\documentclass[preprint,12pt]{elsarticle}
\usepackage{subcaption}
\usepackage{cuted}
\usepackage{todonotes}
\usepackage{amsmath}
\usepackage{amssymb}
\newproof{pf}{Proof}
\usepackage{calc}
\usepackage{nicefrac}
\usepackage{commath}
\usepackage{mathtools}
\usepackage{accents}
\usepackage{interval}
\intervalconfig{
  soft open fences ,
}
\usepackage[normalem]{ulem}
\usepackage{trimclip}
\usepackage[nolist]{acronym}
\usepackage{siunitx}
\usepackage{csquotes}

\usepackage{url}
\usepackage{hyperref}
\hypersetup{
    colorlinks,
    citecolor=black,
    filecolor=black,
    linkcolor=black,
    urlcolor=black,
}

\DeclareMathOperator{\atan2}{atan2}

\DeclareMathOperator{\sign}{sign}

\DeclareMathOperator{\diag}{diag}
\DeclareMathOperator*{\argmax}{argmax}

\DeclareMathOperator{\clamp}{clamp}

\newtheorem{theorem}{Theorem}[section]

\newtheorem{lemma}[theorem]{Lemma}
\newcounter{lemma}

\usepackage[bottom]{footmisc}
\usepackage{graphicx}

\usepackage{diagbox}
\usepackage{multirow}
\usepackage[font=scriptsize]{caption} 

\usepackage{tikz}
\usetikzlibrary{positioning}
\usetikzlibrary{calc}

\newcommand{\red}[1]{{#1}}
\newcommand{\green}[1]{{#1}}

\newcommand{\wrt}{w.r.t.}
\newcommand{\eg}{e.g.}

\newcommand{\ie}{i.e.}

\newcommand\mat{\mathbf}
\newcommand{\vect}[1]{{\mathbf{#1}}}
\newcommand{\hatvect}[2]{\hat{\mathbf{#1}}_{#2}}

\journal{Robotics and Autonomous Systems}
\begin{document}
\begin{frontmatter}

\title{
Distributed \acs{UAV} Formation Control Robust to Relative Pose Measurement Noise}

  \author{Viktor Walter\fnref{ctulabel}}
  \author{Matou\v{s} Vrba\fnref{ctulabel}}
  \author{Daniel Bonilla Licea\fnref{mpulabel}}
  \author{Matej Hilmer\fnref{ctulabel}}
  \author{Martin Saska\fnref{ctulabel}}
  \affiliation[ctulabel]{organization={Faculty of Electrical Engineering, CTU in Prague},
            addressline={Technick\'{a} 2}, 
            city={Prague},
            postcode={16627}, 
            country={Czechia}}

  \affiliation[mpulabel]{organization={Mohammed VI Polytechnic University},
            addressline={Lot 660}, 
            city={Ben Guerir},
            postcode={43150}, 
            country={Morocco}}

  \begin{abstract}
    A technique that allows a \ac{FEC} derived from graph rigidity theory to interface with a realistic relative localization system onboard lightweight \acp{UAV} is proposed in this paper.
    \red{
    The proposed methodology enables reliable real-world deployment of \acp{UAV} in tight formations using relative localization systems burdened by non-negligible sensory noise.
    Such noise otherwise causes undesirable oscillations and drifts in sensor-based formations, and this effect is not sufficiently addressed in existing \ac{FEC} algorithms.
  }
      The proposed solution is based on decomposition of the gradient descent-based \ac{FEC} command into interpretable elements, and then modifying these individually based on the estimated distribution of sensory noise, such that the resulting action limits the probability of \emph{overshooting} the desired formation.
    The behavior of the system was analyzed and the practicality of the proposed solution was compared to pure gradient-descent in real-world experiments where it presented significantly better performance in terms of oscillations, deviation from the desired state and convergence time.%
  \end{abstract}

 \begin{graphicalabstract}
 \includegraphics[width=1.0\linewidth]{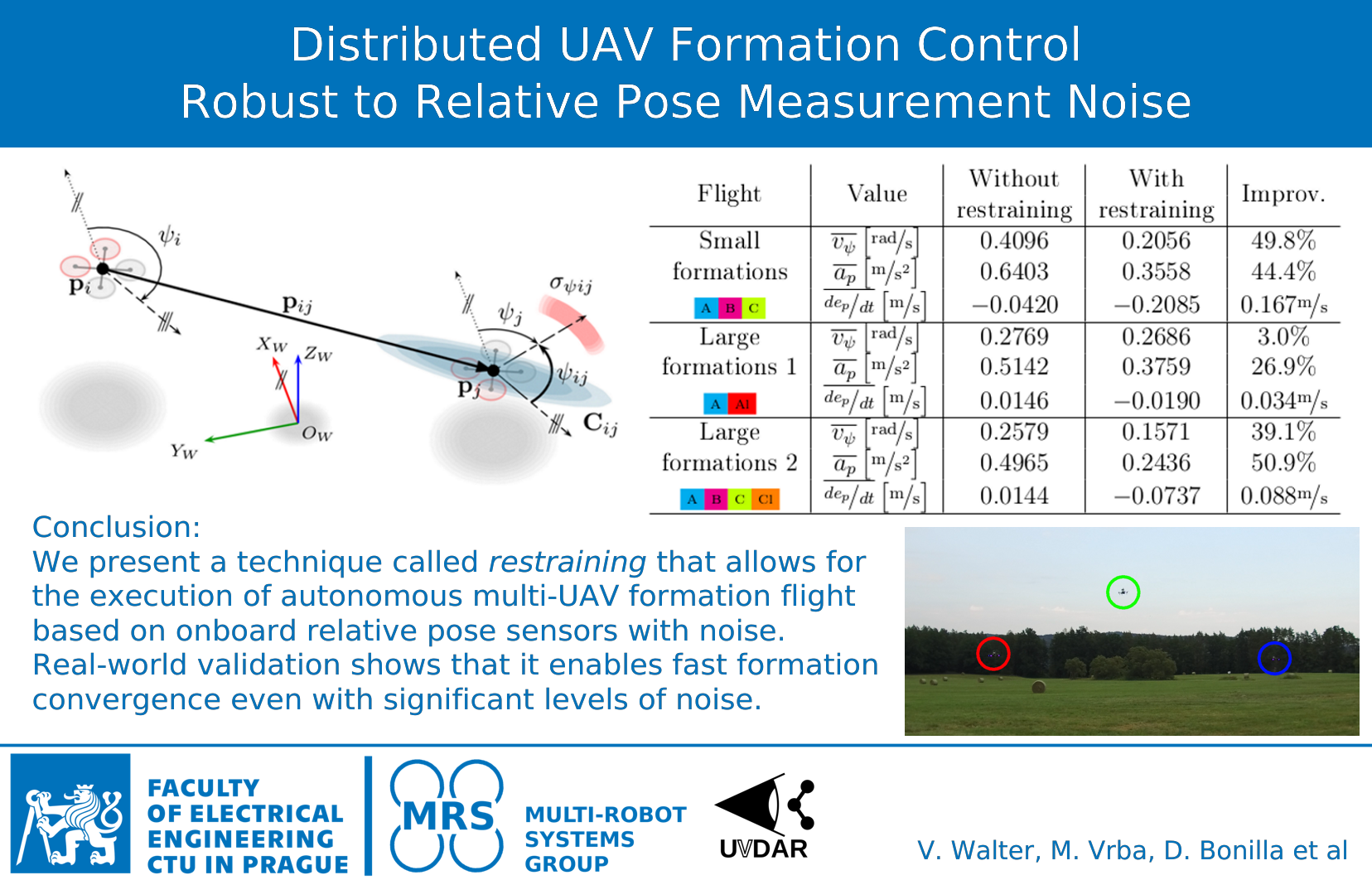}
 \end{graphicalabstract}

 \begin{highlights}
     \item We propose a novel technique for constructing robust distributed \acl{FEC} for autonomous \ac{UAV} teams using onboard relative localization sensors burdened with noise.
     \item Theoretical analysis of this technique is provided, allowing the user to predict its performance under a given configuration.
     \item Experimental verification was performed, showing that application of the proposed technique in real-world scenarios significantly improves performance of real formation flights.
     \item The source code of an implementation of the proposed technique was made publicly available.
 \end{highlights}


\end{frontmatter}

\acresetall
  \begin{figure}
    \includegraphics[width=0.65\linewidth,trim={0.0cm 5cm 0.0cm 9cm},clip]{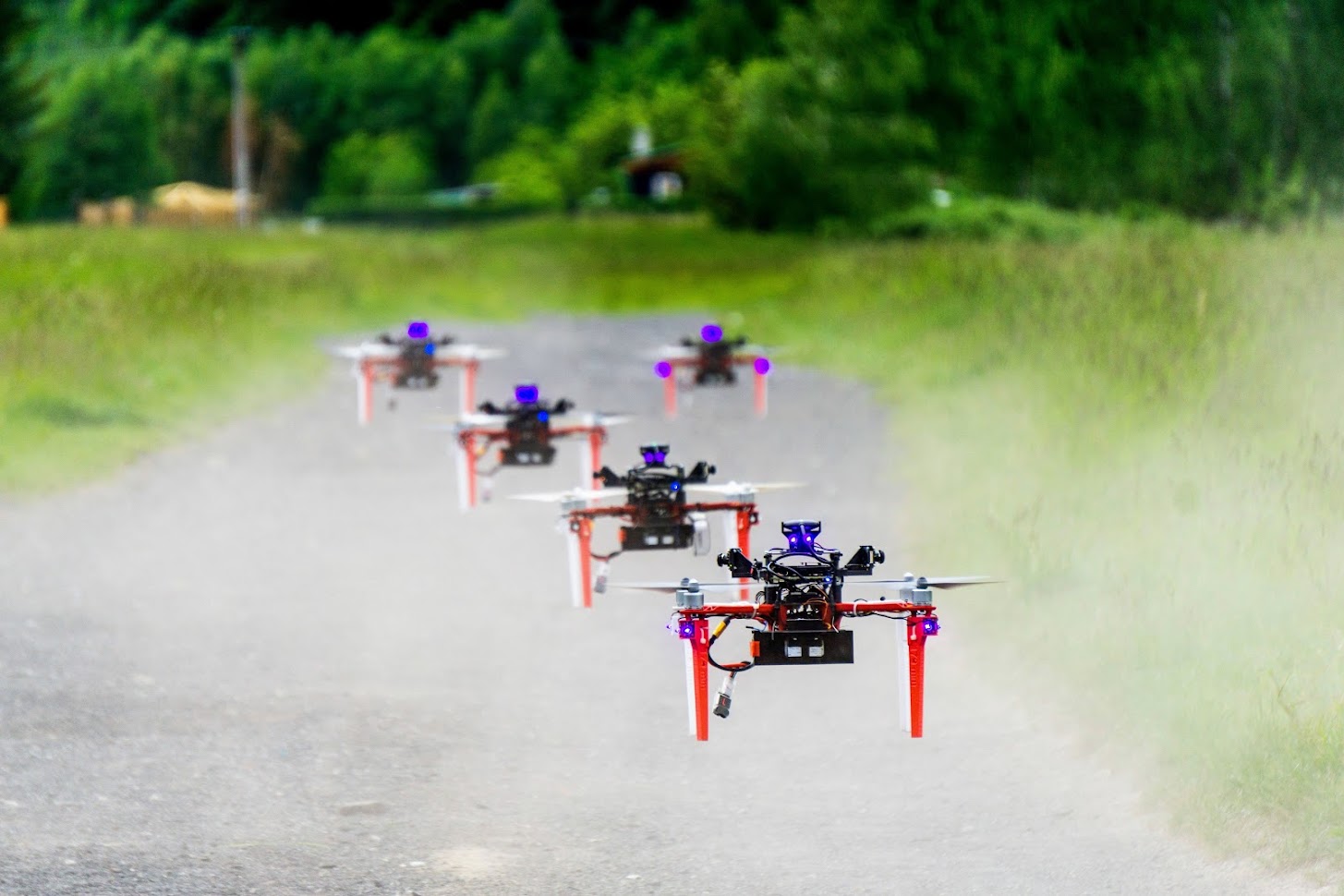}%
    \hfill
    \includegraphics[width=0.30\linewidth,trim={0.0cm -0.5cm -1.0cm -0.5cm},clip]{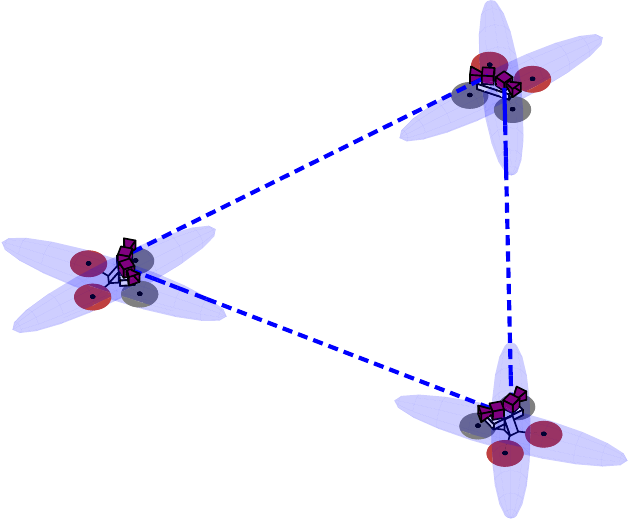}%
    \caption{An example of our fully autonomous \acp{UAV} \cite{MRS2022ICUAS_HW} equipped with the mutual relative localization system \ac{UVDAR}.
    The units depicted are based on the \emph{MRS F450} platforms.
    These devices can cooperate using a mutual relative localization and control scheme, such as the one proposed here.
    The output of the relative localization system is subject to observation noise expressed in terms of covariance of a multivariate Gaussian distribution that is taken into account in the presented method.
     An illustration of 3D ellipsoids representing the distribution of noise in relative position measurement is shown on the right.}
    \label{fig:header_mosaic}
  \end{figure}

  \section{Introduction}
  \label{sec:intro}
  Close cooperation of multiple robotic \acfp{UAV} sharing operational space requires these units to be able to obtain localization of \acp{UAV} in their proximity.
  Such information is necessary for low-level tasks such as collision avoidance, for enforcing a specific layout \cite{swarm_survey}, or for stabilization of the desired shape of a formation.
  Recent technological developments have rapidly progressed towards the possibility of accomplishing cooperative robotic tasks in unstructured environments, such as arbitrary outdoor spaces or underground tunnels.
  While in limited cases the \acp{UAV} can obtain mutual localization from sources of precise global localization, \eg{} \acl{RTK} \acl{GNSS} (\acs{RTK}-\acs{GNSS}) or \ac{MOCAP} systems, this is extremely limiting for a wide range of applications where such a source is unavailable or not reliable enough.
  Typically, these systems require operators to directly access the operational space ahead of robotic deployment in order to prepare the equipment.
  Additionally, these are sensitive to environmental conditions and they require radio communication to share poses of the agents within the team.

  The alternative to global localization infrastructure is to use onboard sensors for mutual relative localization \cite{anonbearingbased,swarm_survey}.
  The two most common modalities of these sensors are distance-based relative localization, which is typically based on the strength or timing of a selected omnidirectionally broadcasted signal \cite{lf_uwb_distance,observability_uwb,valerio_uwb}, and vision-based relative localization using camera systems and image processing.

  In this paper, we propose a novel, distributed, formation-enforcing control law derived from the graph rigidity theory.
  This control law mitigates the effects of noise in vision-based relative localization on the agent motion, without needlessly sacrificing the speed of convergence to the desired formation.

\subsection{Related works}
\label{sec:related}
 Our team has developed an onboard vision-based system for mutual relative localization called \acf{UVDAR} (visible in Fig. \ref{fig:header_mosaic}). This system can operate robustly both indoors and outdoors with challenging conditions (e.g. arbitrary lighting) thanks to its use of emitters and cameras operating together in a shared narrow band of the \ac{UV} spectrum \cite{uvdd1,uvdd2}.
 The system is capable of retrieving the relative position and relative orientation of marked \acp{UAV} \cite{uvdar_ral}.
 Each of these values is provided with the measurement uncertainty expressed as a covariance matrix \cite{midgard}.
The system estimates the distribution of the observation noise using a multi-hypothesis reprojection method for each individual measurement, and the retrieved estimates are processed with a Kalman filter before being used in control.
 Additionally, the system can provide robotic units with unique retrievable identities, and even possesses limited data communication capabilities without radio transmission \cite{horyna2022uvdarcom,licea2023optical}.
 \red{Other systems of relative localization with capability of retrieving relative position and orientation are available as well~\cite{yolo6d,apriltag,uwb_relpos,UWB_LiDAR}, and our work is applicable to use with such systems.}
 In this paper, we assume no communication between the robots. 
 Visual relative localization such as \ac{UVDAR} typically provides accurate information on relative bearing with the option of obtaining information on distance and relative orientation with less precision \cite{vrba2019onboard,vrba2020markerless, markerless:bouman2016, markerless:rozantsev2015, markerless:martinoli2016, markerless:accardo2018, cnnbased:blumenstein2017, markerless:cnn2017,uvdd2}.
A notable recent research also addresses relative localization based on \ac{LIDAR} \cite{vofod}, but such devices are at the moment of writing too costly for deployment on large groups of \acp{UAV}.

 A concurrent research direction addresses new challenges connected with the rise of mutual relative localization systems.
 Particularly relevant is the development of mathematical methods that allow cooperating robots to retrieve, as well as to directly enforce, specific formations in space \cite{cortes2017coordinated,bearingbased,rigiditybased,bearingrigidity_se2,nonholonomic_formation}.
 In \cite{zhang2007coordinated}, an approach where agents are steered in constant motion towards a specific shape is presented.
 Formation control can be de-centralized \cite{decentralized_formation_ground}, which is advantageous for standalone robotic systems where reliance on external infrastructure is problematic.

 The paradigm we base our current research on is built on the graph rigidity theory \cite{graph_rigidity_term}.
 In this work, we will use the term \emph{graph rigidity} to denote the rigidity of an observation graph formed by inter-agent relative pose measurements.
 \subsection{Preliminaries}
 \label{sec:preliminaries}
 It has been shown that a desired formation can, in theory, be achieved merely based on the mutual relative localization of other units if the localization information is either the relative distance \cite{distancerigidity} or the relative bearing \cite{bearingrigidity,bearingbased}.
 Using nomenclature from \cite{bearingrigidity}, all mutual relative measurements within a team form a graph.
 A relative localization-based formation is said to be rigid if there is no more than one formation that can generate a given set of observations.

\begin{figure}
    \includegraphics[trim={0.0cm 11.3cm 0.0cm 11.7cm},clip,width=1.0\linewidth]{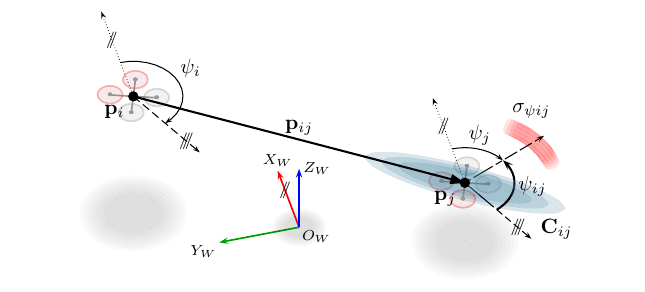}
    \vspace{-2em}
    \caption{Symbols involved in the definition of a formation in this work. Lines marked with $/\!\!/$ or $/\!\!/\!\!/$ are parallel to those with the same marker.}
    \label{fig:formation_geometry}
  \end{figure}

 In the proposed technique, we will consider the relative measurements of multiple neighboring units that can be provided by the \ac{UVDAR} system. This includes the 3D relative position and associated 1D relative orientation comprising the \emph{relative pose measurement} shown in Figure \ref{fig:formation_geometry}, which combines and supersedes the relative bearing and relative distance measurements.
  The relative pose-based formation is always rigid, as long as the observation graph is connected.
  This is because each single relative pose observation in $\mathbb{R}^3 \times \mat{S}^1$ already represents a rigid connection.
  Thus, any two nodes in an observation graph that are connected through a series of such observations are connected rigidly as well.

    To ensure that the \ac{FEC} is distributable,
    the directed observation graph must have no more than a single local sink (representing an observed agent that does not observe others) otherwise the relative pose between such agents is not controllable.
  A single passive agent is permissible and may be used as the de-facto leader of a formation, while the rest of the agents maintains the desired formation around it.

  \red{
    A crucial consideration for our work is that the information provided by real sensors is burdened by observation noise, typically with known statistical properties.
    This aspect is rarely addressed in the theoretical research of formation control, and one of the aims of this work is to bring more attention to this challenge.\\
    Taking this a priori knowledge into account has the potential to significantly increase the reliability of multi-robot systems, thereby addressing the main bottleneck of their applicability.
    Applying real relative pose measurements within a \ac{FEC} derived from graph rigidity theory, while assuming noise-free measurements, leads to undesirable variations in velocity with each new measurement due to the noise.
    Such rapid accelerations and decelerations perturb the vision-based localization system by inducing tilting, which results in large shifts in the image, blurring, loss of tracking, and targets moving outside of the camera's field of view. These effects feed back into the control loop in a way that is difficult to predict.
    In addition to these detrimental effects on the measurements, sensory noise causes persistent, chaotic oscillations of the agents around the desired formation.
    This becomes especially significant when the relative measurements are obtained at discrete times, since an observation error will influence the behavior of the formation for the entire interval before the next measurement is taken.
    Therefore, the aim of the presented work was to develop a sensor-based \ac{FEC} that is robust to such observation noise.
  }

  \red{
    Literature that explicitly addresses imperfect sensing in formation control remains limited.
    In \cite{cortes2009global}, a distributed formation controller is proven robust to bounded measurement errors, converging to a configuration near the desired formation.
    However, the coupling between sensing noise and the resulting parasitic motion is not analyzed, nor is there any attempt to mitigate measurement errors using knowledge of their statistical properties.
    Other studies \cite{distancerigidity2,distancerigidity3} establish stability of \ac{FEC} under certain singular disturbances, but they assume continuous-time control and do not consider persistent noise.

    Sensor noise can be attenuated through repeated measurements and filtering techniques such as Kalman filtering \cite{kalman_filtering,kalman_filtering2,kalman_filtering3}.
    Yet, the effectiveness of filtering depends on factors including noise characteristics, update rate, and knowledge of target dynamics.
    More importantly, filtering only reduces the noise entering \ac{FEC} without altering the underlying sensitivity of the control law; accordingly, we treat the Kalman filter in this work as part of the sensing pipeline rather than a solution to the problem itself.

    In several prior works \cite{convoys,consensus,distributed_adaptive}, relative observation noise is handled implicitly via linear weighting of control actions with respect to distance from desired values.
    This approach neither prevents oscillations near the target formation nor maintains rapid convergence, instead slowing stabilization as demonstrated in our analysis and experiments.
    Another strategy \cite{human_leader_follower} enlarges allowable inter-agent distances in proportion to noise magnitude, which merely introduces collision-avoidance margins while reducing usable operating range.
}

  In this context, the main contributions of this work can be summarized as follows:
  \red{%
\begin{itemize}
  \item A novel technique, called \emph{restraining}, for robust distributed \ac{FEC} of autonomous \ac{UAV} teams based on noisy onboard relative localization measurements.
  \item A theoretical analysis of the proposed technique that enables performance prediction under specified system and noise parameters.
  \item Experimental validation demonstrating the effectiveness of the proposed approach in real-world multi-\ac{UAV} flights.
  \item A publicly available implementation of the proposed technique to support reproducibility and further research.
\end{itemize}
}%
  \section{\red{The proposed method}}
  \label{sec:proposed_method}
  \subsection{Formation-enforcing control}
  \label{sec:fec}
  \green{%
    In this section we derive a formation-enforcing action for $N$ individual \acp{UAV}.
    The action uses for its input the relative pose measurements that each separate \ac{UAV} makes of its neighbors.
  }%
  \begin{figure}
    \centering
    \tikzset{
  rectnode/.style={
    draw,
    black,
    thick,
    anchor=south west,
    align=center
  },
  rectnodegreen/.style={
    draw,
    black,
    fill=green,
    thick,
    anchor=south west,
    align=center
  },
  >=latex
}

\begin{tikzpicture}[x={\textwidth},y={0.8\textwidth}]
  \begin{scope}
    \node[rectnodegreen] (fec) at (0.5,0.5){Formation\\Enforcing\\Control};
    \node[rectnode] (sensor) [left=6.0em of fec] {Relative\\Localization\\Sensor};
    \draw[->, thick] ($(sensor.east) + (0,0.5em)$) -- ($(fec.west) + (0,0.5em)$);
    \draw[->, thick,red] ($(sensor.east) - (0,0.5em)$) -- ($(fec.west) - (0,0.5em)$);
    \draw[->, thick] ($(sensor.east) + (0,0.5em)$) -- ($(fec.west) + (0,0.5em)$) node[midway,above] {$\mathbf{p}_{ij}^m, \psi_{ij}^m$};
    \draw[->, thick,red] ($(sensor.east) - (0,0.5em)$) -- ($(fec.west) - (0,0.5em)$) node[midway,below] {\red{$\mathbf{C}_{ij}, \sigma_{\psi ij}^2$}};
    \coordinate (rl_above) at ($(sensor.north) + (0,2.8em)$);
    \node[rectnode] (sampling) [right=0.5em of rl_above] {Finite\\sampling rate};
    \draw[->, thick] ($(sampling.south) - (2.0em,0)$) -- ($(sensor.north) + (1.85em,0)$) node[midway,right] {$f$};
    \node[rectnode] (noise) [left=0.5em of rl_above] {Measurement\\Noise};
    \draw[->, thick] ($(noise.south) + (2.0em,0)$) -- ($(sensor.north) - (1.78em,0)$) node[midway,right] {$\mathbf{e}_m$};

    \node[draw, thick, circle, minimum size=0.6cm] (cm) [left=7.0em of sensor]{};
    \draw[thick] (cm.north east) -- (cm.south west)
          (cm.north west) -- (cm.south east);
    \fill[black]
    (cm.center) -- (cm.south west)              
          arc[start angle=-135,end angle=-225,radius=0.3cm]  
          -- (cm.center);                   

    \coordinate (cm_left) at ($(cm.west) - (1.5em,0)$);
    \node[rectnode] (neighbor_uavs) [above=1em of cm] {Neighbor\\UAVs};
    \draw[->, thick] (cm_left) -- (cm.west) node[midway,above] {$\mathbf{q}_i$};
    \draw[->, thick] (neighbor_uavs.south) -- (cm.north) node[midway,right] {$\mathbf{q}_j$};

    \node[rectnode] (uav_control) [below=1.0em of fec.south, xshift=-7em] {UAV $i$\\Control stack};
    \coordinate (fec_down) at ($(fec.south) - (0,2.3em)$);
    \draw[-, thick] (fec.south) -- (fec_down) node[midway,right] {$\mathbf{u}_i,\omega_i$};
    \draw[->, thick] (fec_down) |- (uav_control.east);

    \node[rectnode] (physics) [left=1.5em of uav_control] {Physical dynamics};
    \draw[->, thick] (uav_control.west) -- (physics.east);
    \draw[-, thick] (physics.west) -| (cm_left);

    \draw[->, thick] (cm.east) -- (sensor.west) node[midway,below] {$\mathbf{q}_{ij}$};

    \node[rectnode] (desired) [above=1.5em of fec] {Desired\\formation};
    \draw[->, thick] (desired.south) -- (fec.north) node[midway,right] {$\mathbf{q}_{d}$};
  \end{scope}
\end{tikzpicture}%
    \caption{\red{High-level system diagram of a single \ac{UAV} (called Observer) under \ac{FEC} based on real onboard relative localization systems.
    The relative poses $\vect{q}_{ij}$ of neighbors $j$ of an Observer $i$ are observed with a sensor system onboard the Observer.
    The sensor system is imperfect and adds noise $\vect{e}_m$ to the measurement, with the statistical distribution of this noise assumed to be known or estimated.
    This sensor outputs its measurements with finite sampling rate $f$.
    The \ac{FEC} uses this noisy discrete input to generate desired positional and rotational rates $\vect{u}_i$ and $\omega_i$ that minimize the difference between the current estimated formation and the desired formation $\vect{q}_d$.
    The desired rates are assumed constant for the entire sampling period and executed by the Observer's controller.
    All \acp{UAV} involved are assumed to follow this control loop, \ie{} each considers itself to be the Observer and the other \acp{UAV} the neighbors.
    Our contribution is the use of the variances $\mathbf{C}_{ij}$ and $\sigma_{\psi ij}^2$ of the estimated distributions of the sensor noise $\vect{e}_m$ as an additional input into the \ac{FEC}.
    This allows us to mitigate the detrimental effects that the noise and the finite sampling rate have on the overall formation behavior.
    }}
    \label{fig:system_diagram}
  \end{figure}
  \red{We show where the \ac{FEC} under consideration falls within the control loop of a single \ac{UAV} in Figure \ref{fig:system_diagram}.}

  We will be using steps similar to \cite{rigiditybased}, where the formation controller was derived for platforms that can measure only the relative bearings of their neighbors.

  We first denote the world coordinate system as $W$, with origin ${O}_{W}$.
  \green{%
  The horizontal plane ${X}_{W}{Y}_{W}$ is parallel to the surface of the Earth and the axis ${Z_{W}}$ points vertically upwards.
}%
  The pose of the $i$-th agent ($i=1,2,\cdots, N$) in the world coordinate frame is
  \begin{equation}
    \begin{aligned}
    \vect{q}_{i} \triangleq 
    \begin{bmatrix}
      \vect{p}_{i}\\
      \psi_i
    \end{bmatrix},
    \end{aligned}
  \end{equation}
  \green{%
    where $\vect{p}_{i}\in\mathbb{R}^3$ is the position of the agent expressed in the above-mentioned world frame and $\psi_i\in \mat{S}^1$ is its heading.
  The heading is measured as the angle between the axis ${X_{W}}$ and the projection of the front-back axis of an agent onto the horizontal plane.
  }%
  We assume the $i$-th agent follows a first-order integrator dynamic model
  \begin{eqnarray}
    ^{i}\dot{\vect{p}}_i=\vect{u}_i,\\
    ^{i}\dot{\psi}_i=\omega_i,
  \end{eqnarray}
  \green{%
    where $\vect{u}_i$ and $\omega_i$ are the control signals for the velocity and the heading speed respectively.
    The notation $^{i}\cdot_{i}$ indicates that the rate is expressed in the body frame of agent $i$.
    }%
  We also define the formation vector
  \begin{equation}
    \begin{aligned}
    \vect{q} \triangleq 
    {\begin{bmatrix}
      {\vect{q}_{1}}^T, {\vect{q}_{2}}^T,...,{\vect{q}_{N}}^T 
    \end{bmatrix}}^T.
    \end{aligned}
    \label{eq:formation_vector}
  \end{equation}
  
 \green{The noiseless relative pose of the agent $j$  \wrt{} the agent $i$ in the local coordinate frame of the latter agent is}
  \begin{equation}
    \begin{aligned}
    \vect{q}_{ij} = \begin{bmatrix}\vect{p}_{ij}\\\psi_{ij}\end{bmatrix}= \begin{bmatrix}\mat{R}(\psi_i)^T(\vect{p}_{j}-\vect{p}_{i})\\
      \psi_j-\psi_i\end{bmatrix},
    \end{aligned}
    \label{eq:relative_poses}
  \end{equation}
 \begin{equation}
   \mat{R}(\psi_{i}) = \begin{bmatrix}\cos(\psi_{i})& -\sin(\psi_{i})& 0\\ 
     \sin(\psi_{i})& \cos(\psi_{i})& 0\\ 
    0&0&1\end{bmatrix}.
  \end{equation}
 Note that whenever we subtract angles, we wrap the result so that it lies within $\interval{-\pi}{\pi}$.
 This is omitted further in the text for conciseness.
  
  We define the set of all relative poses observed by the agent $i$ as:
  \begin{equation}
   \mathcal{Q}_i\triangleq\{\vect{q}_{ij}\,:\,j=1,\cdots,N,\quad j\neq i,\quad c_{ij}=1 \}, 
  \end{equation}
  where $c_{ij}=1$ if agent $j$ is observed by agent $i$, and $c_{ij}=0$ otherwise.
  Observations $ij$ such that $c_{ij}=1$ form the edges of an \emph{observation graph} $\mathcal{G}$.
  We also define the set of all observed relative poses as
  \begin{equation}
      \mathcal{Q}\triangleq\cup_{i=1}^N\mathcal{Q}_i.
  \end{equation}
  \green{%
    Then the function that transforms the formation vector $\vect{q}$ from eq. (\ref{eq:formation_vector}) to a set of relative poses of neighbors in the body frame of each observing agent is defined as
  }%
  \begin{equation}
    \begin{aligned}
    \varkappa_\mathcal{G}\left(\vect{q}\right) =
    {\begin{bmatrix}
      [\mathcal{Q}]_1^T, [\mathcal{Q}]_2^T,...,[\mathcal{Q}]_{\abs{\mathcal{Q}}}^T
    \end{bmatrix}}^T,
    \end{aligned}
    \label{eq:relative_pose_function}
  \end{equation}
\green{%
  where $[\mathcal{Q}]_i$ is the $i$th element of $\mathcal{Q}$.
  We call $\varkappa_\mathcal{G}\left(\vect{q}\right)$ the \emph{relative pose function}.
  }%
  \green{%
  Thereafter, the action that guides the team to the desired formation $\vect{q}_d$ is obtained by minimizing the squared error between the current and desired relative poses
  }%
    \begin{equation}
    \begin{aligned}
    J\left(\vect{e}_F\right) = \frac{1}{2}  \norm{ \vect{e}_{F}}^2,%
      \label{eq:gradient}
    \end{aligned}
  \end{equation}
  \begin{equation}
    \begin{aligned}
    \vect{e}_F = \varkappa_\mathcal{G}\left(\vect{q}_d\right) - \varkappa_\mathcal{G}\left(\vect{q}\right),
    \end{aligned}
    \label{eq:error_function}
  \end{equation}
where $\norm{\cdot}$ denotes the Euclidean norm.
\green{%
To minimize the error described above, we apply the following local action expressed in the world coordinate frame 
}%
  \begin{equation}
    \begin{aligned}
      \dot{\vect{q}} &= -k_e{\left(\frac{\partial J\left(\vect{e}_F\right)}{\partial \vect{q}}\right)}^T\\
      &= \red{-k_e{\left(\frac{\partial J\left(\vect{e}_F\right)}{\partial \vect{e}_F}\frac{\partial \vect{e}_F}{\partial \varkappa_\mathcal{G}\left(\vect{q}\right)}\frac{\partial \varkappa_\mathcal{G}\left(\vect{q}\right)}{\partial \vect{q}}\right)}^T}\\
      &= -k_e{\left(\frac{\partial \varkappa_\mathcal{G}\left(\vect{q}\right)}{\partial \vect{q}}\right)}^T{\left(\frac{\partial \vect{e}_F}{\partial \varkappa_\mathcal{G}\left(\vect{q}\right)}\right)}^T{\left(\frac{\partial J\left(\vect{e}_F\right)}{\partial \vect{e}_F}\right)}^T\\
      &= -k_e {\left(\frac{\partial \varkappa_\mathcal{G}\left(\vect{q}\right)}{\partial \vect{q}}\right)}^T\left(-\mat{I}\right)\left(\vect{e}_F\right)\\
      &= k_e{\left(\frac{\partial \varkappa_\mathcal{G}\left(\vect{q}\right)}{\partial \vect{q}}\right)}^T\vect{e}_F,%
    \label{eq:gradient_descent}
    \end{aligned}
  \end{equation}
  where $k_e$ is a constant gain.

  \green{%
  If the observation graph $\mathcal{G}$ is connected it is in our case also rigid.
  Therefore, as long as $\mathcal{G}$ is connected the \ac{FEC} can distributed and expressed in the body frame of each agent $i$ as
  }%
  the \ac{FEC} can then be expressed in the body frame of each agent $i$ as
  \begin{footnotesize}
  \begin{equation}
    \label{eq:fec_raw}
    \begin{aligned}
    \vect{u}_{i} &= k_e \!\!\left(  \sum_{
         j\in\mathcal{N}_i}\!\!c_{ij}\left(\vect{p}_{ij}-\vect{p}_{ij}^d\right) - \!\!\sum_{
         j\in\mathcal{N}_i}\!\!c_{ji} {\mat{R}(\psi_{ji})}^T\left( \vect{p}_{ji} -\vect{p}_{ji}^d\right) \!\!\right)\!\!,\\
    \omega_i &= k_e \!\!\left( -\!\!\sum_{
         j\in\mathcal{N}_i}c_{ij}\left( {\vect{p}_{ij}^{T}} \mat{S} \left( \vect{p}_{ij} - \vect{p}_{ij}^d\right)\right)\right.\\
    &\left.+ \sum_{
         j\in\mathcal{N}_i}c_{ij}\left(\psi_{ij} - \psi_{ij}^d \right) - \sum_{
         j\in\mathcal{N}_i}c_{ji} \left(\psi_{ji} - \psi_{ji}^d \right) \!\!\right)\!\!,%
    \end{aligned}%
  \end{equation}%
  \end{footnotesize}%
where $\mathcal{N}_i\!=\!\{\!1,2,\cdots,N\}\!\setminus\! \{i\}$, $\vect{p}_{ij}^d$ and $\psi_{ij}^d$ are the desired relative position and orientation of agent $j$ in the frame of agent $i$ and %
$\mathbf{S}\!=\!\begin{bsmallmatrix}0&-1&0\\1&0&0\\0&0&0\end{bsmallmatrix}\!$.
The full derivation of eq. (\ref{eq:fec_raw}) is provided in \ref{App:fec}.
\green{%
The values of $\vect{p}_{ij}$ and $\psi_{ij}$ are available directly from the local sensors of agent $i$, but $\vect{p}_{ji}$ and $\psi_{ji}$ are not.
To design a fully decentralized control, we will therefore express all the terms $\vect{p}_{ji}$ and $\psi_{ji}$ as a function of $\vect{p}_{ji}$ and $\psi_{ji}$:
}%
  \begin{footnotesize}
  \begin{equation}
    \label{eq:temp_fec}
    \begin{aligned}
    \vect{u}_{i} &= k_e \left(\sum_{j \in \mathcal{N}_i}c_{ij}\left(\vect{p}_{ij}-\vect{p}_{ij}^d\right)\right.\\%
    &\left.- \sum_{j \in \mathcal{N}_i}c_{ij}\mat{R}(\psi_{ij})\left( {\mat{R}(\psi_{ij})}^T(-\vect{p}_{ij}) + {\mat{R}(\psi_{ij}^d)}^T\vect{p}_{ij}^d\right) \vphantom{(\left(\vect{p}_{ij}\right)}\right),\\
    \omega_i &= k_e \left( \sum_{j \in \mathcal{N}_i}c_{ij}\left( {\vect{p}_{ij}^{dT}} \mat{S}^T \vect{p}_{ij} \right) + 2\sum_{j \in \mathcal{N}_i}c_{ij} \left(\psi_{ij} - \psi_{ij}^d \right)\right).
    \end{aligned}
  \end{equation}
  \end{footnotesize}
  Eq. (\ref{eq:temp_fec}) can be further simplified to 
  \begin{footnotesize}
    \begin{equation}
      \begin{aligned}
    \vect{u}_{i} &= k_e \left(  \sum_{j \in \mathcal{N}_i}\!c_{ij}\left(\vect{p}_{ij}-\vect{p}_{ij}^d\right) \!+ \!\!\sum_{j \in \mathcal{N}_i}\!c_{ij}\left( \vect{p}_{ij} - {\mat{R}(\psi_{ij}\!-\!\psi_{ij}^d)}\vect{p}_{ij}^d\right) \!\!\right),\\
    \omega_i &= k_e \left( \sum_{j \in \mathcal{N}_i}c_{ij}\left( {\vect{p}_{ij}^{dT}} \mat{S}^T \vect{p}_{ij} \right) \!+ \!2\!\!\sum_{j \in \mathcal{N}_i}c_{ij} \left(\psi_{ij} - \psi_{ij}^d \right)\!\!\right).
      \end{aligned}
      \label{eq:proportional_clean}
    \end{equation}
  \end{footnotesize}

  \green{%
  To include the sensor measurement uncertainty, we replace $\vect{p}_{ij}$ and $\psi_{ij}$ with their respective measurements.
  These measurements are contaminated with Gaussian noise $\vect{p}_{ij}^m \sim \mathcal{N}\left(\vect{p}_{ij},\mat{C}_{ij}\right)$ and $\psi_{ij}^m \sim \mathcal{N}\left(\psi_{ij},\sigma_{\psi ij}^2\right)$, where  $\mat{C}_{ij}$ is the error covariance matrix for the relative position and $\sigma_{\psi ij}^2$ is the variance for the heading angle difference.
  }%
  We then obtain the \ac{FEC} control scheme to be applied by each agent including (but not yet compensating) noise
  \begin{footnotesize}
    \begin{equation}
      \begin{aligned}
    \vect{u}_{i} &= k_e \left(  \sum_{j \in \mathcal{N}_i}\!c_{ij}\overbrace{\left(\vect{p}_{ij}^m-\vect{p}_{ij}^d\right)}^{\tau_{p_1}}\!+\!\!\sum_{j \in \mathcal{N}_i}\!c_{ij}\overbrace{\left( \vect{p}_{ij}^m - {\mat{R}(\psi_{ij}^m-\psi_{ij}^d)}\vect{p}_{ij}^d\right)}^{\tau_{p_2}} \!\right),\\
    \omega_i &= k_e \left( \sum_{j \in \mathcal{N}_i}c_{ij}\underbrace{\left( {\vect{p}_{ij}^{dT}} \mat{S}^T \vect{p}_{ij}^m \right)}_{\tau_{\psi_1}} + 2\sum_{j \in \mathcal{N}_i}c_{ij}\underbrace{\left(\psi_{ij}^m - \psi_{ij}^d \right)}_{\tau_{\psi_2}}\right).
    \label{eq:proportional}
      \end{aligned}
    \end{equation}
  \end{footnotesize}
  The terms denoted $\tau_{p_1}$, $\tau_{p_2}$, $\tau_{\psi_1}$ and $\tau_{\psi_2}$ are vectors used individually in the following sections to derive a control rule that accounts for observation noise.

  \green{%
  In order to account for observation noise, we interpret the \ac{FEC} input (\ref{eq:proportional}) as a scaled sum of the above vectors.
  Each of these vectors points from the desired relative pose of a neighbor to its measured relative pose.
  }%
  Each of these vectors is subject to noise with its own associated distribution that is Gaussian, or can be approximated as Gaussian. 

\begin{figure}
   \quad\quad\includegraphics[trim={0.0cm 19.5cm 0.0cm 4.5cm},clip,width=\linewidth]{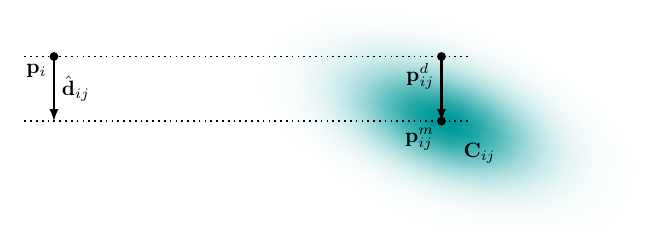}%
  \vspace{-0.8cm}%
  \caption{Simple case of a 2D relative position measurement with Gaussian noise.}%
  \label{fig:gaussian_single_2d}
\end{figure}

An example of such distribution for $\tau_{p_1}$ is in Fig. \ref{fig:gaussian_single_2d} depicting a relative position measurement $\vect{p}_{ij}^m$ measured by agent $i$, where the desired relative position of the neighbor is $\vect{p}_{ij}^d$.
The depicted vector $\vect{d}_{ij}$ represents a control error between the desired and measured relative position of neighbor $j$ \wrt{} agent $i$. 
The measured relative position is burdened by observation noise with a Gaussian distribution, depicted as a cyan ellipsoid.\\
\red{Note that a real relative localization system where \ac{FEC} is applied is assumed to be a discrete system, which means that the relative pose measurements from which $\vect{u}_i$ and $\omega_i$ are derived update with finite rate.
  The values of $\vect{p}_{ij}^m$ and $\psi_{ij}^m$ and their associated variances then relate to the time of the most recent observation made.
  Therefore, the system is assumed to hold the velocity derived from the latest measurement until the next measurement is obtained, thus moving in a piecewise-linear manner.
  We discuss the effects of the above in later sections.%
  }%

\subsection{Mitigation of formation oscillations caused by sensor noise}
\label{sec:restraining}
In this section, we will discuss the core idea used for addressing the sensory noise in a 1D system, and in section \ref{sec:extension_to_4D} we will expand it to the $\mathbb{R}^3\times\mat{S}^1$ case from section \ref{sec:fec}.
\red{%
  We call the proposed technique \emph{restraining}.
  }%

In a real system such as the one discussed here, the sampling rate of relative-localization sensors is limited.
\green{%
This means that a given measurement error will affect the behavior of a formation for a non-zero time period.
}%
Noisy measurements are assumed to be delivered with a constant rate, and upon receiving each new measurement the agents use it to calculate a desired velocity to execute until the next measurement is received.
\red{%
  The state of the system during the time of an upcoming measurement can be predicted based on the state during previous measurement, the values of the measurement and the update period.
  This makes it possible to analyze the \ac{FEC} as a discrete-time system.%
  }%
\subsubsection{Proportional control with noise}
\label{sec:proportional_control_with_noise}
We analyze the behavior of a simple single-agent system using proportional control with measurements burdened with Gaussian noise.
For the sake of simplicity, we will present this idea with a one-dimensional version of the control action.

Consider an agent with a 1D state $x$ (\eg{} its position) with the dynamic model
\begin{eqnarray}
  x[k+1] = x[k] + u[k],
  \label{eq:base_dynamic}
\end{eqnarray}
where $k$ is the discrete time index, and $u[k]$ is the action input.
In our case, the controlled quantity is the rate of change of the state $x$ defined as
\begin{equation}
  u[k] = v[k]T,
  \label{eq:velocity}
\end{equation}
where $v[k]$ is the rate of change and $T$ is the sampling period.
Consider a desired value $d$ that the agent is trying to reach so that $x[N] = d$ at some time step $N$.
Let us define a control error $\Delta_d[k]$ as
\begin{align}
  \Delta_d[k] = d - x[k].
\end{align}

Furthermore, the true control error $\Delta_d[k]$ is not known, but it is measured as
\begin{align}
  \Delta_m[k] = \Delta_d[k] + e_m[k], \hspace{2em} e_m[k] \sim \mathcal{N}\left( 0, \sigma_m \right).
  \label{eq:1d_merr}
\end{align}
The probability distribution of $\Delta_d[k]$ given a measurement $\Delta_m[k]$ is therefore
\begin{equation}
  \Delta_d[k] \sim \mathcal{N}\left( \Delta_m[k], \sigma_m \right).
  \label{eq:1d_x_dist}
\end{equation}

\begin{figure}
  \centering
  \includegraphics[trim={0.0cm 23.1cm 0.0cm 0.0cm},clip,width=0.9\linewidth]{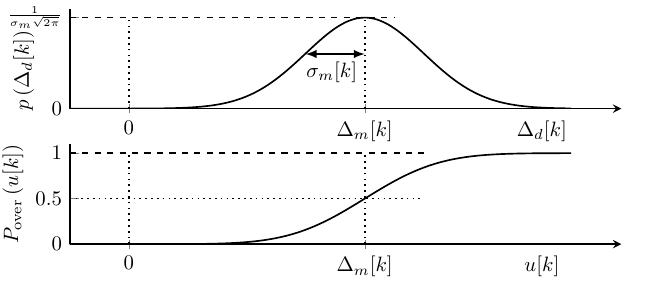}
  \caption{Illustration of the 1D control problem with an unknown true control error $\Delta_d$.
  \textit{Top:} The probability distribution of $\Delta_d[k]$ is Gaussian with mean at the measured value $\Delta_m[k]$ and a standard deviation $\sigma_m[k]$.
  \textit{Bottom:} The probability of overshooting the target $P_{\text{over}}(u[k])$ based on the selected action input $u[k]$ is equal to the CDF of the distribution of~$\Delta_d[k]$.}
  \label{fig:gaussian_cumulative_single_1d}
\end{figure}

\green{%
A maximum-likelihood approach to controlling such system would be to choose the action
}%
\begin{equation}
  v_{\text{ml}}[k] = \argmax_{v[k]} p_{x[k+1]} \left( d \mid \Delta_m[k], ~v[k] \right),
\end{equation}
where $p_{x[k+1]} \left( d \mid \Delta_m[k],~v[k] \right)$ is the probability distribution of $x[k+1]$ given a measurement $\Delta_m[k]$ and an action input $v[k]$.
The solution evaluates to
\begin{equation}
  v_{\text{ml}}[k] = f \Delta_m[k],
\end{equation}
where $f = \frac{1}{T}$ is the system's update rate.
Plugging this into the dynamic equation of the system (\ref{eq:base_dynamic}), we obtain
\begin{equation}
  \begin{aligned}
    x[k+1] &= x[k] + v_{\text{ml}}[k]T \\
           &= x[k] + f \Delta_m[k] T \\
           &= x[k] + \left(d - x[k] + e_m[k] \right)\\
           &= d + e_m[k].
  \end{aligned}
\end{equation}
\green{%
Note that this approach has a \SI{50}{\percent} chance of overshooting the target in every update, as is illustrated in Fig.~\ref{fig:gaussian_cumulative_single_1d}.
This property makes it extremely sensitive to noise.
}%

A conventional approach to control such system is to use proportional control.
The control action $v[k]$ is set at each iteration based on a proportional factor $k_e$ and on the measured control error $\Delta_m[k]$.
This allows to control the aggressivity of the system and reduce the sensitivity to measurement noise, model mismatch, etc.
Specifically, the control action is obtained using proportional control as
\begin{equation}
  \label{eq:clean_regression}
  \begin{aligned}
    v_{\text{p}}\left[k\right] = k_e \Delta_m[k]
  \end{aligned},
\end{equation}
and the resulting action input $u[k]$ is then
\begin{equation}
  \begin{aligned}
    u[k] = v_{\text{p}}\left[k\right]T = \frac{v_{\text{p}}\left[k\right]}{f} = \frac{k_e}{f}\Delta_m[k]
  \end{aligned}.
  \label{eq:clean_displacement}
\end{equation}
Since the system depends on the ratio $\frac{k_e}{f}$, then for notational simplicity we define the constant $k_{ef} = \frac{k_e}{f}$.
Plugging this into eq.~\eqref{eq:base_dynamic}, the system dynamic is then
\begin{equation}
  \begin{aligned}
    x[k+1] = x[k] + k_{ef}\Delta_m[k].
  \end{aligned}
  \label{eq:clean_dynamic}
\end{equation}

In a noiseless scenario, the measurement $\Delta_m[k]$ is equal to the true control error $\Delta_d[k]$ at each time step $k$, and the state moves towards the desired value according to the geometric progression
\begin{equation}
  \label{eq:clean_proportional}
  \begin{aligned}
    x\left[k+1\right]
    &=  x[k] + k_{ef}\Delta_d[k] \\
    &=  x[k] + k_{ef}\left( d - x[k] \right) \\
    &=  d + \left(x\left[k\right] - d\right)\left(1-k_{ef}\right),
  \end{aligned}
\end{equation}
which resolves to
\begin{equation}
  \begin{aligned}
    x\left[k\right] &= d + \left(x\left[0\right]-d\right){\left(1-k_{ef}\right)}^{k}.
    \label{eq:geometric_progression}
  \end{aligned}
\end{equation}
The controlled variable $x\left[k\right]$ will clearly converge to the desired value $d$ if $k_{ef}\in(0,2)$.
The agent will converge monotonically if $k_{ef} \in (0,1]$. 
If $k_{ef} \in \left(1,2\right)$, the agent will converge to the desired value $d$, but it will be affected with damped oscillations, which is contrary to our stated goals.
We will therefore restrict the value of $k_{ef}$ to $(0,1]$.
It is worth noting that for $k_{ef} = 1$, this control law is equivalent to the maximum-likelihood approach.

If we include the measurement noise as defined in eq.~\eqref{eq:1d_merr}, the geometric progression from eq. \ref{eq:clean_proportional} becomes
\begin{equation}
  \label{eq:noisy_proportional}
  \begin{aligned}
    x\left[k+1\right] &= x[k] + k_{ef}\Delta_m[k]\\
    &= x[k] + k_{ef}\left(\Delta_d[k] + e_m[k]\right)\\
    &= d + \left(x\left[k\right] - d\right)\left(1-k_{ef}\right) + k_{ef} e_m[k],\\
  \end{aligned}
\end{equation}
with the additional term $k_{ef} e_m[k]$ representing the influence of sensory noise.
Selecting a suitable $k_{ef}$ allows the user to tune a trade-off between sensitivity to noise and reaction speed of the system.
However, such system with non-zero noise will still oscillate when $x$ is close to the desired state $d$.

\begin{lemma}
  Barring infinite measurement rate or a zero control action, it is \emph{impossible} to use proportional control to enforce a stationary stable state if discrete measurements of this state used in the control law are subject to Gaussian noise.
  \label{lm:gauss_limit}
\end{lemma}
\begin{pf}
  If multiple agents follow the control law described in eq. (\ref{eq:noisy_proportional}), the presented combination of proportional control and observation noise with Gaussian distribution actually leads to a Gaussian distribution of agents%
  \footnote{In \ref{App:A.N}, we show that even with the proposed modification the resulting poses can be accurately approximated with a Gaussian distribution.}.
  \green{%
    Repeated addition of Gaussian noise to a value (represented by the noise inherent in $k_{ef}e_m\left[k\right]$) leads to a behavior called random walk, the uncertainty of which also has Gaussian distribution.
    }%

  The state in system (\ref{eq:noisy_proportional}) has a variance evolving according to
  \begin{equation}
    \label{eq:noise_variance_evolution}
    \begin{small}
      \begin{aligned}
        \mathrm{Var}\left[x\left[k+1\right]\right]&=\mathrm{Var}\left[k_{ef}e_m\left[k\right]\right] + \mathrm{Var}\left[\left(1-k_{ef}\right)x\left[k\right]\right]\\
        &=k_{ef}^2\sigma_{m}^2+(1-k_{ef})^2\mathrm{Var}\left[x\left[k\right]\right].
      \end{aligned}
    \end{small}
  \end{equation}
  Using the equation for geometric series, the closed form of the above is
  \begin{equation}
    \begin{aligned}
      \label{eq:noise_variance_closed_form}
      \mathrm{Var}\left[x\left[k+1\right]\right] &=
      \frac{{\sigma_{m}}^2 k_{ef}\left(1-\left(1-k_{ef}\right)^{2k}\right)}{2-k_{ef}}\\
      &+{\left(1-k_{ef}\right)}^{2k} \mathrm{Var}\left[x\left[0\right]\right].
    \end{aligned}
  \end{equation}
  As $k$ approaches infinity, the variance of the system approaches
  \begin{equation}
    \begin{aligned}
      \label{eq:noise_variance_stabilization}
      \mathrm{Var}\left[x\left[k\rightarrow\infty\right]\right]&= {\sigma_{ss}}^2 = \frac{{\sigma_{m}}^2 k_{ef}}{2-k_{ef}}
    \end{aligned}
  \end{equation}
  for $k_{ef} \in \left(0,2\right)$ (for other values of $k_{ef}$, $\sigma_{ss} = \infty$).
  The two interacting behaviors in the system therefore converge to a Gaussian localization noise of individual agents with non-zero standard deviation of
  \begin{equation}
    \label{eq:sigma_ss}
    \sigma_{ss} = \sigma_{m} \sqrt{\frac{k_{ef}}{2-k_{ef}}}.
  \end{equation}%
\end{pf}
This is the negative effect that is mitigated using the approach proposed in the following section.

\subsubsection{1D restraining}
\label{sec:1d_restraining}
In order to reduce the oscillations while maintaining the required agility of the multi-robot system, we propose to exploit the sensory model as follows.
Consider that the measured control error (as defined in eq.~\eqref{eq:1d_merr}) comprises of the true control error and the measurement noise:
\begin{equation}
  \Delta_m[k] = \Delta_d[k] + e_m[k], \hspace{2em} e_m[k] \sim \mathcal{N}\left( 0, \sigma_m[k] \right).
\end{equation}
Therefore, if $\Delta_m[k]$ is large compared to $\sigma_m[k]$ then $\Delta_m[k]$ likely consists predominantly of the control error $\Delta_d[k]$ with the measurement noise $e_m[k]$ being a minor component.
On the other hand, if the state $x[k]$ is sufficiently close to $d$ so that $\Delta_m[k]$ is small compared to $\sigma_m[k]$, the measured control error $\Delta_m[k]$ likely stems predominantly from the measurement noise.
In other words, the closer the agent measures the desired state $d$ to be, the more likely it is that correcting the measured control error $\Delta_m[k]$ is unnecessary and such correction may in fact increase the true control error $\Delta_d[k]$ by needlessly retreating or by overshooting the target.
If this is not taken into account, an agent near $d$ will tend to frequently change its direction of motion, which is contrary to our goals.
\green{%
Thus, if $\Delta_m[k] \gg \sigma_m[k]$ then the measured difference vector can be safely applied in a proportional control, enabling rapid convergence.
Conversely, if $\Delta_m[k] \ll \sigma_m[k]$ then the control action should be suppressed to minimize reaction to noise.
}%

\begin{figure}
  \centering
  \includegraphics[trim={0.0cm 25.0cm 0.0cm 0.0cm},clip,width=0.9\linewidth]{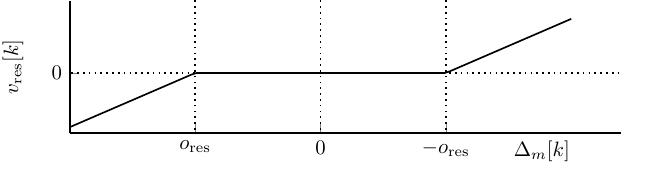}
  \caption{The velocity set with our proposed controller in 1D case.
  The region within $\Delta_m[k] \in \interval{s_{\mathrm{res}}}{-s_{\mathrm{res}}}$ represents a region where the agent remains purposefully passive.
  }
  \label{fig:clamping}
\end{figure}

In order to derive a modified action satisfying these requirements, %
we first propose an abstract control scheme with a region of passivity around the desired state.
Assuming the same dynamics as in eq.~\eqref{eq:base_dynamic}, we define a new control law
\begin{equation}
  v[k] = v_{\text{res}}[k] = k_e\left(\clamp{\left(\Delta_s\left(\Delta_m[k]\right),\Delta_m[k]\right)}\right),
  \label{eq:restrained_velocity}
\end{equation}
where $\Delta_s\left(\Delta_m[k]\right)$ is a function that generates a \emph{restrained control error} reduced from $\Delta_m$ based on the input values as%
\begin{equation}
  \Delta_s\left(\Delta_m[k]\right) =
    \Delta_m + \sign{\left(\Delta_m\right)}o_{\mathrm{res}}
    \label{eq:restrained_control_error}
\end{equation}
The function
\begin{equation}
  \begin{aligned}
    \clamp{\left(y,a\right)} &=\begin{cases}%
      y & \text{if\quad} y\cdot a \in \interval[open left]{0}{{a}^2}\\%
      0 & \text{otherwise},%
    \end{cases}
  \end{aligned}
  \label{eq:clamp}
\end{equation}
serves to nullify the control action within the region $\interval{o_{\mathrm{res}}}{-o_{\mathrm{res}}}$.
Thus, $v_{\text{res}}[k]$ is a piecewise-linear function of $\Delta_m[k]$, as shown in Fig. \ref{fig:clamping}.
  This control aims to steer an agent towards the state $x = d + \sign{\left(\Delta_d[0]\right)} o_\text{res}$.

  However, due to the influence of sensory noise on the measurement $\Delta_m[k]$, the control will still have a stochastic behavior, aiming in each step $k$ towards a \emph{restrained setpoint}
\begin{equation}
  \begin{aligned}
    s_\text{res} = x[k] + \Delta_s.
  \end{aligned}
  \label{eq:restrained_setpoint}
\end{equation}
The agent may incorrectly evaluate itself as being inside the region $\interval{o_{\mathrm{res}}}{-o_{\mathrm{res}}}$ or vice-versa.
This is why the controlled velocity in eq. (\ref{eq:restrained_velocity}) has a dead zone around the desired state. If $\Delta_s$ is set up correctly, it prevents the agent from reacting to measurements that are statistically likely to cause \emph{overshooting} of the desired state $d$ by the agent.
To take into account the distribution of sensor noise, we condition the offset $o_\text{res}$ on the standard deviation $\sigma_m[k]$ of the current measurement $\Delta_m[k]$.
We detail how $o_{\text{res}}$ is chosen in the following section.

\subsubsection{Setpoint selection}
\label{sec:setpoint_selection}

Let us define \textit{overshooting} of the controller as
\begin{equation}
  \label{eq:overshooting}
  \sign{(\Delta_d[k+1])} \neq \mathrm{sign}(\Delta_d[k]),
\end{equation}
or in other words the controller overshoots when the state in the next update lies on the \enquote{opposite side} of the desired value $d$.
We propose to select a setpoint $s[k]$ with offset $o_\text{res}[k]$ \wrt{} measured control error $\Delta_m[k]$, such $o_\text{res}$ is calculated with regard to the probability of overshooting the desired value $d$ assuming that the setpoint is reached in the next step (i.e. $k_{ef} = 1$).
This can also be interpreted as selecting the setpoint based on the probability that it itself overshoots the desired value.
Formally, let us define the overshooting probability for a general setpoint $s$ as
\begin{equation}
  P_{\mathrm{over}} = P\left(\sign{(d - s[k])} \neq \sign{(d - x[k])}\right).
\end{equation}

The proportional approach in section \ref{sec:proportional_control_with_noise} is equivalent to always choosing a setpoint $s_\text{ml} = x[k]+u_\text{ml}[k]$, such that action input $u_\text{ml}[k] = \Delta_m[k]$.
The probability distribution of $\Delta_d[k]$ from the perspective of agent measuring it with noise is Gaussian as per eq.~\eqref{eq:1d_x_dist}.
Therefore, the probability of overshooting based on the action input $u[k]$ is the cumulative probability density function of the Gaussian distribution with the mean $\Delta_m[k]$ and standard deviation $\sigma_m[k]$, as illustrated in Fig.~\ref{fig:gaussian_cumulative_single_1d}.
If we plug the action $u_\text{ml}[k]$ into the equation of this CDF, we get 
\begin{equation}
  \begin{aligned}
    P_{\mathrm{over}, \mathrm{ml}}
    &= \Phi\left( \frac{ u_\text{ml}[k] - \Delta_m[k]}{\sigma_m[k]} \right) \\
    &= \Phi\left( \frac{ \Delta_m[k] - \Delta_m[k]}{\sigma_m[k]} \right) \\
    &= \Phi\left( 0 \right) = 0.5, \\
\end{aligned}
\end{equation}
where $\Phi(\cdot)$ is the CDF of the Gaussian distribution with zero mean and unit standard deviation.
Note that in this \enquote{maximum-likelyhood} approach the overshoot probability is fairly high which, combined with the finite sampling frequency $f$, causes stochastic oscillations of the agent about the desired state $d$ that scale with $k_{ef}$ (as per eq.~\eqref{eq:sigma_ss}) when the state $x[k]$ is close to $d$.

\begin{figure}
  \centering
  \includegraphics[trim={0.0cm 23.5cm 0.0cm 0.0cm},clip,width=0.9\linewidth]{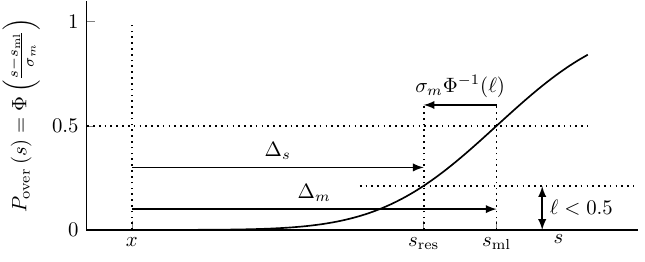}
  \caption{If the measured relative value of a target state is subject to observation noise, controlling the current state of the system towards the mean of that measurement at $s_\mathrm{ml}$ risks overshooting the target with probability $P_\mathrm{over} = 0.5$.
  This probability can be reduced to an arbitrary value $\ell < 0.5$ by choosing a restrained setpoint $s_\mathrm{res}$ that is closer to the current state than $s_\mathrm{ml}$ by $\abs{\sigma_m[k] \Phi^{-1}(\ell)}$.
  If $\ell < 0.5$, then $\sigma_m[k] \Phi^{-1}(\ell) < 0$.}
  \label{fig:setpoint}
\end{figure}

Instead of using the maximum-likelihood setpoint selection, we select the \emph{restrained setpoint} $s_{\text{res}}$ (see Fig. \ref{fig:setpoint}) based on a specified maximum \emph{allowed} probability $\ell \in \left( 0, 0.5 \right]$ of overshooting, as defined above.
Given a sensory noise with a known Gaussian distribution, with mean $\Delta_m[k]$ and standard deviation $\sigma_m[k]$, a $s_\mathrm{res}$ conforming to this probability $\ell$ is
\begin{equation}
  \begin{aligned}
    s_{\text{res}}\left[k\right] &= x[k] + \Delta_m[k] + \sign{\left(\Delta_m[k]\right)}\sigma_m[k] \Phi^{-1}(\ell)\\%
    &= x[k] + \Delta_m[k] + \sign{\left(\Delta_m[k]\right)}o_\text{res}(\sigma_m[k])\\%
    &= x[k] + \Delta_s(\Delta_m[k],\sigma_m[k])%
  \end{aligned}
  \label{eq:level_limit}
\end{equation}
where $\Phi^{-1}(\cdot)$ is the inverse function of $\Phi(\cdot)$.
Deploying the obtained \emph{restrained control error} $\Delta_s$ above into the control law proposed in eq. (\ref{eq:restrained_velocity}), the resulting dynamic equation of the system is
\begin{equation}
  \footnotesize
  \label{eq:position_change}
  \begin{aligned}
    x[k+1] &= x[k] + u_\text{res}\left(\Delta_m[k], \sigma_m[k]\right)\\
    u_\text{res}[k] &=  v_\text{res}\left(\Delta_m[k], \sigma_m[k]\right)T\\
    &=  k_{ef}\clamp{\left(\Delta_s\left(\Delta_m[k], \sigma_m[k]\right), \Delta_m[k]\right)}\\
    &=  k_{ef}\clamp{\left(\Delta_m[k]+\sign{\left(\Delta_m[k]\right)}\sigma_m[k]\Phi^{-1}\left(\ell\right),\Delta_m[k]\right)}.
  \end{aligned}
\end{equation}
This makes for a flexible control that is robust to sensory noise, without the need for manually tuning the proportional factor with regards to the specific standard deviation of the sensor noise present.
The user only needs to set the overshooting probability $\ell$.
The control will steer the agent to a state defined by the limit on the probability of overshooting the target, and if it finds the agent to be located beyond this threshold it will become inactive.
\green{%
In essence, this acts as a spatial probabilistic motion filter.
Despite the fact that it does not actively steer the agents closer to the desired state than $s_\mathrm{res}$, it allows the agents to get closer to the desired state than $s_\mathrm{res}$ through the effects of the sensory noise.
}%
The probability that the agent will actively move in a given iteration decreases with \abs{d-x} according to eq. (\ref{App:A.2.3}) in the \ref{App:A.L}.
Therein, we show that the minimum motion probability is $2\ell$ at $x=d$.
\green{%
Therefore the agent is more likely to remain stationary near the desired state than far from it.
In essence, we are enforcing a narrower probability distribution of the state of the agent about the desired value in comparison with pure proportional control.
}%
\green{%
The system is statistically stable overall, as is shown in \ref{App:A.L}.
The expected state of the agent converges towards $d$ and the expected variance at the stable state is
}%
\begin{equation}
  {\sigma}_{ss,\text{res}}^2=\frac{k_{ef}{\sigma_{m,\text{fin}}}^2}{2-k_{ef}}\exp{\left(\beta(k_{ef})\Phi^{-1}(\ell)\right)}
\end{equation}
where $\beta\left(\cdot\right)$ is an empirical function with example values found in Table \ref{tab:Beta}.
Reducing $\ell$ therefore reduces the stable-state noise in the system.
For example, if $k_{ef} = 0.5$ and $\ell = 0.3$
the distribution of the agents position will have zero-mean around the true set point and variance
\begin{equation}
  \label{eq:reduction_example}
  \sigma_{ss,\text{res}}=0.8051\sigma_{ss}.
\end{equation}

\green{%
If $\ell$ is set to 0.5 the dead zone $\interval{o_{\mathrm{res}}}{-o_{\mathrm{res}}}$ collapses into a point.
In such case the control will be equal to the naive proportional control from section \ref{sec:proportional_control_with_noise}, as in the case $s_\mathrm{res} = d$.
}%
\green{%
If $\abs{d-x} \gg \sigma_m$, the agent will behave similarly to being driven by a proportional control with convergence to the desired formation largely unimpeded by restraining.
}%

The net effect of the above is that the \emph{stable state} variance of the agent's displacement from $d$ is \emph{reduced} compared to $\sigma_{ss}$ from eq. (\ref{eq:sigma_ss}).
The magnitude of said noise reduction is dependent on $\ell$ and on $k_{ef}$, and the specifics of the behavior are analyzed in depth in \ref{App:A.L}.

\subsubsection{Analysis of the control performance}
\label{sec:analysis_simulated_1d}

\green{%
To first demonstrate the operation of the proposed control scheme in a simulated 1D system we establish a large number (10,000) of virtual agents.
These agents are set to act according to the dynamics definition in eqs. (\ref{eq:base_dynamic}) and (\ref{eq:velocity}).
}%
Without loss of generality, we can choose the unit of the state $x$ to be meters, so that we may provide an intuitive understanding in terms of robotic motion control.
The parameters of the simulation are selected such that the relevant effects of the proposed control can easily be demonstrated on reasonably scaled plots.
However, the presented effects persist across a wide range of parameters, albeit with changes in their relative magnitudes.

The agents were first randomly displaced according to a Gaussian distribution around the mean equal to $d$, with a standard deviation of their positions being $\sigma_I = $ \SI{100}{\meter}.
\green{%
  Agents then measured their distances to the desired value $\Delta_m$ at a rate $f=\SI{10}{\hertz}$.
  Each such measurement is burdened by random additive noise with a Gaussian distribution of $\sigma_m = \SI{3}{\meter}$.
  }%
The proportional factor $k_{ef}$ was set to 0.5.
The whole simulation was executed for a range of thresholds $\ell \in
    {\begin{bmatrix}
      0.05,0.1,0.2,0.25,0.3,0.4,0.45,0.5 
    \end{bmatrix}}$.
For $\ell = 0.5$, the control is equivalent to the proportional control without the proposed modification.%

\begin{figure}
  \centering
  \includegraphics[width=0.9\linewidth]{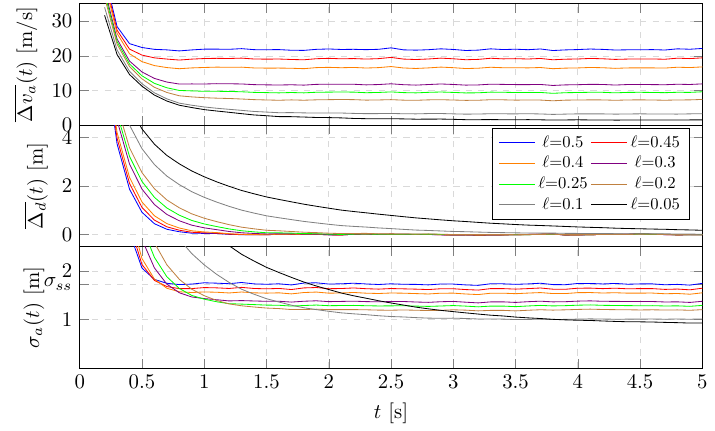}\vspace{-1em}
  \caption{Simulation results of the 1D case of proportional control with the restrained target value.
  Each curve is obtained by averaging the motion of 10,000 virtual agents with random initial distribution and observation noise.
  The initial conditions were the same for each value of $\ell$.
  Top: the average change in velocity between two subsequent target measurements.
  Middle: the average distance from the target position.
  Bottom: standard deviation from the target.}
  \label{fig:1d_behavior}
\end{figure}

\green{%
Fig. \ref{fig:1d_behavior} shows the time evolution of three relevant variables in the simulation:
\begin{itemize}
  \item{the average difference $\overline{\Delta v_a}(t)$ in rate of change between two consecutive measurements}
  \item{the average distance $\overline{\Delta_d}(t)$ from the desired position $d$}
  \item{the root mean square deviation $E_{\text{RMSD}}(t)$ of the agents.}
\end{itemize}
}%
As seen in the plots, $\overline{\Delta_d}$ converged to zero.
For $\ell = 0.5$ the value of $E_{\text{RMSD}}$ converged to the theoretical value from eq. (\ref{eq:sigma_ss}) of $\sigma_{ss} = \SI{1.73}{\meter}$ for the selected parameters, following Lemma \ref{lm:gauss_limit}.
This represents the upper bound of the stable-state localization noise in the system.
By decreasing the threshold $\ell$, the rate of change in $x$ becomes progressively more steady, which is exactly the goal we set out to achieve.
In addition, although the coalescence of the agents towards a stable distribution gets slower with decreasing $\ell$, the standard deviation of their distribution \emph{actually becomes smaller}.
  \green{%
    This outcome may be seen as counter-intuitive - despite  the \emph{restrained control error} $\Delta_s$ being less accurate to the expected relative position of $d$ than the measured control error $\Delta_m$, agents steered using $\Delta_s$ this way end up closer to $d$ than if we were to steer them using $\Delta_m$.
    }%

Lastly, we analyzed how the proposed system compares to a pure proportional control in terms of the tradeoff between the speed of convergence and the stable-state distribution and velocity.
To do this, we have performed multiple simulations of the above system for a set of values of $\ell$, including $\ell = 0.5$.
\green{%
  We have swept through values of $k_{ef} \in \interval{0.02}{0.97}$ to establish whether the proposed \emph{restraining} technique can provide better results than merely tuning the proportional gain.
  }%
For each run of the simulation, we have evaluated three parameters:
\begin{itemize}
  \item{time $t_c$ to convergence to stable-state oscillations,}
  \item{the standard deviation $\sigma_t$ from the desired state in the stable-state,}
  \item{the average change $\overline{\Delta v}$ in velocity between subsequent linear movements based on their latest measurements.}
\end{itemize}
\green{%
To evaluate these parameters, we will define the time of convergence to the stable state in the simulation data with $M$ samples as follows:
convergence is achieved at the first such iteration that the state becomes closer to the desired state than 3 times the standard deviation of the rest of the recorded states, or: 
}%
\begin{equation}
  \begin{aligned}
    t_c &= \frac{k_c}{f}, \\
    k_c &= \min\left\{k \in 0..M : \abs{x[k-1]-d} < 3\sigma_{\text{fin}}[k]\right\}, \\
    \sigma_{\text{fin}}[k] & = \sqrt{ \text{var}\left(\left\{x[k']-d : k' \in k..M \right\}\right) }.
  \end{aligned}
  \label{eq:convergence_definition}
\end{equation}
The stable-state standard deviation $\sigma_t$ and average convergence velocity change $\overline{\Delta v}$ are then defined as
\begin{align}
  \sigma_t &= \sigma_{\text{fin}}[k_c], \label{eq:stable_state_sigma} \\
  \overline{\Delta v} &= \frac{1}{(M\!-\!2)}\sum_{k=2}^M{\left( \left(x[k]\!\!-\!\!2x[k\!\!-\!\!1]\!\!+\!\!x[k\!\!-\!\!2]\right)f\right)}. \label{eq:average_velocity}
\end{align}

\begin{figure}
  \centering
  \includegraphics[width=1.0\linewidth]{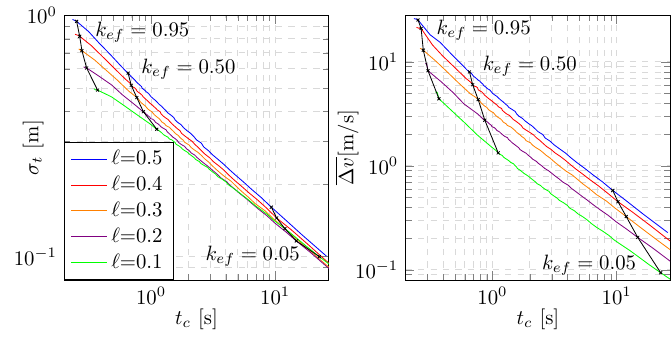}\vspace{-1em}
  \caption{Left: a comparison of the trade-off between the stable-state standard deviation $\sigma_t$ from the desired state and the time $t_c$ it took to reach the stable-state oscillation threshold for various values of $\ell$.
  The curves show the presented value \wrt{} a sweep across values of $k_{ef} \in \interval{0.02}{0.97}$.
  On the right, we show a similar comparison between the average change in velocity between subsequent measurement-induced motions and the time of convergence, in the same ranges of $k_{ef}$.
  These plots are produced by 1D simulation of the presented dynamic system, with framerate $f = 10$ and $\sigma_m = 1$ for a specific scale.
  The plot for $\ell = 0.5$ corresponds to pure proportional control; in either case, we obtain better combination of parameters by applying our restraining technique, since the other plots are contained under the curve of $\ell=0.5$.
  }
  \label{fig:1d_tradeoff}
\end{figure}

Using the above definitions, the results of our simulation are presented in Fig. \ref{fig:1d_tradeoff}, where each point in the plots was averaged from 1,000 runs with the same parameters.
The figure shows that using the proposed technique can obtain better combinations of behavior properties than what is possible by merely changing the value of $k_{ef}$.
This is evident by all the plots with $\ell < 0.5$ contained \emph{under} the plot of $\ell = 0.5$ corresponding to the pure proportional control.
We arbitrarily chose $\sigma_m = 1$ and $f = \SI{10}{\hertz}$, since these variables merely scale the values of $\sigma_t$ and $\overline{\Delta v}$ without changing the relations.

\green{%
The smaller the value of $k_{ef}$, the lower will be the improvement achieved by using the proposed technique compared to pure proportional control.
The above implies that the proposed system is especially useful when the involved relative localization system has limited output rate.
}%

It must be noted that, in a general formation control, agents strive to reach a relative displacement with respect to other agents.
If the observed agent is static, the observing agent behaves as detailed above, where we substitute the desired state $d$ with the desired displacement with respect to the other agent.
However, two agents may observe each other at the same time.
\green{%
In case of mutual observation, the agents will both move towards their desired relative displacement \wrt{} each other, each according to their current observation.
}%
In \ref{App:C}, we show that in such a case the range of $k_{ef}$ for which the system is generally stable reduces from $\interval[open]{0}{2}$ to $\interval[open]{0}{1}$, which in turn translates to the range $\interval[open]{0}{0.5}$ for monotonical convergence to the desired displacement.
This is not a stringent requirement, since $k_{ef} = 0.5$ is in practice an extreme value - for example, with $f = \SI{10}{\hertz}$ this translates to a local proportional factor $k_e = 5$ times the displacement.
The observation does imply however, that the increasing number of connections in the formation leads to increased limits on the dynamics that enable convergence in a controlled manner. 
As we note in the appendix, the dynamics of the mutual-observation case will eventually turn into the behavior of a single-observation case, since one of the two agents will stop with high probability near the desired displacement, after which the remaining agent will move as if with respect to a static desired state.

\subsection[Extension to 4D case]{Restraining for \ac{FEC} in $\mathbb{R}^3\times\mat{S}^1$}
  \label{sec:extension_to_4D}
Returning to the \ac{FEC} derived in eq. (\ref{eq:proportional}), we will now explore how the \emph{restraining} technique above can be exploited for a more complex, multi-dimensional problem.

We aim to mitigate the oscillations in the proportional control caused by noisy measurements.
To do this, we apply the restraining techniques developed in section \ref{sec:1d_restraining}, particularly eqs. (\ref{eq:restrained_velocity}) and (\ref{eq:level_limit}).
However, these equations are developed for a one-dimensional system while (\ref{eq:proportional}) is multi-dimensional.
\green{%
  Applying the dead zone directly to modify the original cost function (\ref{eq:gradient}) is not practical in our case, primarily due to the fact that the resulting \ac{FEC} would still depend on ground-truth states of the agents.
  This is mainly due to the geometrical interdependence of the relative position and orientation of agents.

  Instead, we apply the proposed restraining technique \emph{post-hoc} directly on the \ac{FEC} as formulated for an ideal system (\ref{eq:proportional}).
  We do this by addressing noise individually for each of the terms $\tau_{p_1}$, $\tau_{p_2}$, $\tau_{\psi_1}$, and $\tau_{\psi_4}$, viewing each of these components as an equivalent to the control error $\Delta_m$ from eq. (\ref{eq:1d_merr}).

  The distribution of noise for relative pose measurements is defined as a 3D covariance of relative position $\mat{C}_{ij}$ and a 1D standard deviation $\sigma_{\psi ij}$ for relative orientation.
  We will first convert these values into 1D standard deviations equivalent to $\sigma_m$ from the 1D control law in (\ref{eq:position_change}) through specific projections as detailed in subsections below.
We then then determine the corresponding setpoint (\ref{eq:level_limit}) for each term and compose the terms back to provide the control velocity based on (\ref{eq:restrained_velocity}).
}%

We also have to extend the overshooting definition provided for the one dimensional case in (\ref{eq:overshooting}) to generalize to the higher dimensional case.
\green{%
  We define overshooting as such a change from the original state $\vect{p}_{ij}[k]$ to a new state $\vect{p}_{ij}[k+1]$, where its orthogonal projection onto the line connecting $\vect{p}_{ij}[k]$ and the desired state $\vect{p}_{ij}^d$ falls beyond the desired target from the perspective of the original position, \ie{} \emph{if}:
  }%
\begin{equation}
  \label{eq:ND_overshooting}
  \langle\vect{p}_{ij}[k+1]-\vect{p}_{ij}[k], \vect{p}_{ij}^d-\vect{p}_{ij}[k] \rangle> \norm{\vect{p}_{ij}^d-\vect{p}_{ij}[k]}^2.
\end{equation}
where $\langle\cdot,\cdot\rangle$ is the inner product. 
Note, that for clarity of explanation in section \ref{sec:1d_restraining} the observer controls its own state towards the desired state.
\green{%
As a result, the definition of the control errors is reversed here with respect to said section.
This is because in the situation of real \acp{UAV} the desired relative poses of neighbors are attached to the body of each agent.
The agent can thus control these desired  poses in the environment by its own body motion, while it strives to match them with measured relative poses beyond its direct control.
}%

In cases of bilateral observation, we assume that two mutually-observing agents both follow the proposed control.
In \ref{App:C} we show in a 1D case that this behavior transforms into unilateral observation once one of the agents enters its dead zone.

\green{%
In the following sections, we will be discussing the four terms $\tau_{p_1}$, $\tau_{p_2}$, $\tau_{\psi 1}$ and $\tau_{\psi 2}$ from \ac{FEC} in equation (\ref{eq:proportional}).
We will show how each of them can be adjusted with the proposed \emph{restraining} technique to obtain a fully restrained \ac{FEC} for a team of multirotor \acp{UAV}.
}%

\subsubsection{\red{Direct positional difference term $\tau_{p_1}$}}
The term $\tau_{p_1}$ can be interpreted as the control error pertaining to translational difference between the measured and desired relative position of agent $j$ from the perspective of agent $i$.
\green{%
Consider that the measurement of the $j$-th agent's position by agent $i$ has a multivariate Gaussian distribution with a known covariance $\mat{C}_{ij}$ and mean $\vect{p}_{ij}$
}%
\begin{equation}
  \vect{p}_{ij}^{m}[k] \sim\mathcal{N}\left(\vect{p}_{ij},\mat{C}_{ij}\right),
  \label{eq:gaussian_relpos}
\end{equation}
and thus, conversely, given the measurement $\vect{p}_{ij}^{m}[k]$, the likelihood that $\vect{p}_{ij}$ is located at point $\vect{p}{}$ is
\begin{equation}
\!\!\!\!P\left(\vect{p}_{ij}\!=\!\vect{p}{}\right)\!=\!\frac{1}{\sqrt{{(2\pi)}^3\abs{\mat{C}_{ij}}}} e^{\frac{1}{2}{\left(\vect{p}{}-\vect{p}_{ij}^{m}[k]\right)}^T\mat{C}_{ij}^{-1}\left(\vect{p}{}-\vect{p}_{ij}^{m}[k]\right)}.
\end{equation}
\green{%
We restrict the point $\vect{p}{}$ to lie along the line defined by the points $\vect{p}_{ij}^d$ and $\vect{p}_{ij}^{m}[k]$.
This way the likelihood function becomes a 1D Gaussian distribution with mean $\vect{p}_{ij}^{m}[k]$ and variance $\sigma^2_{pij}$. 
}%

The standard deviation $\sigma_{pij}$ of the 1D Gaussian distribution is obtained by calculating the Mahalanobis distance between $\vect{p}_{ij}^{m}[k]$ and $\vect{p}_{ij}^d$, and dividing their Euclidean distance by it:
\begin{equation}
  \begin{aligned}
    \sigma_{pij} &= \frac{\norm{\vect{p}_{ij}^m - \vect{p}_{ij}^d}}{\sqrt{\left(\vect{p}_{ij}^m - \vect{p}_{ij}^d\right)^T {\mat{C}_{ij}}^{-1} \left(\vect{p}_{ij}^m - \vect{p}_{ij}^d\right)}}.
  \end{aligned}
  \label{eq:sigma_pij}
\end{equation}
We can use this standard deviation to obtain a \emph{restrained setpoint} $\vect{s}_{p_1}$ equivalent to $s_\text{res}$ from eq. (\ref{eq:level_limit}):
\begin{equation}
  \begin{aligned}
    &\vect{s}_{p_1} = \sigma_{pij} \frac{\vect{p}_{ij}^m - \vect{p}_{ij}^d}{\norm{\vect{p}_{ij}^m - \vect{p}_{ij}^d}} \left(\Phi^{-1}\left(\ell\right)\right)+\vect{p}_{ij}^m\\%
    &= \frac{\vect{p}_{ij}^m - \vect{p}_{ij}^d}{\sqrt{\left(\vect{p}_{ij}^m - \vect{p}_{ij}^d\right)^T {\mat{C}_{ij}}^{-1} \left(\vect{p}_{ij}^m - \vect{p}_{ij}^d\right)}} \Phi^{-1}\left(\ell\right)+\vect{p}_{ij}^m.
  \end{aligned}
  \label{eq:level_limit_3d}
\end{equation}
\subsubsection{\red{Reciprocal position difference term $\tau_{p_2}$}}
\begin{figure}
  \centering
  \includegraphics[trim={0.0cm 2.0cm 0.0cm 2.0cm},clip,width=0.8\linewidth]{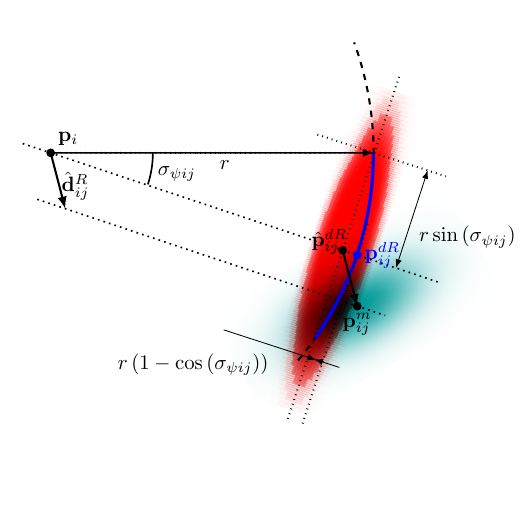}
  \caption{Extended case of 2D relative pose measurement with both ends of the difference vector being burdened by observation noise.}
  \label{fig:gaussian_extended_2d}
\end{figure}

The term $\tau_{p_2}$ can be interpreted as the control error pertaining to the estimate by agent $i$ of how its own relative position in the frame of observed agent $j$ differs from the desired state.
It is composed of a three-dimensional Gaussian random variable (\ie{} $\vect{p}_{ij}^m$) and a three-dimensional non-Gaussian random variable (\ie{} $\mat{R}(\psi_{ij}^m-\psi_{ij}^d)\vect{p}_{ij}^d$). The non-Gaussian variable is the result of a three-dimensional nonlinear function applied to a single Gaussian variable ($\psi_{ij}^m$).
We will refer to this non-Gaussian random variable as $\vect{p}_{ij}^{dR}$. This is illustrated in Fig. \ref{fig:gaussian_extended_2d}, where the noise of $\vect{p}_{ij}^m$ is illustrated as a cyan Gaussian blob and one standard deviation of the probability distribution of $\vect{p}_{ij}^{dR}$ is shown in blue.\\
We start by approximating $\vect{p}_{ij}^{dR}$ with a new Gaussian variable $\hatvect{p}{ij}^{dR}$,
such that one standard deviation of the distribution of $\hatvect{p}{ij}^{dR}$  encloses one standard deviation of the distribution of $\vect{p}_{ij}^{dR}$ (Fig. \ref{fig:gaussian_extended_2d}). 
\green{%
This method differs from the more standard Taylor expansion-based method \cite{hendeby2007nonlinear}.
We decided against using the standard method because it would either not take into account the true distribution being biased in the direction of $-\vect{p}_{ij}^{dR}$ in the case of first-order Taylor approximation, or would allow this bias to shift the mean of the linearized distribution beyond $\vect{p}_i$ in the case of the second-order Taylor approximation.
}%

We obtain the mean $\hatvect{p}{ij}^{dR}$ according to
\begin{equation}
  \label{eq:t2_mean_shift}
  \begin{aligned}
    \hatvect{p}{ij}^{dR} &= {\begin{bmatrix}\begin{bmatrix}[\vect{p}_{ij}^{dR}]_1 & [\vect{p}_{ij}^{dR}]_2\end{bmatrix}  \cos\left(\sigma_{\psi ij}\right) & [\vect{p}_{ij}^{dR}]_3\end{bmatrix}}^T,
  \end{aligned}
\end{equation}
where $[\cdot]_n$ denotes the $n$-th element of a vector.

\green{%
We then generate a Gaussian distribution centered on this new starting point $\hatvect{p}{ij}^{dR}$.
We set the eigenvalue corresponding to the vertical eigenvector to zero, making this a degenerate or \enquote{flat} distribution.
}%
\green{%
The other two eigenvalues correspond to the eigenvectors set to point tangentially and perpendicularly to horizontal components of $\hatvect{p}{ij}^{dR}$, respectively.
The covariance $\mat{C}_{tij}$ of the new distribution is constructed as:
}%
\begin{equation}
  \begin{aligned}
    \vect{v}_{ij1}^d &= {\left[[\vect{p}_{ij}^{dR}]_1 \quad [\vect{p}_{ij}^{dR}]_2 \quad 0\right]}^T,\\
    \vect{v}_{ij2}^d &= {\left[-[\vect{p}_{ij}^{dR}]_2 \quad [\vect{p}_{ij}^{dR}]_1 \quad 0\right]}^T,\\
    \vect{v}_{ij3}^d &= {\left[0 \quad 0 \quad 1\right]}^T,\\
    \mat{V}_{ij} &= \left[\frac{\vect{v}_{ij1}^d}{\norm{\vect{v}_{ij1}^d}} \quad\frac{\vect{v}_{ij2}^d}{\norm{\vect{v}_{ij2}^d}} \quad \frac{\vect{v}_{ij3}^d}{\norm{\vect{v}_{ij3}^d}}\right],\\
    \sigma_{\psi ij}^c &= \min\left(\sigma_{\psi ij}, \frac{\pi}{2}\right),\\
    \mat{L}_{ij} &= \diag{\!\!\left(\!\norm{\vect{v}_{ij1}^d}\!\!\begin{bmatrix}{1\!-\!\cos\!\left(\!\sigma_{\psi ij}^c\!\right)}^2 ~ {\sin\left(\!\sigma_{\psi ij}^{c}\!\right)}^2 ~ {\delta}^2 \end{bmatrix}\!\right)},\\
    \mat{C}_{tij} &= \mat{V}_{ij} \mat{L}_{ij} \mat{V}_{ij}^{-1},
  \end{aligned}
\end{equation}
where $\delta$ is an arbitrarily small, non-zero value chosen to prevent numerical instabilities.
This new distribution is shown in red in Fig. \ref{fig:gaussian_extended_2d}
and indicates that the starting point can also lie \emph{inside} the circle containing the original distribution, which is a conservative assumption.

Thus, we approximate the term $\tau_{p_2}$ with a random Gaussian variable with the mean $\vect{p}_{ij}^m - \hatvect{p}{ij}^{dR}$ and covariance matrix: 
\begin{equation}
  \begin{aligned}
    \mat{C}_{cij} = \mat{C}_{ij} + \mat{C}_{tij}.
  \end{aligned}
\end{equation}
Then the set point for this term is 
\begin{equation}
    \vect{s}_{p_2} = \frac{\vect{p}_{ij}^m - \hatvect{p}{ij}^{dR}}{\sqrt{{\left(\vect{p}_{ij}^m - \hatvect{p}{ij}^{dR}\right)}^{T} {\mat{C}_{cij}}^{-1} \left(\vect{p}_{ij}^m - \hatvect{p}{ij}^{dR}\right)}} \Phi^{-1}\left(\ell\right)+\vect{p}_{ij}^m.
  \label{eq:level_limit_3d_combined}
\end{equation}

\subsubsection{\red{Relative bearing difference term $\tau_{\psi_1}$}}
The term $\tau_{\psi_1}$ can be interpreted as the control error of the orientation of agent $i$ obtained from the difference in relative bearing of the desired and measured position of agent $j$.
Let us start by factorizing this term as follows:
\begin{equation}
  \begin{aligned}
    \tau_{\psi_1} = \left( \vect{p}_{ij}^{dT} \mat{S}^T \vect{p}_{ij}^m \right)& = \norm{\mat{F}\vect{p}_{ij}^d }\cdot \norm{\mat{F}\vect{p}_{ij}^m}\cdot\sin{\left(\alpha\right)},
    \label{eq:term_4}
  \end{aligned}
\end{equation}
where $\alpha$ is the angle formed by the desired and measured bearings of agent $j$ from the local perspective of agent $i$, with the positive orientation being the same as the positive orientation of the angles $\psi$, and with $\mat{F} = \diag{\left({\begin{bmatrix}1 & 1 & 0\end{bmatrix}}^T\right)}$ removing the vertical components from the two vectors.
  Note that the magnitude of the term $\tau_{\psi_1}$ also increases with the norms of $\mat{F}\vect{p}_{ij}^m$ and $\mat{F}\vect{p}_{ij}^d$, which is due to linearization used to obtain the local control eq. (\ref{eq:proportional}).

\begin{figure}
  \includegraphics[trim={0.0cm 19.5cm 0.0cm 3.5cm},clip, width=\linewidth]{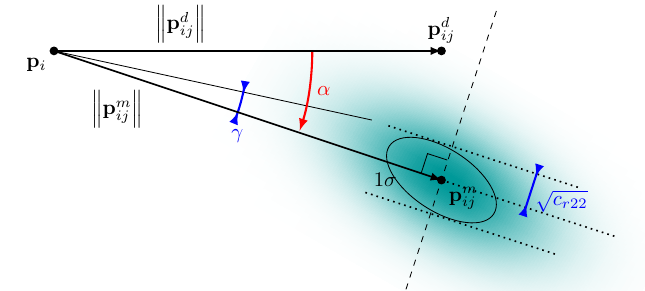}
    \caption{Special case of 2D relative pose measurement for the orientation command based on the relative bearing error. The ellipse denotes the probability density corresponding to one standard deviation.}
    \label{fig:gaussian_bearing}
  \end{figure}

\green{%
  A change in the orientation of the observing agent does not affect the norm of $\vect{p}_{ij}^m$.
  Therefore, only the angular bearing error $\alpha$ will be the restrained variable.
  }%
  As before, we need to express a setpoint that is offset from the mean of the statistical distribution of the term expressed as a mean and a variance.
We will need to be able to evaluate the Gaussian distribution function of $\vect{p}_{ij}^m$ at a given relative bearing $\alpha$.
For this purpose, we need to know the standard deviation $\sigma_{\beta ij}$ of the bearing measurement ${\beta}_{ij} = \zeta\left(\vect{p}_{ij}^m\right)$ where $\zeta\left(\cdot\right)$ calculates the horizontal bearing angle of a vector in the argument as $\zeta\left(\vect{x}\right) = \atan2\left([\vect{x}]_2,[\vect{x}]_1\right)$.
This value is related to noise in the relative position $\vect{p}_{ij}^m$, in the sense that the true distribution of the bearing measurement noise corresponds with the projection of the distribution of the relative position measurement noise onto a circle centered over the observer, conditioned on the bearing angle.
\green{%
The exact distribution of the bearing error after such non-linear projection is non-Gaussian and thus we need a linear approximation.
}%
\green{%
  We will use the following orthogonal projection, obtaining the 1D standard deviation $\gamma$ of bearing measurement:%
  }%
\begin{equation}
  \begin{aligned}
    \mat{C}_r &= \mat{R}({-\beta}_{ij})\mat{C}_{ij}{\mat{R}({-\beta}_{ij})}^T =
  \begin{small}
    \begin{bsmallmatrix}{c_r}_{11} & {c_r}_{12} & \\{c_r}_{21} & {c_r}_{22}&\\ & & \ddots\end{bsmallmatrix}
  \end{small}
      ,\\
      \gamma &= \arctan{\left(\frac{\sqrt{{c_r}_{22}}}{\norm{\vect{p}_{ij}^m}}\right)}.
      \end{aligned}
\end{equation}
This geometric consideration is illustrated in Fig. \ref{fig:gaussian_bearing}.
Thus, for our 3D case an approximation ${\sigma}_{\beta ij}^{'}$ of the standard deviation ${\sigma}_{\beta ij}$ of the above distribution can be obtained as%
\begin{equation}
  \begin{aligned}
    \sigma_{\beta ij} &\approx \gamma \approx \tan{\left(\gamma\right)},\\
    \sigma_{\beta ij}^{'} &= \frac{\sqrt{{c_r}_{22}}}{\norm{\vect{p}_{ij}^m}}.
  \end{aligned}
\end{equation}
Next, we modify the proportional action corresponding to $\left( {\vect{p}_{ij}^{dT}} \mat{S}^T \vect{p}_{ij}^m\right)$ for an equivalent effect as eq. (\ref{eq:level_limit_3d}) and (\ref{eq:level_limit_3d_combined}).
\green{%
We obtain a restrained modification of $\vect{p}_{ij}^m$ by rotating it in the horizontal plane towards the horizontal projection of $\vect{p}_{ij}^d$, in the sense of the shortest angle.
}%
Therefore, 
\begin{equation}
{\hatvect{p}{ij}^{c3}} = \mat{R}\!\left(\! {\sigma}_{\beta ij}^{'} \sign{\!\left(\!\zeta\!\left(\vect{p}_{ij}^d\right)\!-\!\zeta\!\left(\vect{p}_{ij}^m\right)\!\!\right)}\Phi^{-1}\left(\ell\right)\!\right)  \vect{p}_{ij}^m.
    \label{eq:level_limit_bearing}
\end{equation}

Applying this to eq. (\ref{eq:term_4}) yields the modified term: 

\begin{equation}
  \begin{footnotesize}
  \begin{aligned}
    \left( {\vect{p}_{ij}^{dT}} \mat{S}^T \hatvect{p}{ij}^{c3}\right) &= \norm{\mat{F}\vect{p}_{ij}^d}\cdot \norm{\mat{F}\vect{p}_{ij}^m}\cdot\sin{\left(\alpha-\sign{\left(\alpha\right)}{\sigma}_{\beta ij}^{'} \Phi^{-1}\left(\ell\right)\right)}.
  \end{aligned}
  \end{footnotesize}
\end{equation}

\green{%
  Note that this term scales not only with the orientation of the agent, but also with the square of the neighbor distance.
}%
\green{%
  Thus, implementations of this system should contain safety mechanisms that limit the resulting rotation to less than $\pm\pi$\SI{}{rad} in one step.
  }%
This can be enforced by choosing a sufficiently small $k_{ef}$ for the final control.

\subsubsection{\red{Relative heading term $\tau_{\psi_2}$}}
The term $\tau_{\psi_2}$ may be interpreted as the control error of the orientation of agent $i$ obtained from the difference in the measured and desired relative heading of agent $j$.
\green{%
  The term already represents a one-dimensional Gaussian variable for which we have a known standard deviation $\sigma_{\psi ij}$ so we can directly apply the eq. (\ref{eq:level_limit}):
}%
\begin{equation}
  \begin{aligned}
    s_{\psi_{2}} &= \sigma_{\psi ij} \sign{\left(\psi_{ij}^m - \psi_{ij}^d\right)} \Phi^{-1}\left(\ell\right)+\psi_{ij}^m.
    \label{eq:level_limit_orientation}
  \end{aligned}
\end{equation}

\subsubsection{Combining the restrained terms}

\green{%
We now have a complete modified variant of the local control action from eq. (\ref{eq:proportional}):
}%
\begin{equation}
  \begin{small}
  \begin{aligned}
    \vect{u}_{i} &= k_e \sum_{\substack{j=1,...,N\\c_{ij}=1}} \clamp{\left(\vect{s}_{p_1}-\vect{p}_{ij}^d,\vect{p}_{ij}^m-\vect{p}_{ij}^d\right)}\\
    &+ k_e \sum_{\substack{j=1,...,N\\c_{ij}=1}} \clamp{\left( \vect{s}_{p_2} - \hatvect{p}{ij}^{dR},\vect{p}_{ij}^m-\hatvect{p}{ij}^{dR}\right)},\\%
    \omega_i &= k_e \sum_{\substack{j=1,...,N\\c_{ij}=1}} \clamp{\left( {\vect{p}_{ij}^{dT}} \mat{S}^T \hatvect{p}{ij}^{c3},{\vect{p}_{ij}^{dT}} \mat{S}^T \vect{p}_{ij}^m\right)}\\
    &+ 2 k_e \sum_{\substack{j=1,...,N\\c_{ij}=1}} \clamp{\left(s_{\psi_{2}} - \psi_{ij}^d,\psi_{ij}^m-\psi_{ij}^d\right).}
    \label{eq:proportional_modded}
  \end{aligned}
  \end{small}
\end{equation}
\green{%
This new control law that reduces changes in velocity induced by measurement noise, thus mitigating oscillations.
}%
\green{%
  The $\clamp\left(\cdot\right)$ function for vectors is defined as
  }%
\begin{equation}
  \begin{aligned}
  \clamp{\left(\vect{y},\vect{a}\right)} &=\begin{cases}%
    \vect{y} & \text{if\quad} \vect{y}\cdot \vect{a} \in \interval[open left]{0}{\norm{\vect{a}}^2}\\%
      \vect{0} & \text{otherwise}.%
    \end{cases}
  \end{aligned}
\end{equation}
Note, that the geometrical considerations used with some of the terms from eq. (\ref{eq:proportional}) are reminiscent of those used for controlling purely bearing-based \cite{bearingbased} and distance-based \cite{distancerigidity,distancerigidity2,distancerigidity3,distancerigidity4} formations.
We believe that our example can also be used as a template for applying the proposed technique to such systems.

  \begin{figure}
    \centering
    \begin{subfigure}[t]{0.8\linewidth}
      \caption{Two mutually observing \acp{UAV}.}
      \vspace{-0.5em}
      \includegraphics[width=\linewidth]{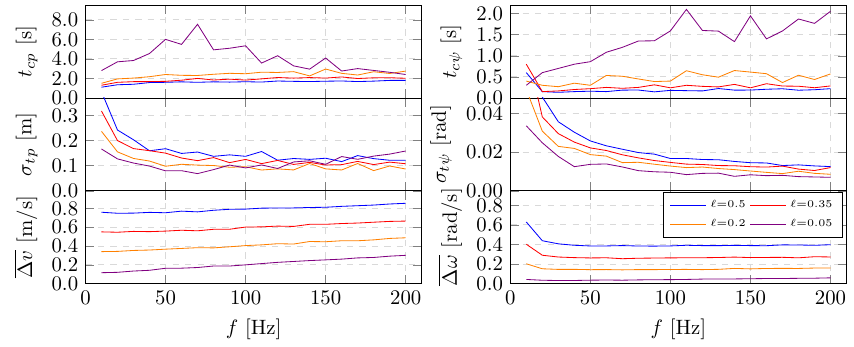}
      \label{fig:4D_N2F}
    \end{subfigure}
    \begin{subfigure}[t]{0.8\linewidth}
      \vspace{-0.8em}
      \caption{Three mutually observing \acp{UAV}.}
      \vspace{-0.5em}
      \includegraphics[width=\linewidth]{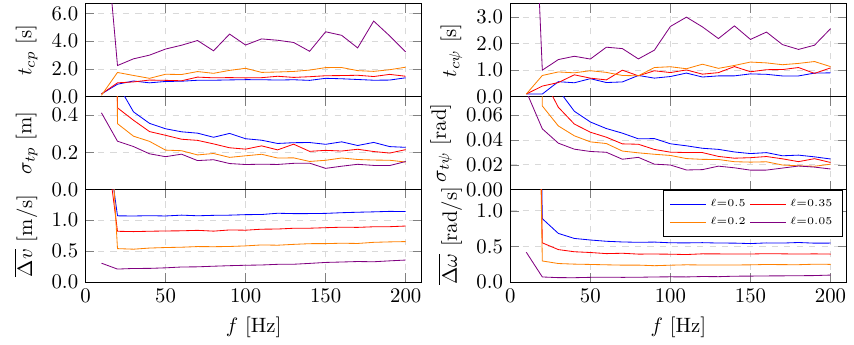}
      \label{fig:4D_N3F}
    \end{subfigure}
    \begin{subfigure}[t]{0.8\linewidth}
      \vspace{-0.8em}
      \caption{Six mutually observing \acp{UAV}.}
      \vspace{-0.5em}
      \includegraphics[width=\linewidth]{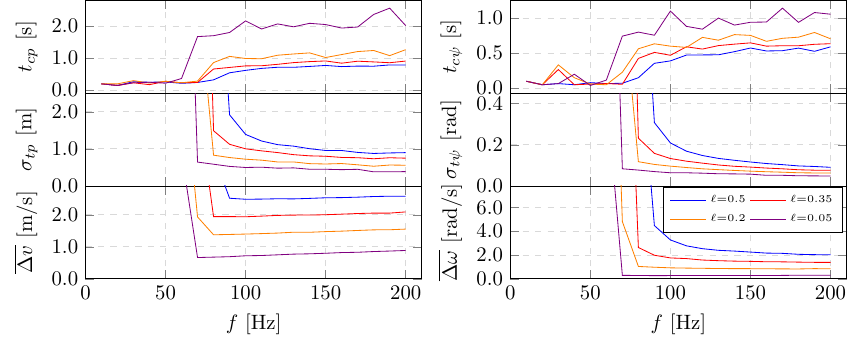}
      \label{fig:4D_N6F}
    \end{subfigure}
    \begin{subfigure}[t]{0.8\linewidth}
      \vspace{-0.8em}
      \caption{Six \acp{UAV} with incomplete observation graph $\mathcal{G}$.}
      \vspace{-0.5em}
      \includegraphics[width=\linewidth]{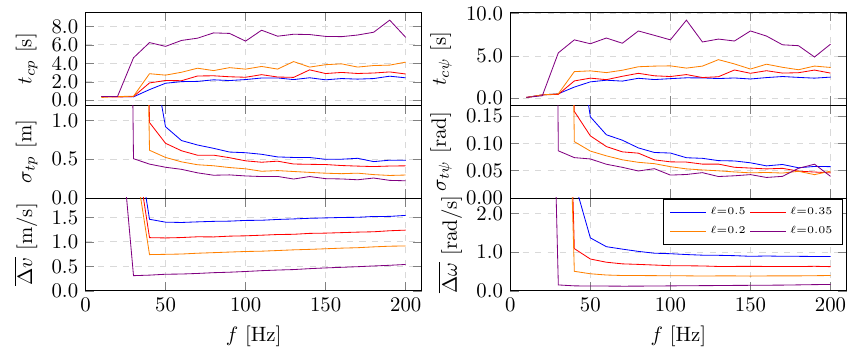}
      \label{fig:4D_N6P}
    \end{subfigure}
    \vspace{-1.5em}
    \caption{Results from the simulated formation flights of various configurations.}
    \label{fig:4D_exps}
  \end{figure}
  \section{Simulation testing}
  \label{sec:simulation}
  \green{%
  In order to test and showcase the presented technique, we have implemented a simulation of the proposed formation control.
  Within this simulation the agents move as first-order integrators and they base their actions on simulated relative localization with characteristics mimicking those measured in \ac{UVDAR} \cite{uvdar_ral,midgard,uvdd1}.
  }%
  In this simulation, virtual \acp{UAV} can set their velocity and observe their neighbors with noisy measurements of their relative positions and relative orientations about the vertical axis.
  No blind spots are considered to exist for the relative localization sensor. If one \ac{UAV} is connected to another in a pre-defined observation graph, then it will successfully detect it in every measurement.
  Relative localization measurements are retrieved with a fixed rate.
  \green{%
    After each new measurement, every agent calculates the control action using eq. (\ref{eq:proportional_modded}).
    It then sets its velocity accordingly and maintains it until the next measurement is obtained.
    }%
  Thus, between two consecutive measurements, the agents follow a linear trajectory and rotate at a constant rate.
  \begin{figure}
    \includegraphics[trim={7.0cm 5.0cm 7.0cm 5.0cm},clip,height=0.22\linewidth]{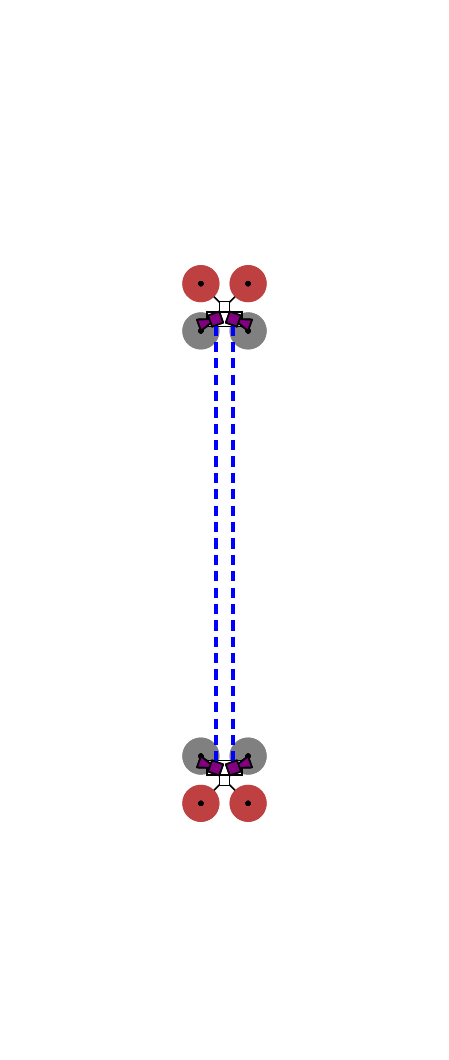}
    \hfill
    \includegraphics[trim={5.0cm 5.0cm 5.0cm 5.0cm},clip,height=0.22\linewidth]{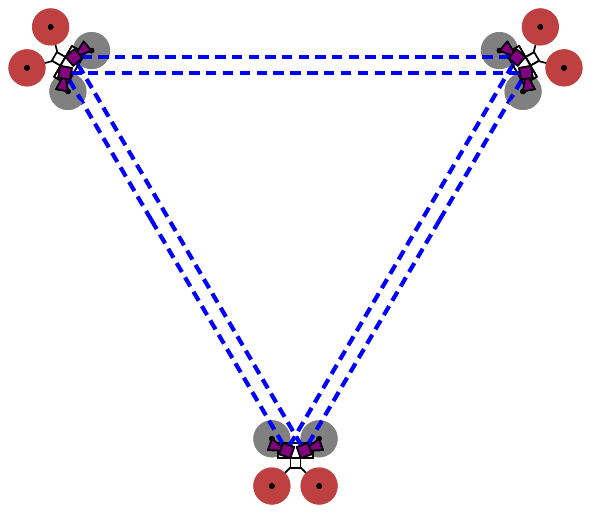}
    \hfill
    \includegraphics[trim={2.0cm 5.0cm 1.0cm 6.0cm},clip,height=0.22\linewidth]{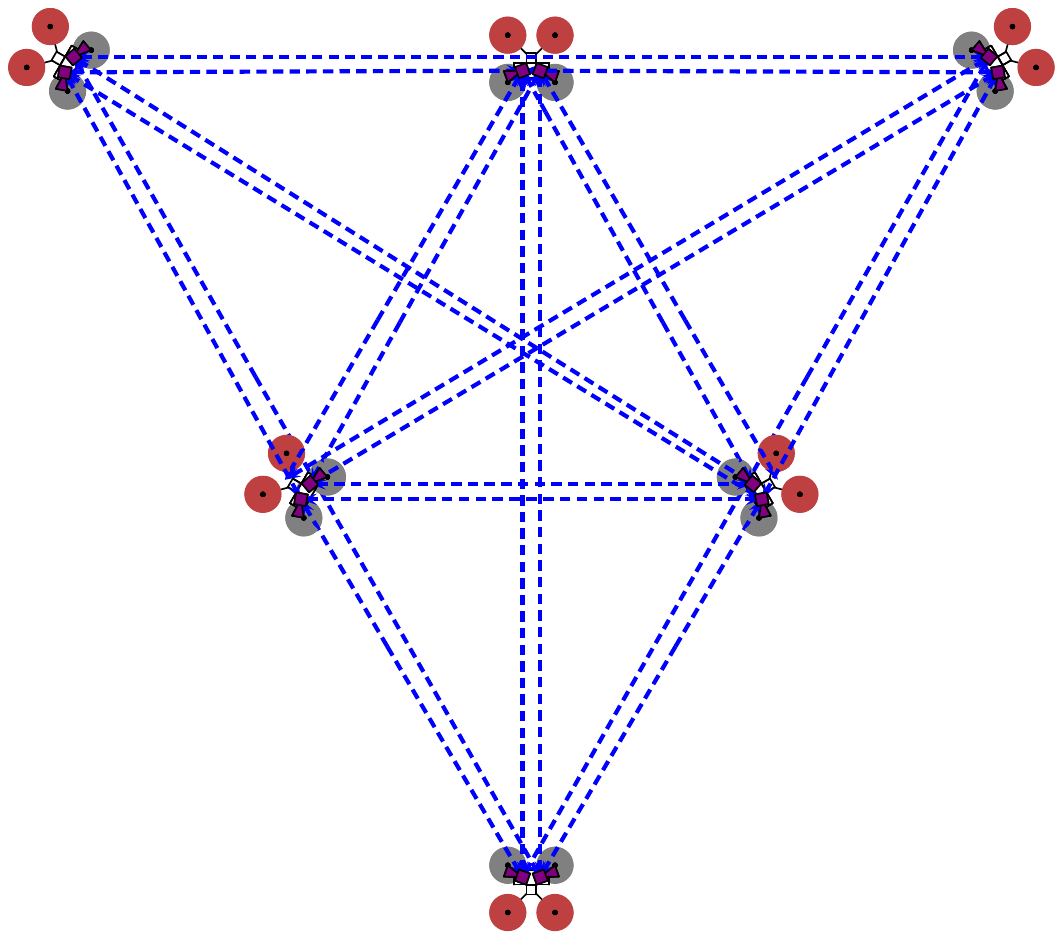}
    \hfill
    \includegraphics[trim={2.0cm 5.0cm 1.0cm 6.0cm},clip,height=0.22\linewidth]{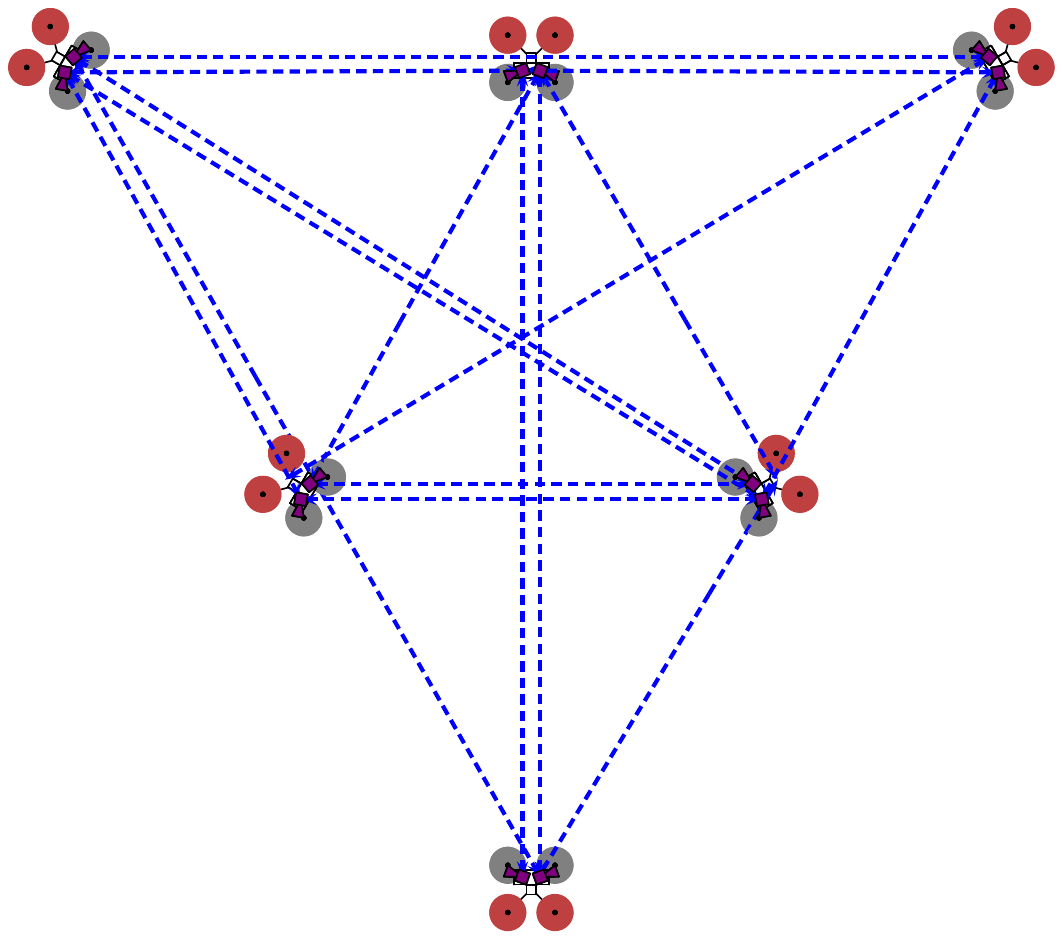}
    \vspace{-0.5em}
    \caption{The desired formations in our simulated experiments. The distance between the closest pair of agents is \SI{5}{\meter}. The latter three formations are based on equilateral triangles. Blue lines denote observations between agents.}
    \label{fig:desired_formations}
  \end{figure}
  To explore how various formations are affected by our technique, we have employed four distinct formation scenarios, as seen in Fig. \ref{fig:desired_formations}:
  \begin{itemize}
    \item{two mutually observing \acp{UAV},}
    \item{three mutually observing \acp{UAV} with a fully connected observation graph $\mathcal{G}$ and the desired formation in the shape of an equilateral triangle,}
    \item{six mutually observing \acp{UAV} with a fully connected observation graph $\mathcal{G}$ and the desired formation in the shape of a flat equilateral triangle, where three units represent its vertices and three units are in the center of each of its sides,}
    \item{six units with the same desired formation as above, with half of the edges in the observation graph $\mathcal{G}$ randomly removed such that the graph remains connected.}
  \end{itemize}

Each scenario was executed with four values of the parameter $\ell$: 0.05, 0.2, 0.35, and 0.5.
  When $\ell =$ 0.5, the control is equal to the pure proportional \ac{FEC}, according to the eq. (\ref{eq:proportional}).
  The proportional factor $k_e$ was set to 0.5.
  The simulated measurement noise was set to emulate the properties of a real vision-based relative measurement system.
  Specifically, each measurement was burdened with noise sampled from a Gaussian probability distribution, such that:
\begin{itemize}
  \item the distance component of the measured relative position had an additive noise with a standard deviation of \SI{10}{\percent} of the true distance;
  \item the bearing component had an additive noise corresponding to \SI{0.03}{\radian};
  \item the relative heading measurement noise had a standard deviation of \SI{0.26}{\radian}.
\end{itemize}

  Additionally, each scenario was tested with measurement rates $f$ from \SI{10}{\hertz} to \SI{200}{\hertz} in increments of \SI{10}{\hertz}.
  \green{%
    Because the movement between the measurements is simulated with a constant velocity, this change in rate can be alternatively interpreted as a change in the proportional factor $k_{ef}$ as shown in eq. (\ref{eq:clean_displacement}).
    The range of values of $k_{ef}$ for the above range of rates and chosen $k_e$ is from 0.05 to 0.0025.
    }%

  In every case, the agents were initialized with random orientations and positions up to \SI{20}{\meter} from the origin of the world coordinate frame.
  Every test was set to run for 2000 steps (corresponding to a duration of $\frac{2000}{f}$\SI{}{\second}) in order to ensure that transient effects have passed.
  Each of these tests was evaluated according to three separate criteria for position ($p$) and orientation ($\psi$):
  \begin{itemize}
    \item{Convergence times $t_{cp}$ and $t_{c\psi}$ approximating the time it took to reach an equilibrium.
      These values were obtained similarly to the definition in (\ref{eq:convergence_definition}) as:}
  \end{itemize}
\begin{equation}
  \begin{small}
    \begin{aligned}
    k_{cp} &= \min\!\left\{\!k \in 0..M\!:\!\sqrt{\sum_{i,j:c_{ij}=1}{\norm{\vect{p}_{ij}[k\!\!-\!\!1]-\vect{p}_{ij}^d}^2}} < 3\sigma_{\text{fin}p}[k]\right\}\\
      k_{c\psi} &= \min\!\left\{\!k \in 0..M\!:\!\sqrt{\sum_{i,j:c_{ij}=1}{\!\left(\psi_{ij}[k\!\!-\!\!1]-\psi_{ij}^d\right)^2}} < 3\sigma_{\text{fin}\psi}[k]\!\right\}\\
    t_{cp} &= k_{cp}/f \quad\quad\quad  t_{c\psi} = k_{c\psi}/f
  \end{aligned}
  \end{small}
\end{equation}
\quad\quad where
\begin{equation}
  \begin{aligned}
    \sigma_{\text{fin}p}[k] & = \sqrt{ \text{var}\left(\left\{\sum_{i.j:c_{ij}=1}{\norm{\vect{p}_{ij}[k']-\vect{p}_{ij}^d}^2} : k' \in k..M \right\}\right) }\\
    \sigma_{\text{fin}\psi}[k] & = \sqrt{ \text{var}\left(\left\{\sum_{i.j:c_{ij}=1}{\left(\psi_{ij}[k']-\psi_{ij}^d\right)^2} : k' \in k..M \right\}\right) }.
  \end{aligned}
\end{equation}
  \begin{itemize}
    \item{The \emph{stable-state} noise $\sigma_{tp}$ and $\sigma_{t\psi}$ of the formation obtained according to (\ref{eq:stable_state_sigma}):}
  \end{itemize}
  \begin{equation}
\begin{aligned}
  \sigma_{tp} &= \sigma_{\text{fin}p}[k_{cp}],\\
  \sigma_{t\psi} &= \sigma_{\text{fin}\psi}[k_{c\psi}].
\end{aligned}
  \end{equation}
  \begin{itemize}
    \item{Mean control-induced velocity and rotation rate change $\overline{\Delta v}$ and $\overline{\Delta\omega}$ obtained according to (\ref{eq:average_velocity}):}
  \end{itemize}
\begin{equation}
\begin{aligned}
  \overline{\Delta v} &= \frac{1}{N(M\!-\!2)}\sum_{k=2}^M\sum_{i=1}^N{\left( \norm{\vect{p}_{i}[k]\!\!-\!\!2\vect{p}_{i}[k\!\!-\!\!1]\!\!+\!\!\vect{p}_{i}[k\!\!-\!\!2]}f\right)},\\
  \overline{\Delta \omega} &= \frac{1}{N(M\!-\!2)}\sum_{k=2}^M\sum_{i=1}^N{\left( \left(\psi_{i}[k]\!\!-\!\!2\psi_{i}[k\!\!-\!\!1]\!\!+\!\!\psi_{i}[k\!\!-\!\!2]\right)f\right)}.
\end{aligned}
\end{equation}
      The results of these tests are shown in Fig.~\ref{fig:4D_exps}.
      Decreasing the maximal overshoot probability parameter $\ell$ generally decreases the mean velocity change $\overline{\Delta v}$ and mean rotation rate change $\overline{\Delta\omega}$, as well as the stable-state standard deviations $\sigma_{tp}$ and $\sigma_{t\psi}$.
      This comes at the cost of an increase in the convergence time $t_{cp}$ and $t_{c\psi}$ 

      It is also worth noting that if the measurements are obtained with a rate $f$ that is too low, the agents do not achieve convergence at all, which is evident particularly in cases with six agents.
      They instead oscillate randomly, much like the behavior of a swarm.
      \green{%
      This effect is seen in the plots as a very high stable-state errors and low convergence time at the low rate $f$.
      }%
      \green{%
      The convergence time is evaluated as low for these cases because the system rapidly reached a chaotic state whose high error and oscillations do not improve over time.
      }%
      The agents in this state move in large linear steps, making them deviate too far from the region where the local linearization of the gradient according to eq. (\ref{eq:gradient_descent}) is a reasonable approximation.

      \green{%
      Notably, the results from tests with six agents seen in Figures \ref{fig:4D_N6F} and \ref{fig:4D_N6P} show that convergence can be achieved at lower rates if the parameter $\ell$ is set to low values.
      This illustrates an additional advantage of our technique - beyond merely reducing oscillations, it enables convergent behavior for slower relative localization systems.
      }%

      The results also indicate that
      \begin{itemize}
        \item{increasing the number of agents in the formation increases the effect of degraded convergence, as seen in plots with more agents where the non-convergent behavior occurs with a higher $f$.%
          }
        \item{
            \green{%
      If the number of agents remains unchanged, the same is the case with more edges in the observation graph $\mathcal{G}$.
          A potential explanation is that with more agents, the error function $\vect{e}_F$ becomes increasingly non-linear \wrt{} to the local motion of an agent.}
          }%
        \item{
            \green{%
              In the theoretical noise-less case more edges in the observation graph lead to a more rigid formation.
              However, with noise included more edges lead to a more noisy formation control, which may initially appear counter-intuitive.
          }%
              Because the agents do not share their measurements, each agent is also negatively influenced by noisy measurements of the other agents' poses.
              Similarly to a sum of Gaussian random variables, the resulting output has a larger uncertainty than each of the contributing measurements.
    }
      \end{itemize}
      This also explains how the stable-state $\sigma_{tp}$ and $\sigma_{t\psi}$ seem to increase with an increasing number of agents or observations, even if $f$ is sufficiently high to avoid the aforementioned non-convergent behavior.

\section{Experimental verification}
\label{sec:experimental_verification}
\begin{figure}
    \includegraphics[trim={0cm 0cm 0cm 2cm},clip,height=0.3\linewidth]{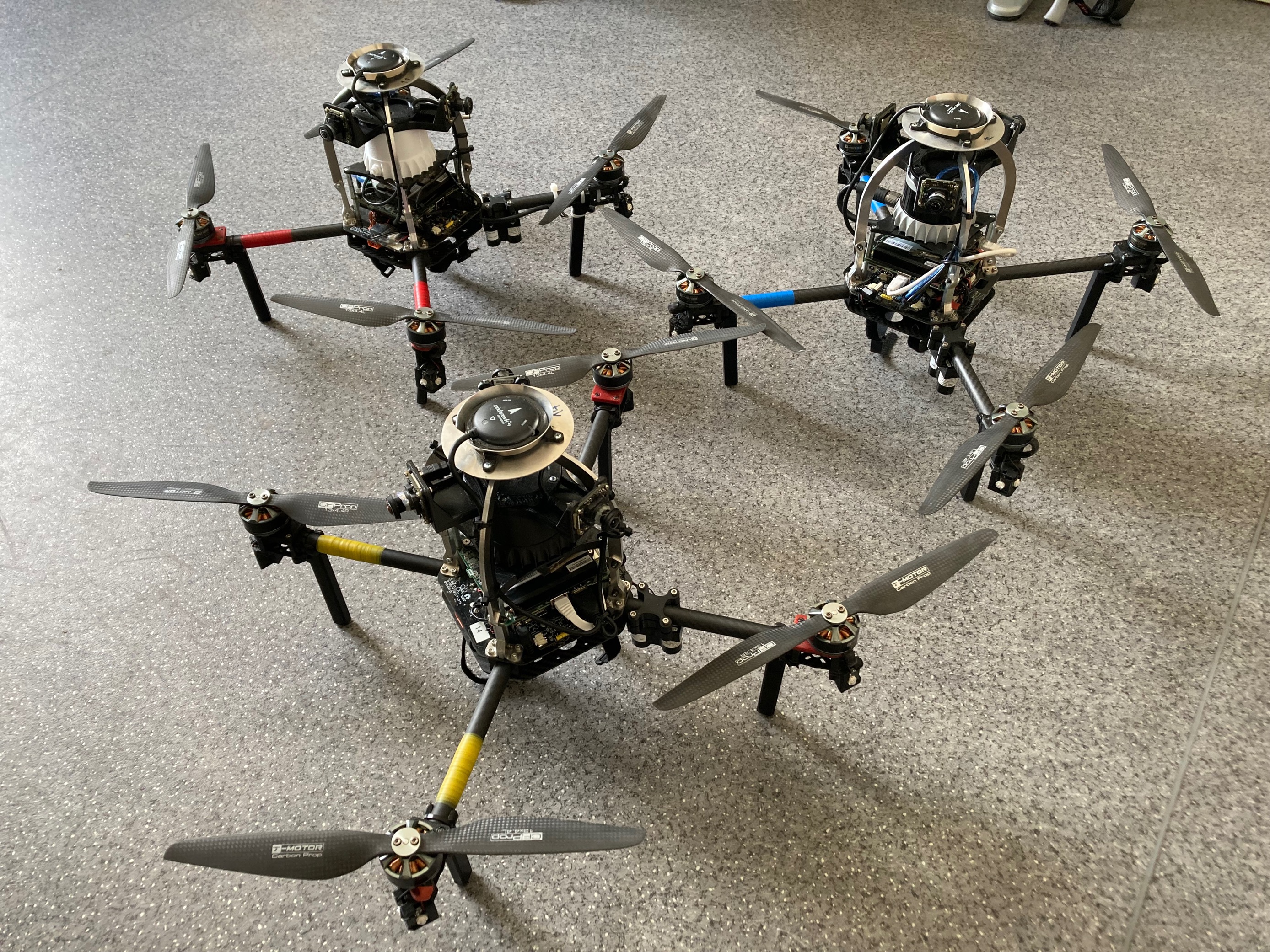}%
    \includegraphics[trim={61.3cm 23cm 63.3cm 43cm},clip,height=0.3\linewidth]{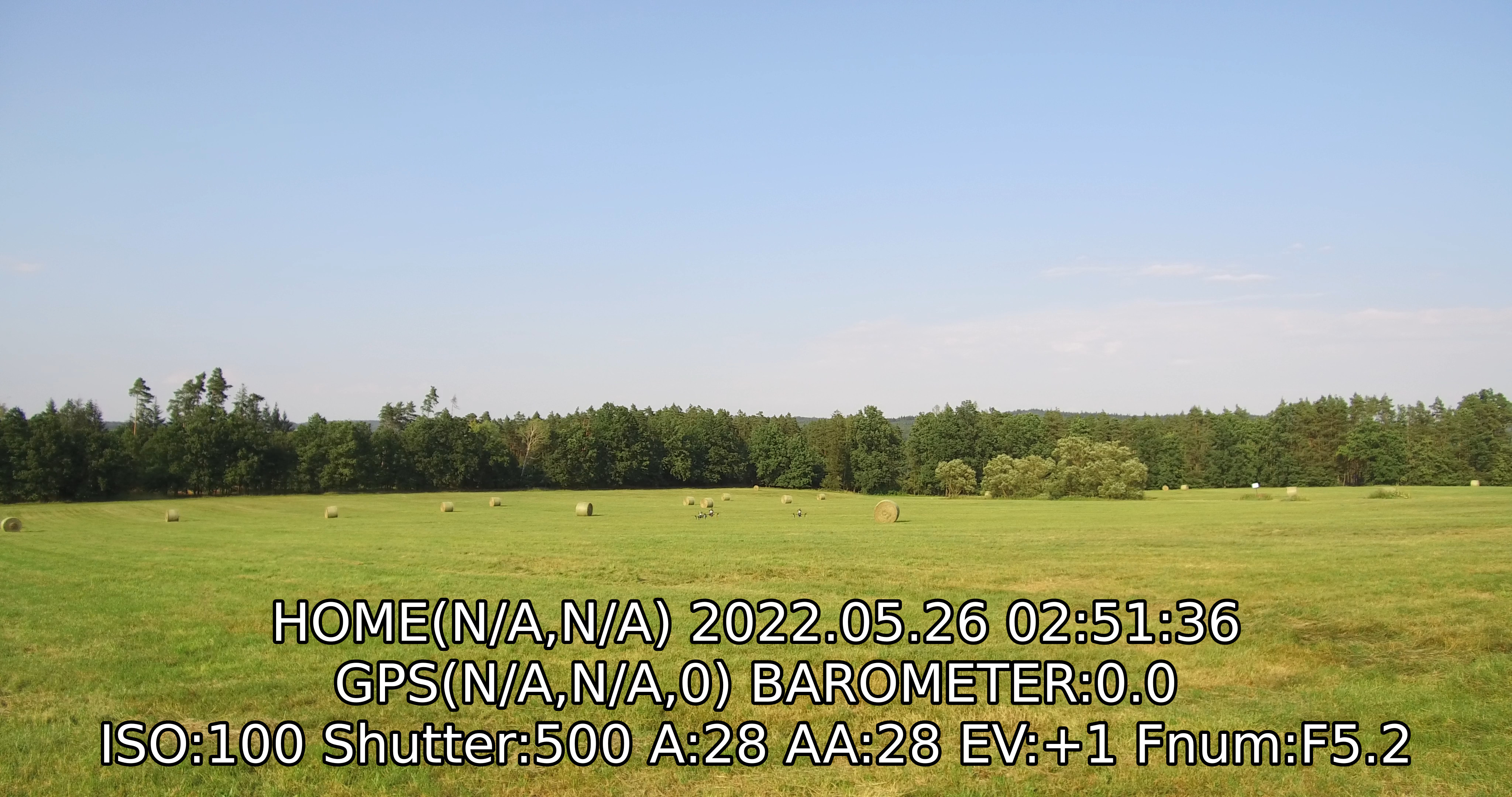}
    \caption{The three \ac{UAV} platforms used in the real-world experiments.}
    \label{fig:uav_platforms}
  \end{figure}

\begin{figure*}
  \includegraphics[trim={5.0cm 1.5cm 20.0cm 10.5cm},clip,width=0.246\linewidth]{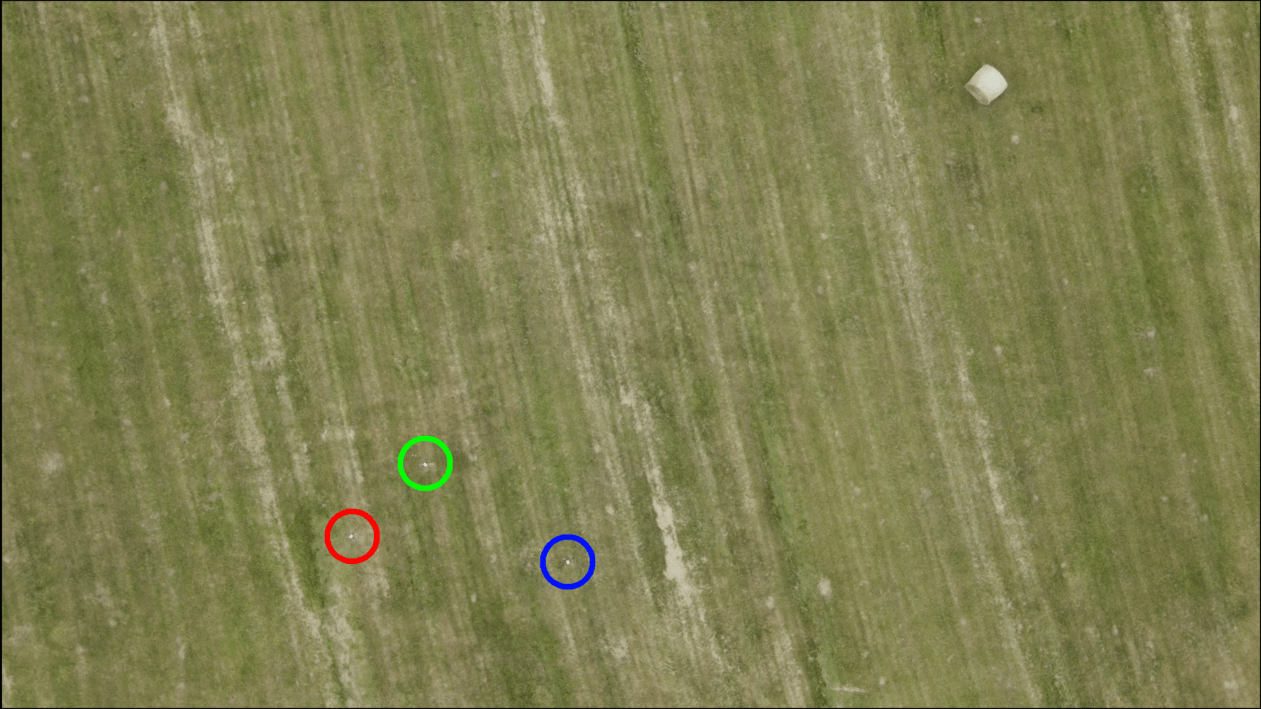}%
  \includegraphics[trim={5.0cm 1.5cm 20.0cm 10.5cm},clip,width=0.246\linewidth]{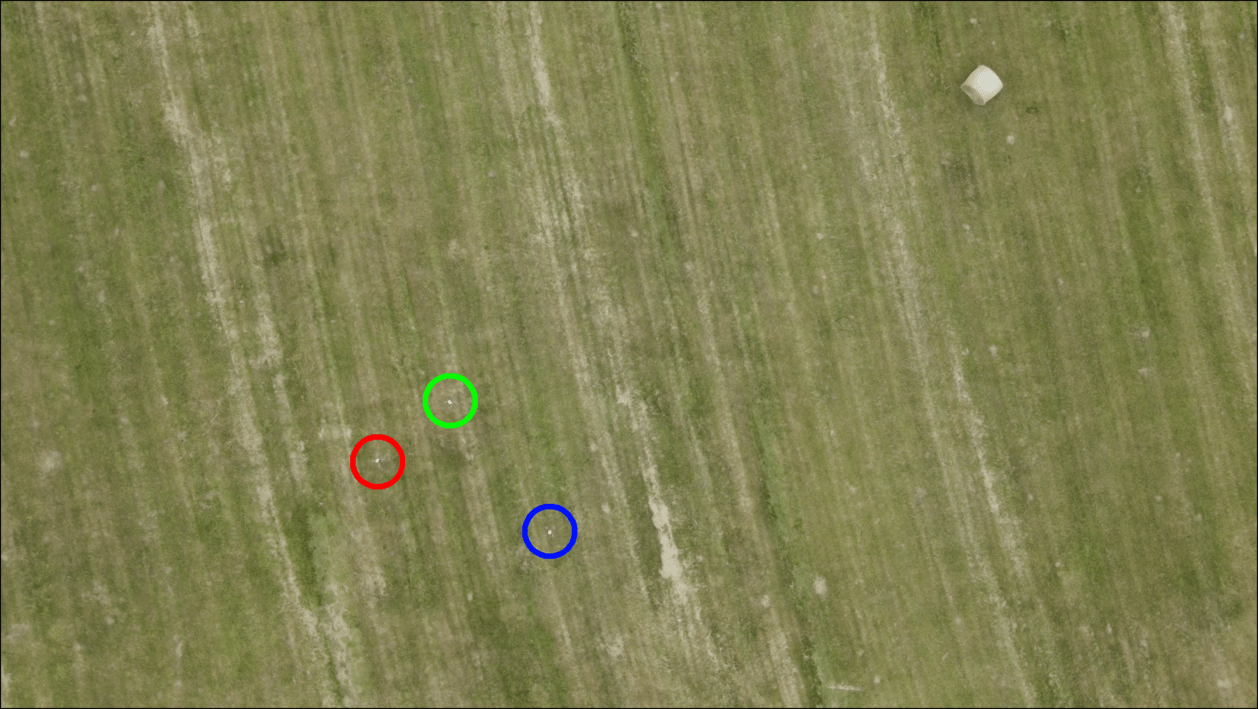}%
  \includegraphics[trim={13.5cm 4.0cm 11.5cm 8.0cm},clip,width=0.246\linewidth]{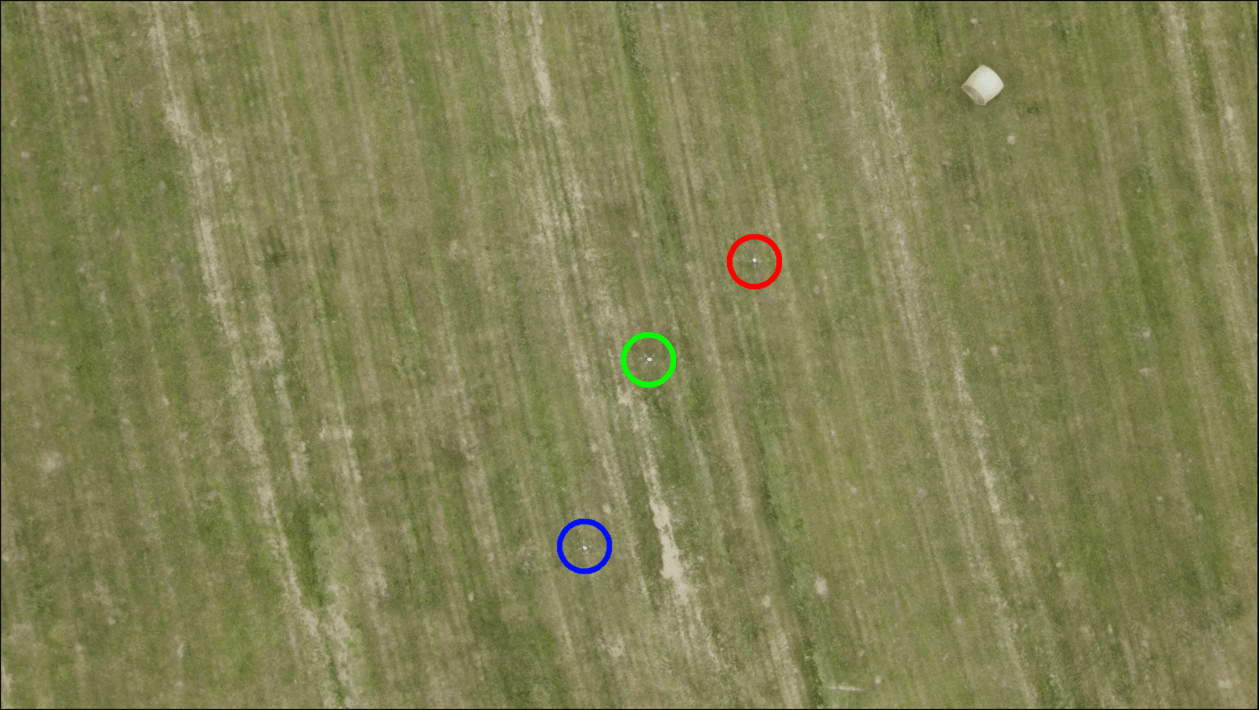}%
  \includegraphics[trim={18.0cm 0.05cm 7.0cm 12.0cm},clip,width=0.246\linewidth]{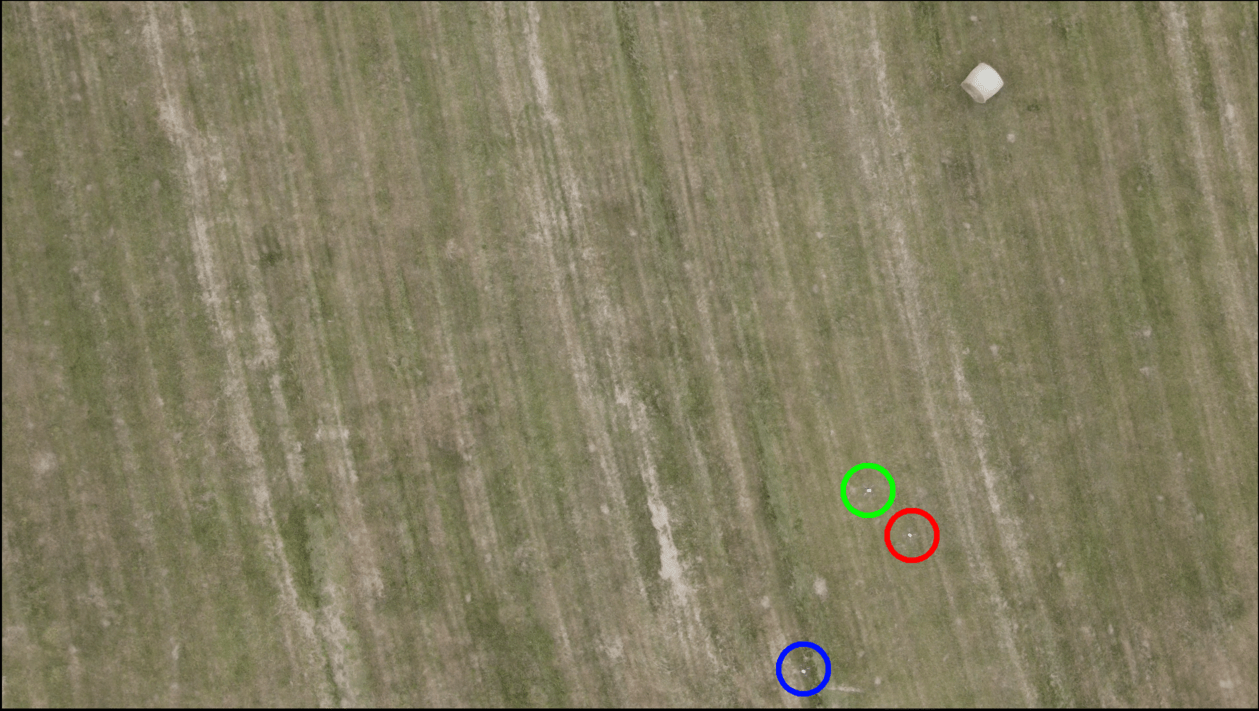}\\
  \includegraphics[trim={15.0cm 12.5cm 13.5cm 5.0cm},clip,width=0.246\linewidth]{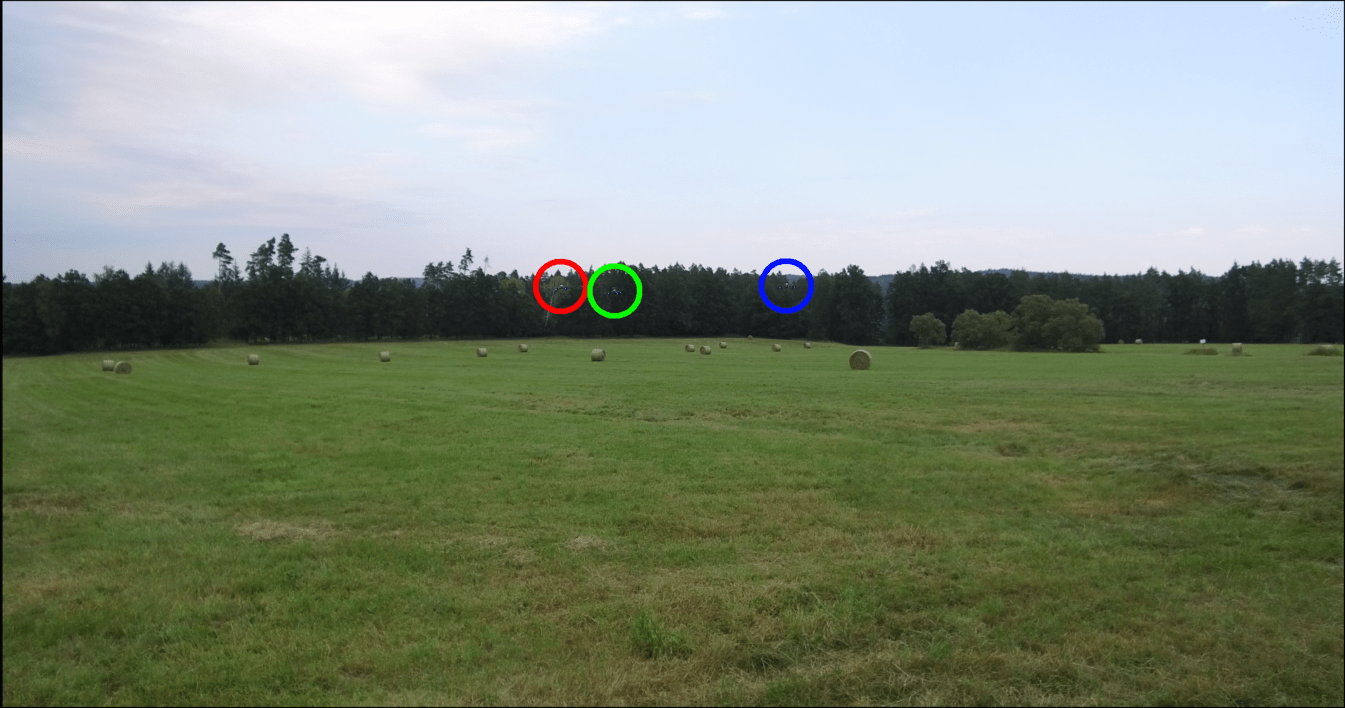}%
  \includegraphics[trim={15.0cm 12.5cm 13.5cm 5.0cm},clip,width=0.246\linewidth]{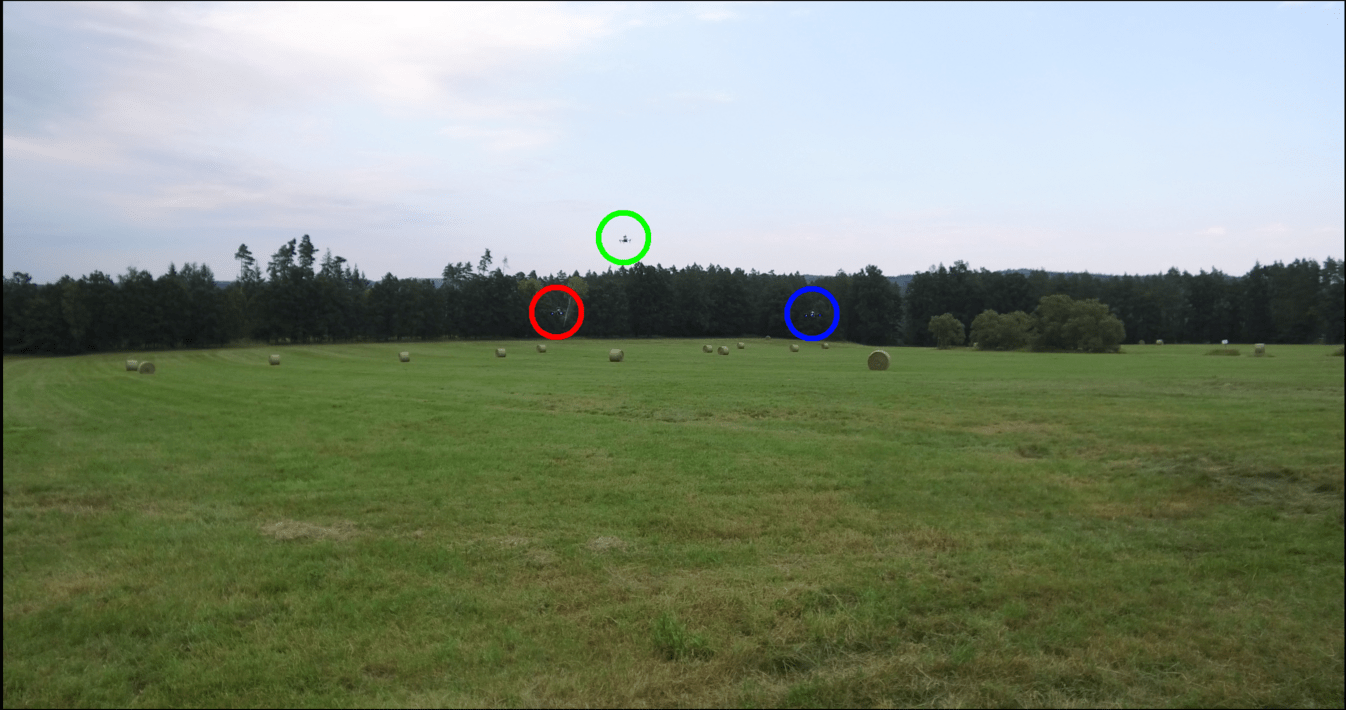}%
  \includegraphics[trim={15.0cm 12.5cm 13.5cm 5.0cm},clip,width=0.246\linewidth]{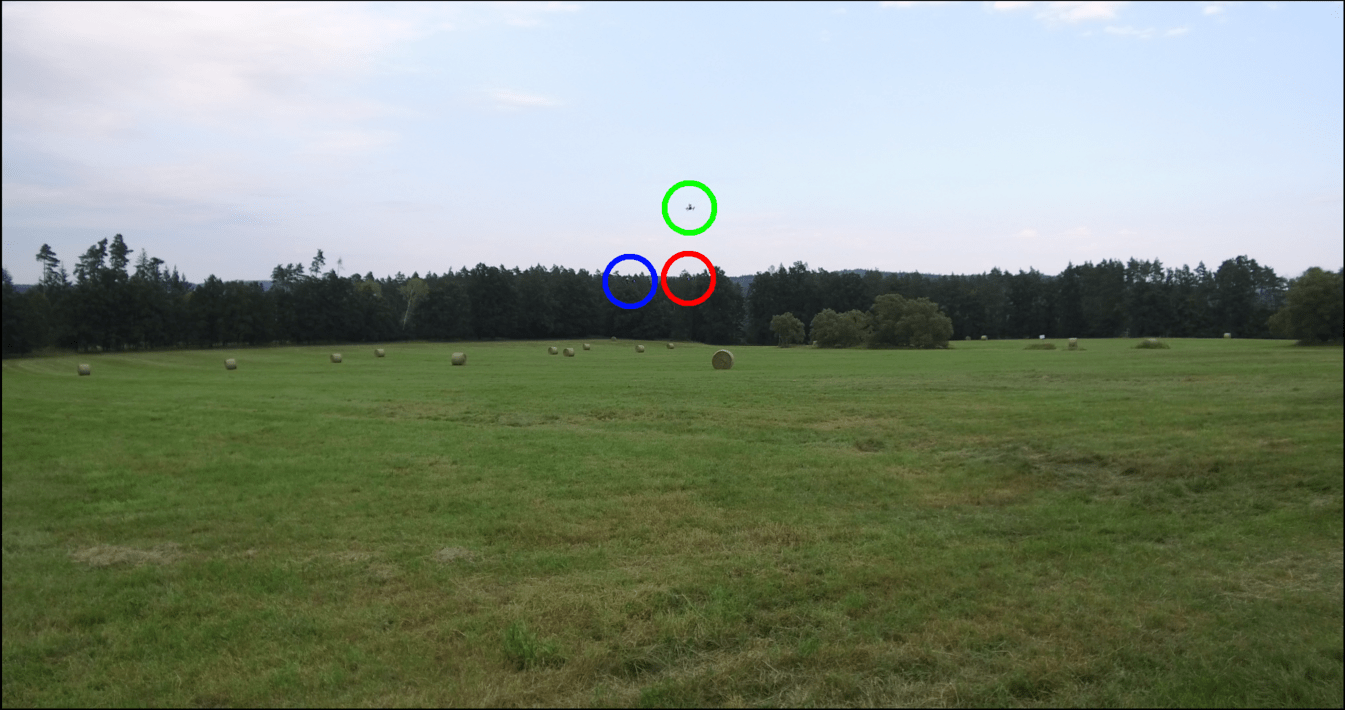}%
  \includegraphics[trim={15.0cm 12.5cm 13.5cm 5.0cm},clip,width=0.246\linewidth]{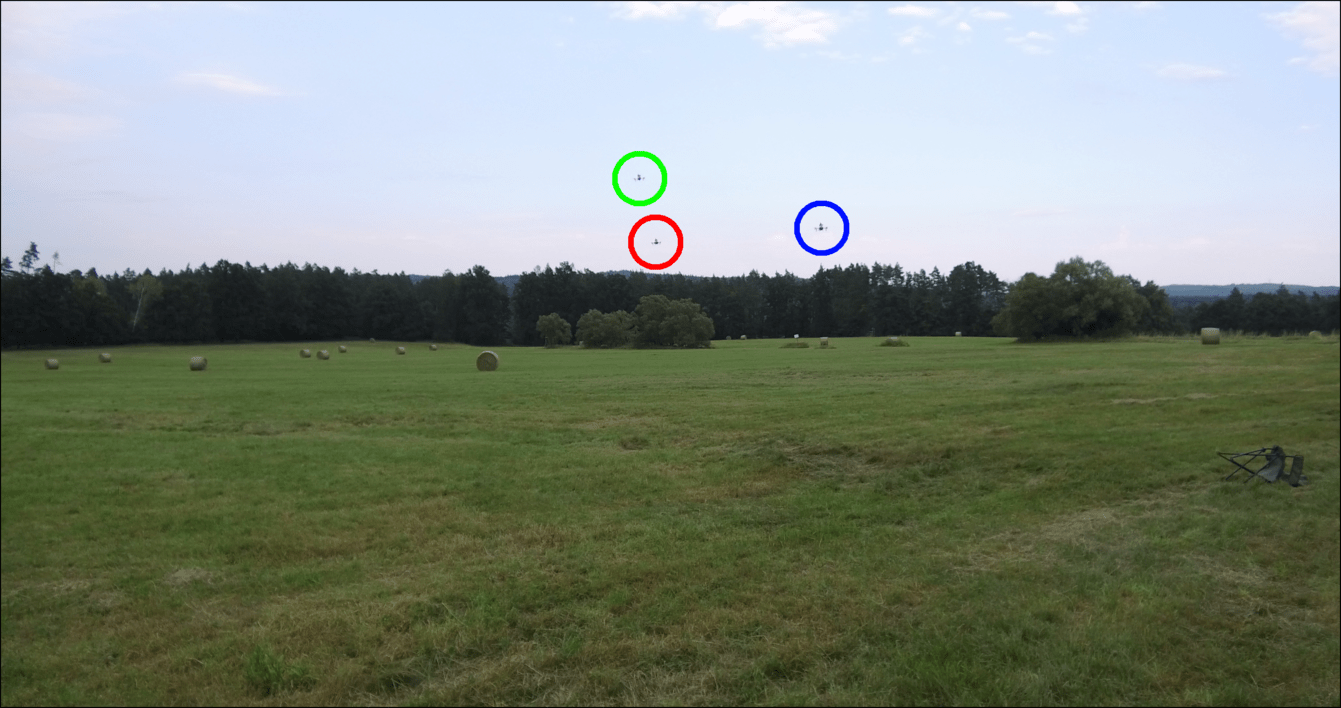}%
  \vspace{0.05cm}\\
  \includegraphics[trim={5.0cm 0.5cm 20.0cm 11.5cm},clip,width=0.246\linewidth]{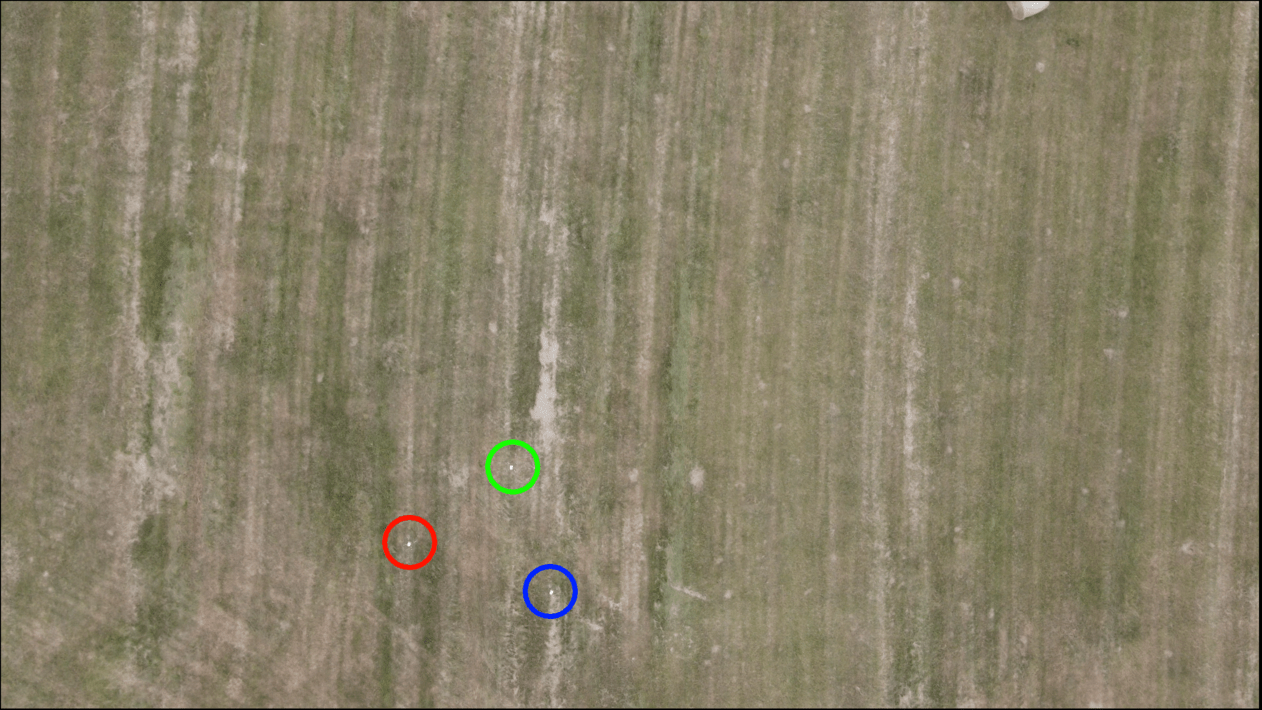}%
  \includegraphics[trim={5.0cm 0.5cm 20.0cm 11.5cm},clip,width=0.246\linewidth]{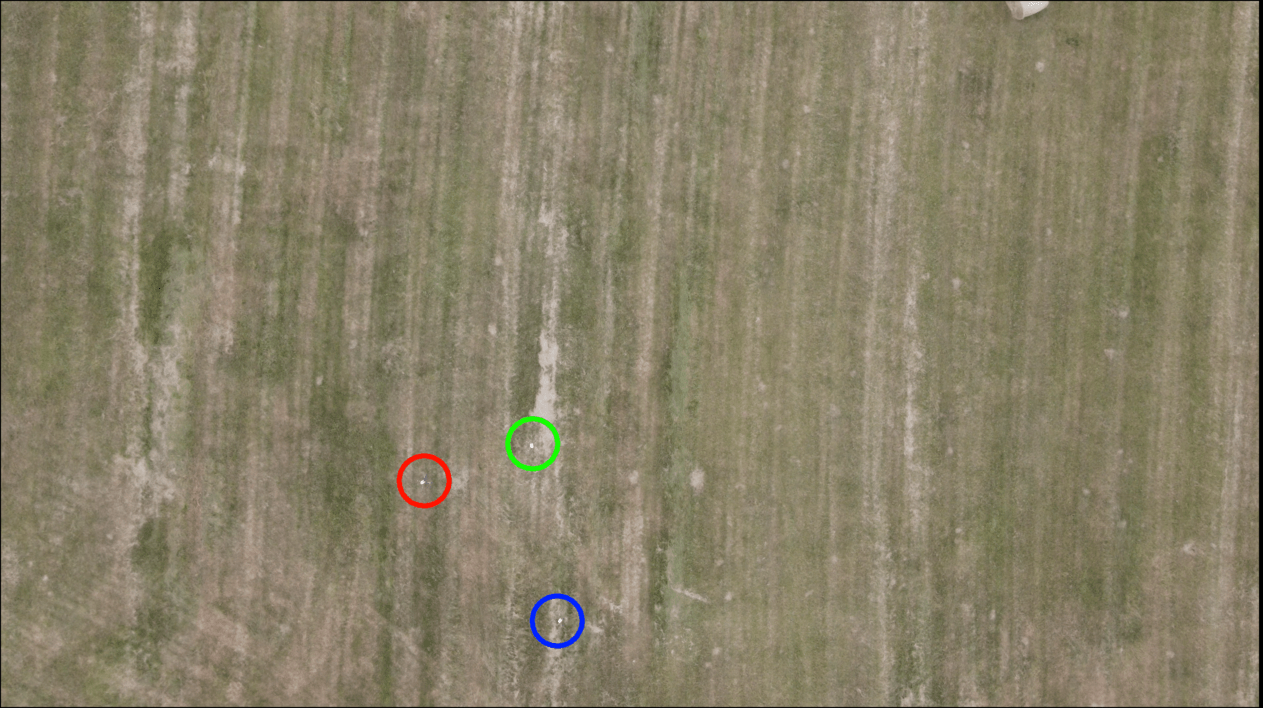}%
  \includegraphics[trim={5.0cm 0.05cm 20.0cm 12.0cm},clip,width=0.246\linewidth]{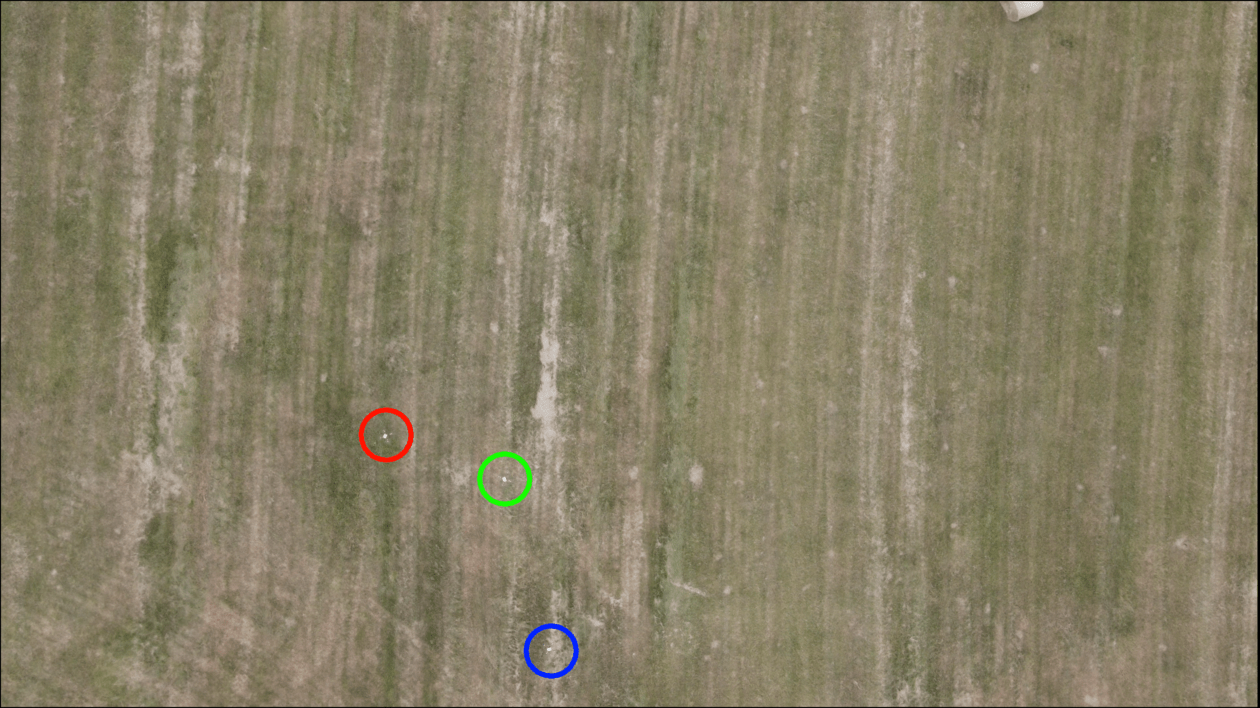}%
  \includegraphics[trim={5.0cm 1.5cm 20.0cm 10.5cm},clip,width=0.246\linewidth]{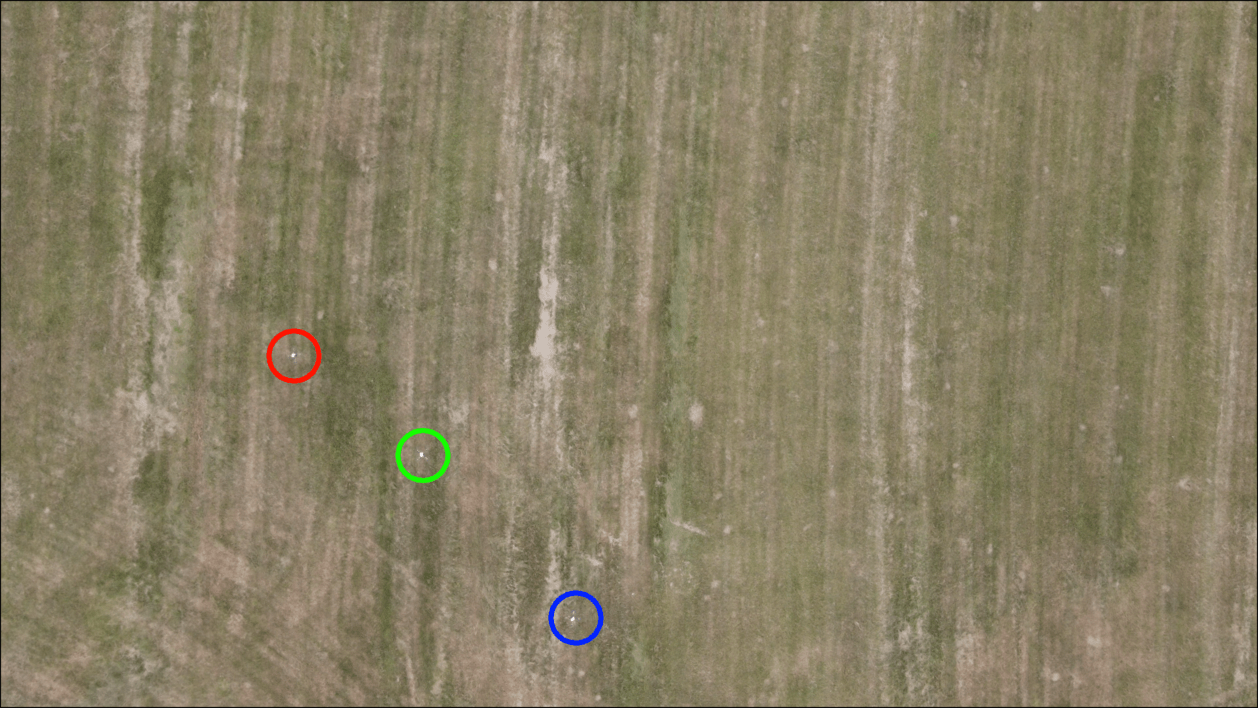}\\
  \includegraphics[trim={15.0cm 9.0cm 13.5cm 8.0cm},clip,width=0.246\linewidth]{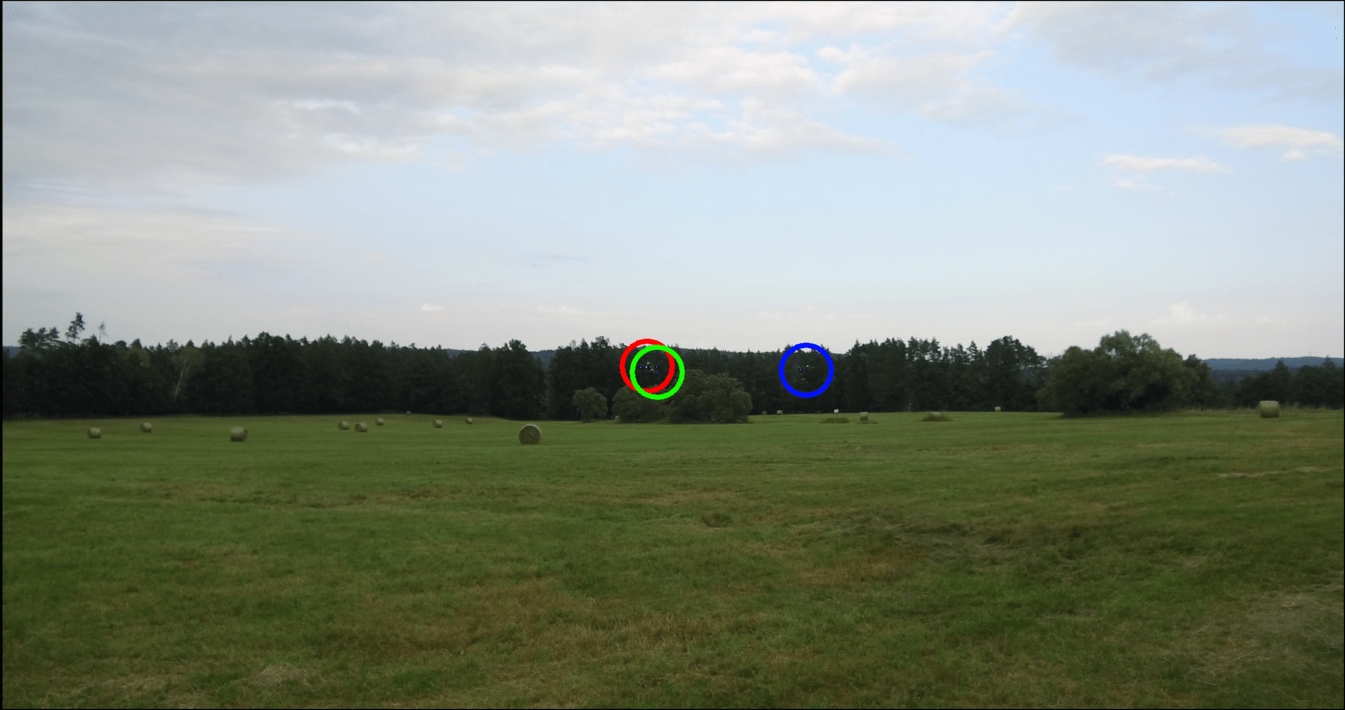}%
  \includegraphics[trim={15.0cm 9.0cm 13.5cm 8.0cm},clip,width=0.246\linewidth]{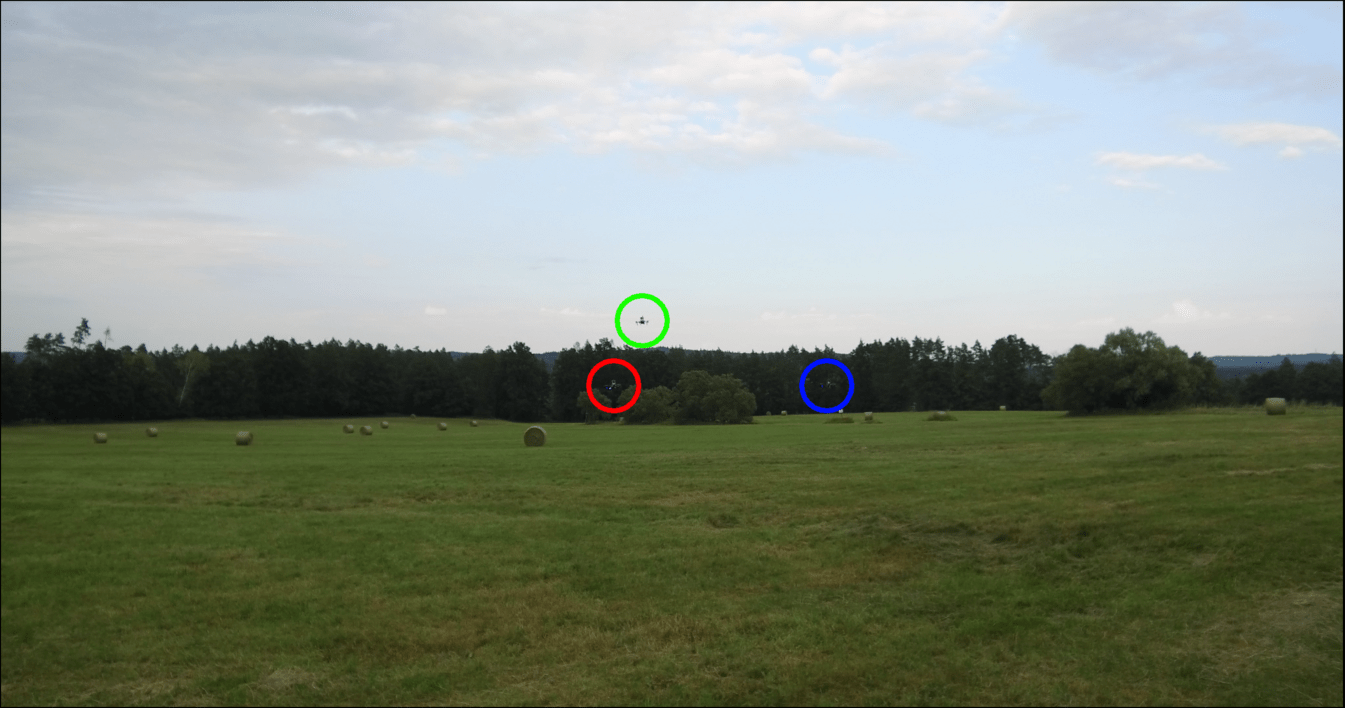}%
  \includegraphics[trim={15.0cm 9.0cm 13.5cm 8.0cm},clip,width=0.246\linewidth]{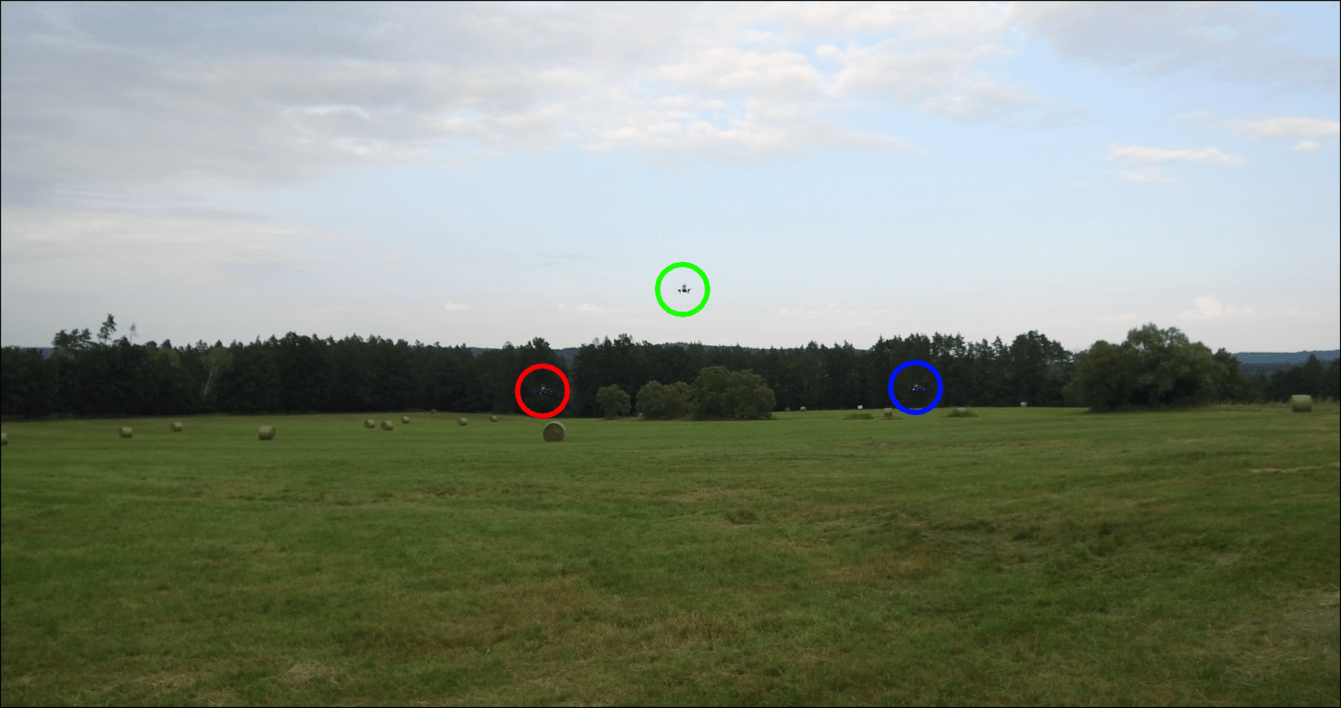}%
  \includegraphics[trim={15.0cm 9.0cm 13.5cm 8.0cm},clip,width=0.246\linewidth]{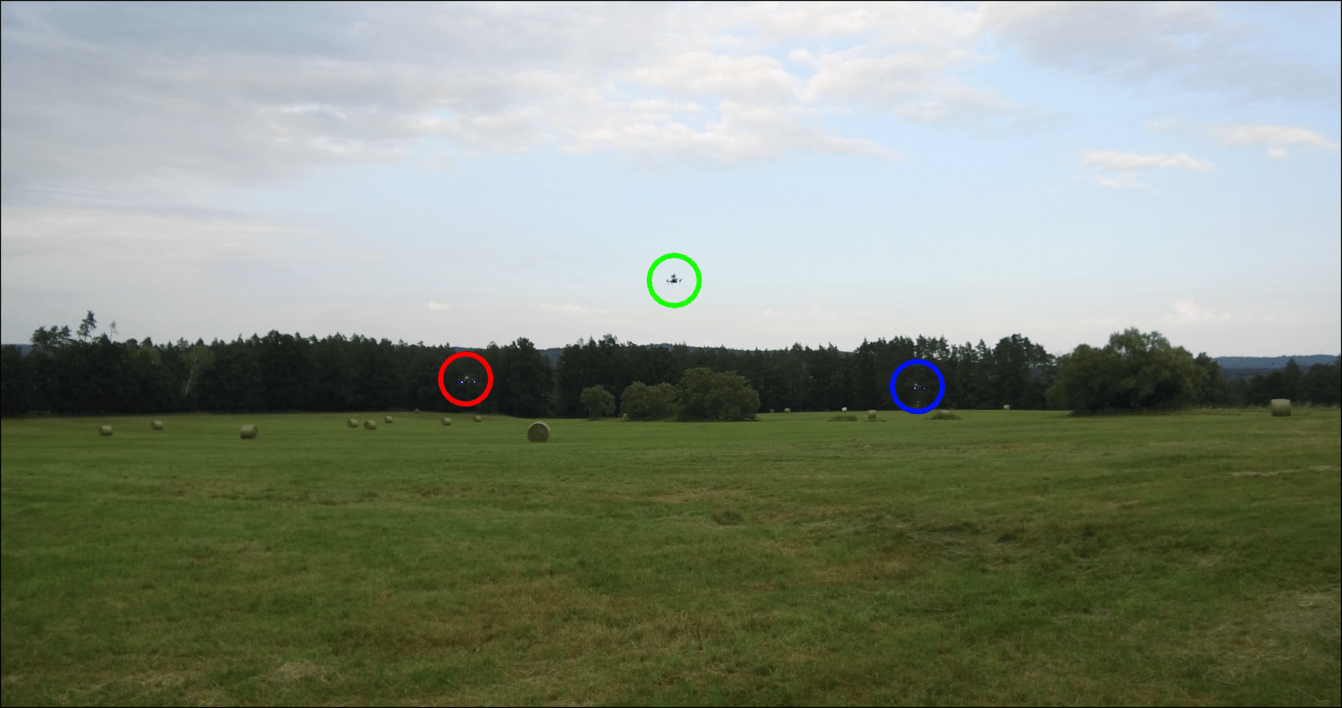}\\
    \hspace*{1.8cm} \colorbox{cyan}{A}\hfill \colorbox{magenta}{B}\hfill\colorbox{lime}{C}\hfill \colorbox{orange}{Cl}\hspace*{1.8cm}\\
  \vspace{-1.5em}
    \caption{Flight with a changing desired formation. 
    Photographs of the top and side view of the formation are from the time of the closest convergence of each case. 
    The top line of pictures presents flights without the proposed restraining technique, and flights with restraining are at the bottom.
    Without restraining, the formation significantly drifted and rotated in the third desired formation due to an accumulation of noise.
    It also took significantly longer for convergence to occur for this formation.
    The fourth desired formation was only achievable with our restraining technique applied. In the case without restraining, chaotic movement occurred and the achieved formation differed significantly from the desired one.}
    \label{fig:difecron_progress}
\end{figure*}

\green{%
  Since the proposed method is intended to enable real-world robotic deployment of distributed UAV formations, it was necessary to verify that our algorithm works outside of simulation.
  Thus, we have conducted a series of flights with three \acp{UAV} outdoors in an open space, as shown in Figure \ref{fig:difecron_progress}.
  }%
  Each unit was equipped with three \ac{UV}-sensitive cameras, and UV LED markers comprising the full \ac{UVDAR} system \cite{uvdar_ral}.
  These platforms are shown in Fig. \ref{fig:uav_platforms}.
  \green{%
    We have published the implementation of the proposed control technique used for the experiments on-line\footnote{\url{https://github.com/ctu-mrs/difec-ron}}.
  }%
  It is based on the \ac{ROS} and the \ac{MRS} \ac{UAV} system \cite{baca2021mrs} that provides velocity-tracking functionality.
  The \ac{MRS} \ac{UAV} system relies on a SE(3) geometric controller for low-level control of the \ac{UAV} and a \ac{MPC} trajectory tracker that takes the desired trajectory (or in this case, desired velocity) and pre-shapes it to provide a dynamically feasible full-state reference for the SE(3) controller to execute.
  \begin{figure*}
    \includegraphics[trim={1.8cm 11.0cm 1.5cm 11.2cm},clip,width=0.31\linewidth]{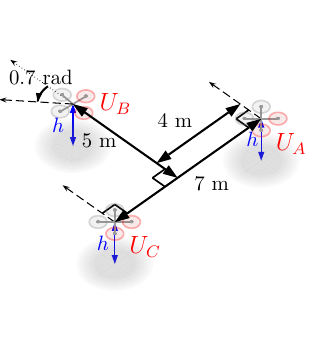}
    \hfill
    \includegraphics[trim={1.5cm 10.3cm 1.5cm 10.6cm},clip,width=0.31\linewidth]{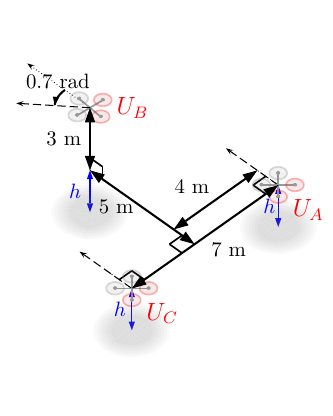}
    \hfill
    \includegraphics[trim={1.4cm 10.4cm 1.3cm 10.8cm},clip,width=0.33\linewidth]{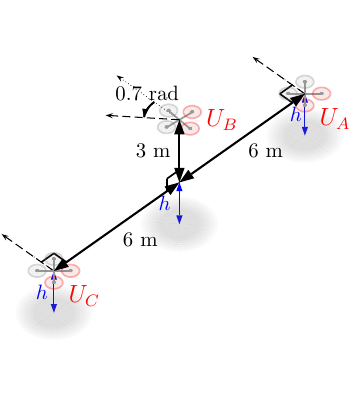}\\%
    \vskip -2.8em
    \hspace*{1cm} \colorbox{cyan}{A}\hfill \colorbox{magenta}{B}\hfill\colorbox{lime}{C} \hspace*{1.5cm}\\
    \vskip -2em
    \includegraphics[trim={0.0cm 11.0cm 0.0cm 11.2cm},clip,width=0.48\linewidth]{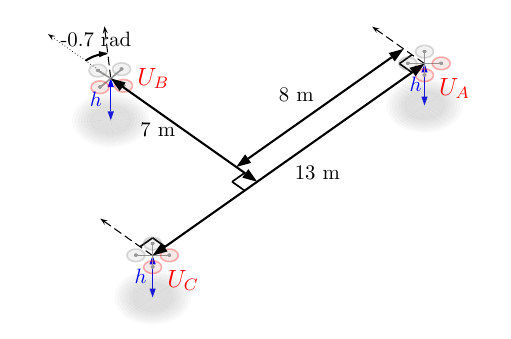}
    \hfill
    \includegraphics[trim={0.0cm 10.4cm 0.0cm 10.8cm},clip,width=0.48\linewidth]{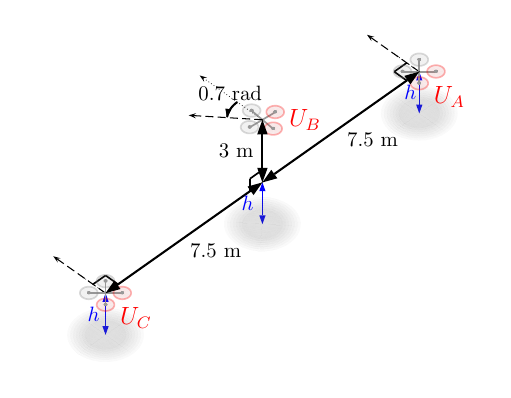}\\
    \vskip -2.0em%
    \hspace*{2.5cm} \colorbox{red}{Al}\hfill \colorbox{orange}{Cl}\hspace*{2.5cm}\\
   \vspace{-1.0em}%
    \caption{The desired formations used in our experimental verification. The red propellers denote the tail side of the \acp{UAV}. The larger formations Al and Cl contain mutual distances at which \ac{UVDAR}, at the used setting, can not reliably provide direct localization between \ac{UAV} {\color{red} {$U_A$}} and \ac{UAV} {\color{red} {$U_C$}}, such that the formation is \enquote{held together} through \ac{UAV} {\color{red} {$U_B$}}.}
    \label{fig:formations}%
  \end{figure*}

\green{%
  In the experiments, we alternated between a series of the desired formations seen in Fig. \ref{fig:formations}.
  These desired formations were communicated to all three \acp{UAV} using a wireless network.
  }%
  \green{%
  The \acp{UAV} then applied the presented algorithm using relative pose measurements from \ac{UVDAR}.
  }%

  In order to showcase the need for addressing the effects of sensory noise in \ac{FEC}, we have performed experiments with our modified action according to eq. (\ref{eq:proportional_modded}), and with the original action without restraining according to eq. (\ref{eq:proportional}).
  \green{%
  We have also performed experiments where we alternated between the two modes in order to provide a more direct comparison.
  }%
  \green{%
    For evaluation purposes the \acp{UAV} were equipped with \acs{GNSS}-\acs{RTK} system.
    This system provided us with a ground truth recorded for post-hoc analysis below, but we did not use it in the execution of the \ac{FEC} itself.
  }%
  \green{%
  The results of the experiments are shown in attached plots, with the following values of interest:
  }%
  \begin{itemize}
    \item{The norm $e_F$ of the error vector $\vect{e}_F$ from eq. (\ref{eq:error_function}), which was obtained using absolute positions.}
    \item{The \ac{UAV}-wise average norm $e_p$ of the relative position difference between individual current and desired relative positions of neighbors.}
    \item{The \ac{UAV}-wise average difference $e_{\psi}$ between individual current and desired relative orientations of neighbors.}
    \item{The \ac{UAV}-wise average angular velocity $v_{\psi}$. Higher values are detrimental for visual relative localization.}
    \item{The \ac{UAV}-wise average positional acceleration $a_{p}$. Higher values are detrimental for relative pose estimation.}
    \item{The connectivity of the observation graph $\mathcal{G}$, expressed through the Fiedler eigenvalue $f$ of the Laplace matrix $L$ of this graph.}
  \end{itemize}
  \green{%
    We will discuss the individual experimental runs below.
    }%
  \begin{figure*}[p]
    \includegraphics[width=\linewidth]{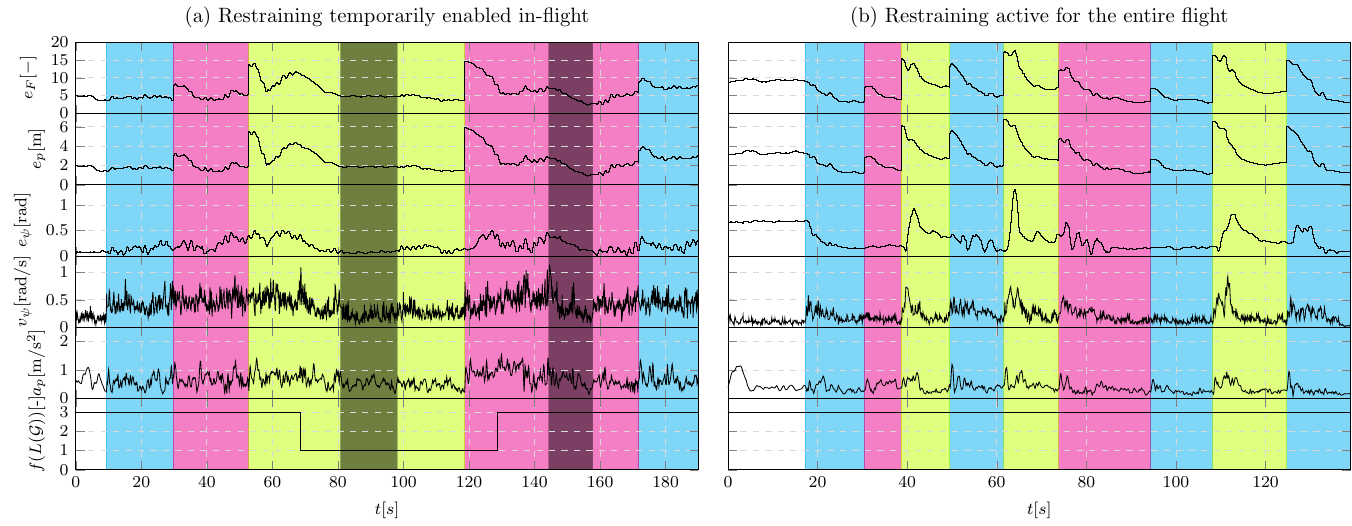}
\vspace{-2em}
    \caption{Plots of selected metrics during the flight of three \acp{UAV} where the desired formations were dynamically switched between formation A, formation B, and formation C.
    In the left plot, restraining was disabled for most of the flight and enabled for the intervals marked by the darker color.
    In the right plot, restraining was active during the entire flight time.
    Note, how enabling restraining rapidly improves performance of the control.
    }
    \label{fig:flight_first_plot}
  \end{figure*}
  \begin{figure*}[p]
  \includegraphics[width=\linewidth]{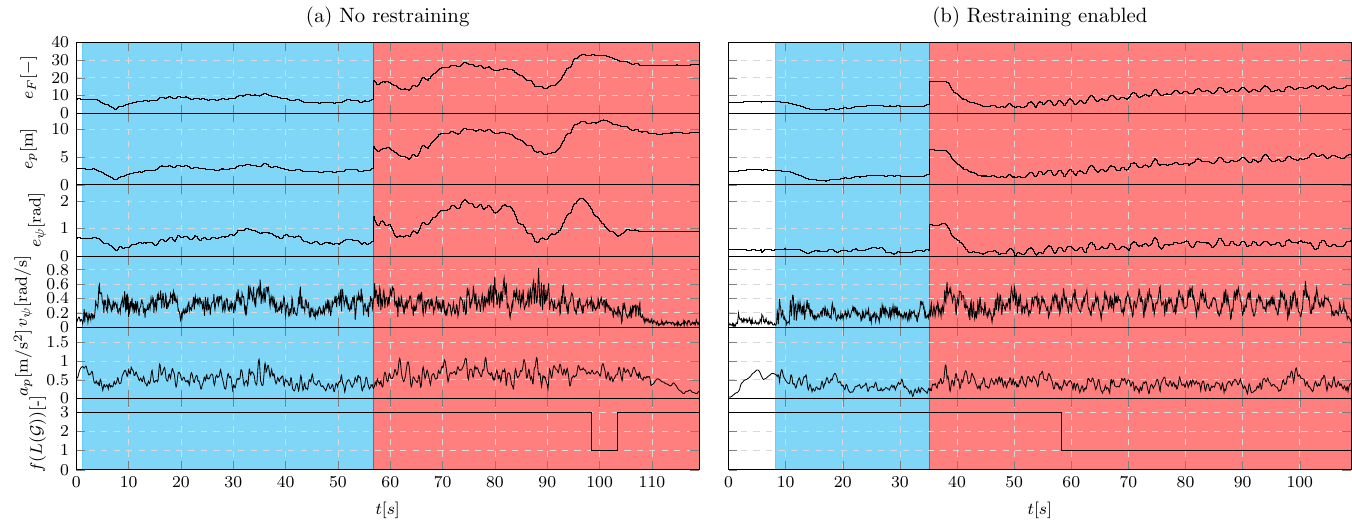}
\vspace{-2em}
    \caption{Plots of selected metrics during the flight of three \acp{UAV} where the desired formations were switched dynamically between formation A and the larger formation formation Al.
    Note that, without the proposed restraining technique, the formation rapidly loses adherence to the desired shape.
    }
    \label{fig:flight_temesvar_a_plot}
  \end{figure*}
  \begin{figure*}[p]
  \includegraphics[width=\linewidth]{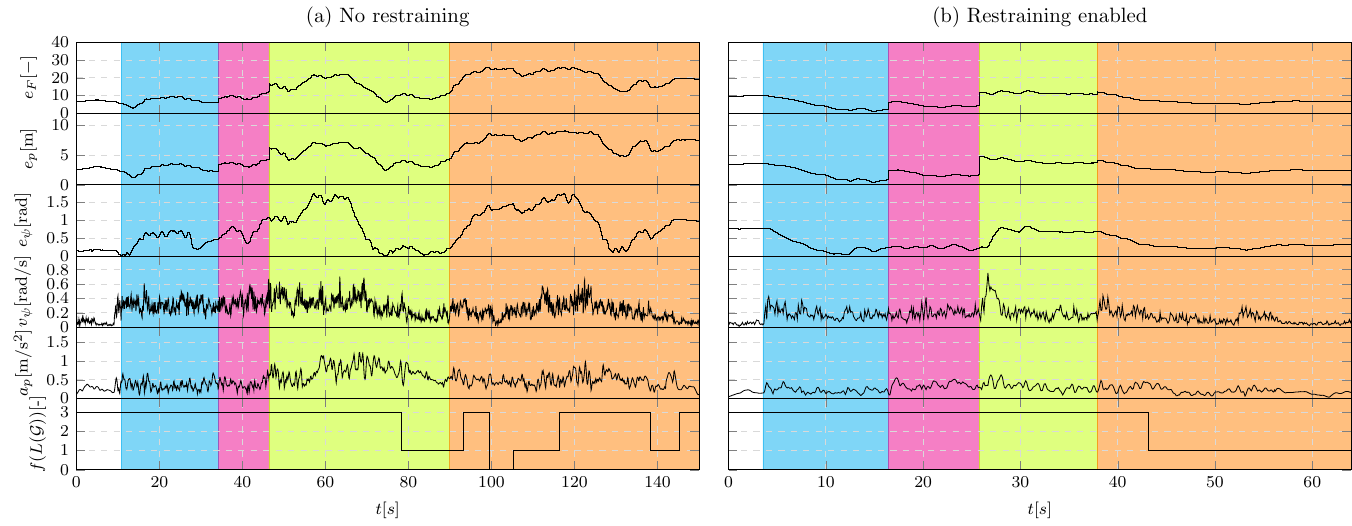}
\vspace{-2em}
    \caption{Plots of selected metrics during the flight of three \acp{UAV} where the desired formations were switched dynamically between formation A, formation B, formation C, and the larger formation Cl.
    Without the proposed restraining technique, the \acp{UAV} occasionally lost contact while following the largest desired formation Cl.
    This loss of contact occurred due to the \acp{UAV} reaching undesirably large distances from each other as a consequence of measurement noise and their individual disordered motion.
    The proposed restraining technique enabled formation preservation, even in these challenging sensory conditions.}
    \label{fig:flight_temesvar_c_plot}
\end{figure*}
  \subsection{Small formations}%
  In these experiments, we have alternated between formations A, B, and C, as shown in Fig. \ref{fig:formations}.
  Formation C has a relatively large distance between \acp{UAV} $U_A$ and $U_C$, leading to greater noise in estimation of both relative position and relative orientation.
  First, we have executed the sequence with restraining activated according to eq. (\ref{eq:proportional_modded}).
  The results of this test are shown in Fig. \ref{fig:flight_first_plot}b.
  The proportional factor $k_{e}$ was set to 0.06, the sensor sampling frequency to \SI{10}{\hertz}, and the threshold $\ell$ was set to 0.3 - a value we empirically determined to be an adequate compromise between convergence speed and restraining reaction to sensory noise.
  Each time a new desired formation was selected, the actual formation converged smoothly towards it.

  \green{%
  We have then performed this scenario with restraining disabled, while only activating at specific intervals.
  This was done to show that doing so stabilized the oscillating motion of the \acp{UAV}.
  }%
  The results of this experiment are shown in Fig. \ref{fig:flight_first_plot}a, with darker areas signifying where restraining was activated.
  The formation had a significantly worse rate of convergence to the desired formation, as well as exhibiting more unnecessary accelerations and rotations.
  In the sections where restraining was activated, this effect is visibly mitigated.
  
  \subsection{Large formations}
  \label{sec:large_formations}
  To test the real-world limits of the proposed algorithm, we have performed two sequences that end in the more extreme formations Al and Cl.
  Formation Al includes distances between the \acp{UAV} at which \ac{UVDAR} performs poorly, and its ability to estimate the relative orientation of the target shows significant noise.
  \green{%
    In formation Cl the distance between \acp{UAV} $U_A$ and $U_C$ is set to be at the limit of the \ac{UVDAR} operational range.
    This leads to a state where the two are only held in formation by mutual visibility with \ac{UAV} $U_B$.
    }%

  We have performed two scenarios, each with \ac{FEC} according to eq. (\ref{eq:proportional}) and with restraining according to eq. (\ref{eq:proportional_modded}), where the proportional factor $k_{e}$ was set to 0.06, the average sensor sampling frequency at \SI{10}{\hertz}, and the threshold $\ell$ set to 0.3.

  In the first scenario, we set the formation A as the desired formation. Once it converged, we set the desired formation to Al.
  The results of this experimental scenario are shown in Fig. \ref{fig:flight_temesvar_a_plot}.
  \green{%
    As is evident from the plots that without restraining, the system took significantly longer for convergence to the desired formation A.
    Additionally, it completely failed to converge to the desired formation Al due to the larger measurement noise at greater relative distances.
    }%

    \red{%
In the experiment with restraining, a slow but steady drift is visible in the formation region Al.
This behavior occurs because at this large separation distance the relative localization system operates near the limits of its effective range.
The \acp{UAV} $U_A$ and $U_C$ are linked together only through their mutual observations with \ac{UAV} $U_B$ and both neighbors observed \ac{UAV} $U_B$ with high measurement noise, particularly affecting the relative heading estimates.
The periodic oscillations superimposed on the same time period arise from the central \ac{UAV} $U_B$ alternating between minimizing the formation error relative to \ac{UAV} $U_A$ and \ac{UAV} $U_C$.
Despite these adverse effects, the overall flight characteristics are clearly improved compared to the case without restraining, exhibiting a significantly reduced divergence of the formation.
  }

  \green{%
  The second scenario included a sequence of desired formations A, B, C, and Cl, with and without restraining.
  }%
  The results are shown in Fig. \ref{fig:flight_temesvar_c_plot}.
  \green{%
  It is clearly visible in these plots that the accelerations and angular velocities were significantly reduced with restraining.
  Additionally, convergence to the desired formations was more precise in comparison with the non-restrained approach.
  }%
  \green{%
    Application of restraining enabled convergence to the desired formation even in the challenging cases where without restraining the agents failed to converge entirely.
  }%
  \green{%
    As evidenced by the Fiedler eigenvalue dropping to zero, the unrestrained \ac{FEC} even caused the observation graph to disconnect at one point.
  }%

  These experiments are shown in the video documentation\footnote{\url{http://mrs.felk.cvut.cz/difec-ron}}.
Results of the three experimental scenarios are summarized in Table \ref{tab:ExperimentalResults}, where we show the average rotation rates and acceleration for all three \acp{UAV} during each scenario.
\green{%
  The error change rate was averaged from the whole active \ac{FEC} period.
  We have excluded the instances of desired formation switching, since an abrupt change in the desired formation leads to sudden jump in the value of the formation error unrelated to the performance of the \ac{FEC} itself.
  }%

  \green{%
For the \emph{Small formations} flight without restraining, we have excluded the data where restraining was temporarily activated (seen as dark areas in Figure \ref{fig:flight_first_plot}a).
}%
\green{%
  These results show that the proposed technique improves \ac{FEC} both in terms of reduced tilting, which is beneficial for computer vision performance.
  The technique also improves the overall ability of a \ac{UAV} team to converge to a desired formation, especially in cases of significantly noisy relative pose measurements.
  }%

  \green{%
In the video documentation, the overall formation exhibits drifting over time.
This is expected, since the \ac{FEC} does not account for the positions of agents in the world.
}%
\begin{table}
  \centering
    \begin{tabular}{c|c|c|c|c} 
      \multirow[c]{2}{*}[0em]{Flight}&\multirow[c]{2}{*}[0em]{Value}& Without & With & \multirow[c]{2}{*}[0em]{Improv.}\\ 
      && restraining & restraining & \\ 
      \hline     
      Small&
      $\overline{v_\psi}\left[\nicefrac{\mathrm{rad}}{\mathrm{s}}\right]$&$0.4096$& $0.2056$ & $49.8\%$\\
      formations&
      $\overline{a_p}\left[\nicefrac{\mathrm{m}}{\mathrm{s}^2}\right]$&$0.6403$& $0.3558$ & $44.4\%$ \\ 
      \tiny{{\colorbox{cyan}{A}}{\colorbox{magenta}{B}}{\colorbox{lime}{C}}}&
      $\overline{\nicefrac{d e_p}{dt}}\left[\nicefrac{\mathrm{m}}{\mathrm{s}}\right]$&$-0.0420$& $-0.2085$ & $0.167\nicefrac{\mathrm{m}}{\mathrm{s}}$ \\ 
      \hline
      Large&
      $\overline{v_\psi}\left[\nicefrac{\mathrm{rad}}{\mathrm{s}}\right]$&$0.2769$& $0.2686$ & $3.0\%$\\
      formations 1&
      $\overline{a_p}\left[\nicefrac{\mathrm{m}}{\mathrm{s}^2}\right]$&$0.5142$ & $0.3759$ & $26.9\%$ \\ 
      \tiny{{\colorbox{cyan}{A}}{\colorbox{red}{Al}}}&
      $\overline{\nicefrac{d e_p}{dt}}\left[\nicefrac{\mathrm{m}}{\mathrm{s}}\right]$&$0.0146$& $-0.0190$ & $0.034\nicefrac{\mathrm{m}}{\mathrm{s}}$ \\ 
      \hline
      Large&
      $\overline{v_\psi}\left[\nicefrac{\mathrm{rad}}{\mathrm{s}}\right]$&$0.2579$& $0.1571$ & $39.1\%$\\
      formations 2&
      $\overline{a_p}\left[\nicefrac{\mathrm{m}}{\mathrm{s}^2}\right]$&$0.4965$ & $0.2436$ & $50.9\%$\\ 
      \tiny{{\colorbox{cyan}{A}}{\colorbox{magenta}{B}}{\colorbox{lime}{C}}{\colorbox{orange}{Cl}}}&
      $\overline{\nicefrac{d e_p}{dt}}\left[\nicefrac{\mathrm{m}}{\mathrm{s}}\right]$&$0.0144$& $-0.0737$ & $0.088\nicefrac{\mathrm{m}}{\mathrm{s}}$ \\ 
      \hline
    \end{tabular}
    \caption{
      Evaluation of the three real-world experiments.
      Average angular velocities $\overline{v_\psi}$ and positional accelerations $\overline{a_p}$, as well as the rate of change in average positional formation error $\overline{\nicefrac{d e_p}{dt}}$, are shown.
      For all three variables, lower values are better.
      The average angular velocity and acceleration show the amount of oscillations in the system, where a higher value leads to a degradation of the onboard sensing performance.
    The results demonstrate that the presented technique significantly reduces such tilting and accelerations.
    It also improved the convergence to the desired formation, as seen by the lower error change rate.
    \label{tab:ExperimentalResults}}
\end{table}
\section{Conclusion}
\green{%
In this work, we present a novel technique to allow real-world deployment of the classical \acf{DIFEC} based on relative neighbor observation.
}%
The proposed technique, called \emph{restraining}, addresses the problems inherent in such formation control if the relative measurements by onboard sensors are obtained in discrete time-steps, and are subject to noise with a known distribution.
Since most of the available onboard relative localization sensors provide discrete data with known characteristics, this technique has a wide range of applicability.
\green{%
  Our proposal involves the design the control action, such that we exploit the knowledge of the statistical properties of the observation noise.
  }%

  \green{%
The user of a multi-robot system utilizing this technique can tune the behavior to his needs by specify the value of a statistical parameter $\ell$.
The value of this parameter influences the trade-off between the smoothness of flight and ability to converge to formation on one side and the speed and aggressiveness of the agents on the other.
}%

\green{%
We have applied our technique to formations based on the relative pose observations consisting of the combined 3D positions and headings of \acp{UAV}.
}%

\green{%
  However, the general statistical principle of the \emph{restraining} technique can also be applied to other situations, such as bearing-based or distance-based formations, and even to tasks beyond formation enforcement.
  }%

We have verified the performance of the presented technique both in simulation and in real-world experiments with the use of the \ac{UVDAR} sensor.
All the various experiments demonstrated significant improvements in flight characteristics - see Table \ref{tab:ExperimentalResults} - and also numerous situations where formation flight fails without the proposed technique.
\appendix
\addcontentsline{toc}{section}{Appendices}
\counterwithin{lemma}{section}
\setcounter{lemma}{0}
\counterwithin{equation}{section}
\renewcommand\theequation{\arabic{equation}}
\counterwithin{table}{section}
\renewcommand\thetable{\Roman{table}}
\section{Formation enforcing action in $\mathbb{R}^3\times\mat{S}^1$}
\label{App:fec}
We will derive a control action that moves the current formation down the gradient of the overall error \wrt{} a given desired formation.
As established in section \ref{sec:fec}, such control can be expressed as
\begin{equation}
\label{App:fec:eq:1}
  \dot{\vect{q}} = k_e{\left(\frac{\partial \varkappa_\mathcal{G}\left(\vect{q}\right)}{\partial \vect{q}}\right)}^T\vect{e}_F.
\end{equation}
From this form, we shall derive control equations usable in a group of \acp{UAV} that do not explicitly communicate their poses and individually rely only on their observations.
  Following similar steps to those used to obtain a bearing-based \ac{FEC} in \cite{rigiditybased}, we define the \emph{relative pose rigidity matrix} $\mat{H}_{\mathcal{G}}^W$:
  \begin{equation}
  \label{App:fec:eq:2}
    \begin{aligned}
    \mat{H}_{\mathcal{G}}^W\left(\vect{q}\right) \triangleq \frac{\partial \varkappa_\mathcal{G}\left(\vect{q}\right)}{\partial\vect{q}}.
    \end{aligned}
  \end{equation}
  This matrix expresses how much the formation $\varkappa_\mathcal{G}\left(\vect{q}\right)$ changes with the agent poses $\vect{q}$.
  Thereafter, the four lines of $\mat{H}_\mathcal{G}^W$ corresponding to the edge $e = \left(i,j\right)$ are
\newcommand\bundermat[2]{%
  \makebox[0pt][l]{$\smash{\underbrace{\phantom{%
    \mathmakebox[\widthof{$\begin{smallmatrix}\!\!#2\!\!\end{smallmatrix}$}][l]{
        \begin{smallmatrix}\frac{1}{1}\\\frac{1}{1}\\\frac{1}{1}\end{smallmatrix}}
        }}_{\text{#1}}}$}#2}%
\newcommand\zerodash{%
  ...0...}%
  \begin{equation}
\mat{H}_{\mathcal{G}ij}^W\left(\vect{q}\right) =%
      \begin{bsmallmatrix}
        \zerodash & \bundermat{$\nicefrac{\partial \varkappa_\mathcal{G}}{\partial \vect{p}_{i}}$}{-{\mat{R}(\psi_i)}^T} & \bundermat{$\nicefrac{\partial \varkappa_\mathcal{G}}{\partial \psi_{i}}$}{{\frac{\partial \mat{R}(\psi_i)}{\partial \psi_i}}^T\vect{p}_{ij}^W} &  \zerodash & \bundermat{$\nicefrac{\partial \varkappa_\mathcal{G}}{\partial \vect{p}_{j}}$}{\mat{R}(\psi_{i})^T} & \bundermat{$\nicefrac{\partial \varkappa_\mathcal{G}}{\partial \psi_{j}}$}{\hspace*{1.3em}\mathbf{0}\hspace*{1.3em}} & \zerodash
        \\
        \zerodash & {0} & -1 & \zerodash & 0 & 1 & \zerodash
      \end{bsmallmatrix}.
    \label{eq:rigidity_matrix_world}
  \vspace{1.0em}
  \end{equation}
  This expression can be transformed into the local frame of the agent $i$, which is then observing agents $j$ through relative pose measurements.
    This transformation yields the matrix $\mat{H}_{\mathcal{G}}^l$ and comprises of per-edge sub-matrices as follows:
  \begin{equation}
    \mat{H}_{\mathcal{G}ij}^l\left(\varkappa_\mathcal{G}\right) =%
    \begin{bmatrix}
      \zerodash & -\mat{I}_3 & \mat{S}^T \vect{p}_{ij} & \zerodash & -\mat{R}(\psi_{ij}) & \mathbf{0} & \zerodash%
      \\%
      \zerodash & 0 & -1 & \zerodash & 0 & 1 & \zerodash%
    \end{bmatrix}%
    \label{eq:rigidity_matrix_local}
  \end{equation}
  where $\mat{S} = \begin{bsmallmatrix*} 0&-1&0\\1&0&0\\0&0&0 \end{bsmallmatrix*}$, $\vect{p}_{ij}$ is the relative position of agent $j$ measured by agent $i$ in its own body frame, and $\psi_{ij}$ is the relative heading of agent $j$ measured by $i$.
  Note that, the columns corresponding to changes in $\vect{p}_{j}$ in the world frame were also rotated to express their motion \wrt{} the perspective of agent $i$.
    The action $\dot{\vect{q}}$ viewed in the local frames of each individual agent is converted to $\dot{\vect{q}}^l$, resolving to
  \begin{equation}
    \begin{aligned}
      \dot{\vect{q}}^l &= k_e{\mat{H}_\mathcal{G}^{l}}^T {\left(     \varkappa_\mathcal{G}\left(\vect{q}_d\right) - \varkappa_{\mathcal{G}}\left(\vect{q}\right) \right)} \\
    &= [[{\vect{u}_{1}}^T,\omega_1],[{\vect{u}_{2}}^T,\omega_2],...,[{\vect{u}_{n}}^T,\omega_n]]^T.
      \label{eq:action}
    \end{aligned}
  \end{equation}
  If the relative observation graph $\mathcal{G}$ is connected, we can construct a decentralized control scheme that enforces the desired formation shared by all observing agents.
  After plugging in eqs. (\ref{eq:relative_poses}), (\ref{eq:relative_pose_function}) and (\ref{eq:rigidity_matrix_local}) to (\ref{eq:action}) the resulting scheme corresponds to eq. (\ref{eq:fec_raw}):
  \begin{footnotesize}
  \begin{equation*}
    \begin{aligned}
    \vect{u}_{i} &= k_e \left(  \sum_{
         j\in\mathcal{N}_i}c_{ij}\left(\vect{p}_{ij}-\vect{p}_{ij}^d\right) - \sum_{
         j\in\mathcal{N}_i}c_{ji} {\mat{R}(\psi_{ji})}^T\left( \vect{p}_{ji} -\vect{p}{ji}^d\right) \right), \\
    \omega_i &= k_e \left( -\sum_{
         j\in\mathcal{N}_i}c_{ij}\left( {\vect{p}_{ij}^{T}} \mat{S} \left( \vect{p}_{ij} - \vect{p}_{ij}^d\right)\right)\right.\\
    &\left.+ \sum_{
         j\in\mathcal{N}_i}c_{ij}\left(\psi_{ij} - \psi_{ij}^d \right) - \sum_{
         j\in\mathcal{N}_i}c_{ji} \left(\psi_{ji} - \psi_{ji}^d \right) \right).
    \end{aligned}
  \end{equation*}
  \end{footnotesize}

\section{Stability analysis}
\label{App:stability}
In order to analyze the stability of the \ac{FEC} system obtained in section \ref{sec:fec}, we propose the following Lyapunov function:
\begin{equation}
  V\left(\vect{e}_F\right) = {\vect{e}_F}^T \mat{P} \vect{e}_F
\end{equation}
where $\mat{P}$ is a constant positive-definite matrix.
The time-derivative of $V$ is 
\begin{equation}
  \dot{V} = {\dot{\vect{e}}_F}^T \mat{P} \vect{e}_F + {\vect{e}_F}^T \mat{P} \dot{\vect{e}}_F.
  \label{eq:lyapunov_dynamic}
\end{equation}
The error dynamics $\dot{\vect{e}}_F$ are obtained as
\begin{equation}
  \dot{\vect{e}}_F = \frac{\partial\vect{e}_F}{\partial t} = \frac{\partial\vect{e}_F}{\partial \varkappa_\mathcal{G}} \frac{\partial \varkappa_\mathcal{G}}{\partial \vect{q}}  \frac{\partial \vect{q}}{\partial t}.
\end{equation}
Using (\ref{App:fec:eq:1}) and (\ref{App:fec:eq:2}), the above resolves to
\begin{equation}
  \dot{\vect{e}}_F = -\mat{I_4} \mat{H}_{\mathcal{G}}^W k_e  {\mat{H}_{\mathcal{G}}^W}^T \vect{e}_F.
\end{equation}
For notational simplicity, we will define $\mat{M} \triangleq  \mat{H}_{\mathcal{G}}^W {\mat{H}_{\mathcal{G}}^W}^T$ and simplify the above to
\begin{equation}
  \dot{\vect{e}}_F = -k_e  \mat{M}\vect{e}_F.
\end{equation}
Plugging the above into (\ref{eq:lyapunov_dynamic}) yields
\begin{equation}
  \begin{aligned}
    \dot{V} &= -k_e\left({\vect{e}_F}^T \mat{M} \mat{P} \vect{e}_F + {\vect{e}_F}^T \mat{P} \mat{M} \vect{e}_F\right)\\
    &= -k_e {\vect{e}_F}^T \left(\mat{M}\mat{P} + \mat{P}\mat{M}\right)\vect{e}^F.
  \end{aligned}
  \label{eq:mp_matrix}
\end{equation}

For the system to be stable in some neighborhood $\mathcal{B}$ of equilibrium $\vect{e}_F = \vect{0}$, it must hold that
\begin{equation}
   \dot{V} < 0 \quad \forall \vect{e}_F \in \mathcal{B} \backslash \{\vect{0}\},
  \label{eq:lyapunov_condition}
\end{equation}
therefore  $\left(\mat{M}\mat{P} + \mat{P}\mat{M}\right)$ must be a positive definite matrix.
This condition is difficult to evaluate for a general case, since $\mat{M}$ depends on the observation graph, desired formation, and current state $\vect{q}$.
We will evaluate the condition for a basic example formation of two agents where agent $1$ observes agent $2$.
In this case,
\begin{equation}
    \mat{H}_{\mathcal{G}}^W = \begin{bmatrix}
      -{\mat{R}(\psi_1)}^T & \mat{S}^T\vect{p}_{12}^W &  \mat{R}(\psi_{1})^T & \vect{0_3} \\
         \vect{0_3}^T & -1 &  \vect{0_3}^T & 1 
    \end{bmatrix},
\end{equation}
and therefore $\mat{M} $ is
\begin{equation}
  \begin{small}
  \mat{M} =
    \begin{bmatrix}
      2+{\left(\left[\vect{p}_{12}^W\right]_2\right)}^2 & -\left[\vect{p}_{12}^W\right]_2 \left[\vect{p}_{12}^W\right]_1 & 0 &  \left[\vect{p}_{12}^W\right]_2 \\
      -\left[\vect{p}_{12}^W\right]_2 \left[\vect{p}_{12}^W\right]_1 &  2+{\left(\left[\vect{p}_{12}^W\right]_1\right)}^2 & 0 &  -\left[\vect{p}_{12}^W\right]_1 \\ 
      0 & 0 & 2 & 0 \\
      \left[\vect{p}_{12}^W\right]_2 & -\left[\vect{p}_{12}^W\right]_1 & 0 & 2
    \end{bmatrix}.
  \end{small}
  \label{eq:default_m}
\end{equation}
Since $\mat{M}$ is symmetric, if its leading principal minors are positive, then it is positive definite.
To prove this is the case, we evaluate all four leading principal minors.
These are
\begin{equation}
\begin{small}
\begin{aligned}
  m_1 &= 2+{\left(\left[\vect{p}_{12}^W\right]_2\right)}^2\\
  m_2 &= 4+2{\left(\left[\vect{p}_{12}^W\right]_2\right)}^2+2{\left(\left[\vect{p}_{12}^W\right]_1\right)}^2\\
  m_3 &= 8+4{\left(\left[\vect{p}_{12}^W\right]_2\right)}^2+4{\left(\left[\vect{p}_{12}^W\right]_1\right)}^2\\
  m_4 &= 16+4{\left(\left[\vect{p}_{12}^W\right]_2\right)}^2+4{\left(\left[\vect{p}_{12}^W\right]_1\right)}^2
\end{aligned}
\end{small}
\end{equation}
In the above, $m_o$ denotes the leading principal minor corresponding to the determinant of the upper-left submatrix of size $o\times o$.
All of the above leading principal minors are positive for $\vect{p}_{12}^W \in \mathbb{R}^3$, and thus $\mat{M}$ is a positive definite matrix.
Therefore, choosing trivial $\mat{P} = \mat{I_4}$ for (\ref{eq:mp_matrix}), condition (\ref{eq:lyapunov_condition}) holds for arbitrary $\mathcal{B}$ and thus the system is stable, converging to $\vect{e}_F = \vect{0}$. 

The above can not be trivially generalized to an arbitrary formation.
In general, $\mat{M}$ will have the format
\begin{equation}
  \begin{small}
    \mat{M} \!=
\mat{H}_{\mathcal{G}}^W {\mat{H}_{\mathcal{G}}^W}^T\!=
    \begin{bmatrix} \mat{E}_1\\\mat{E}_2\\\vdots\end{bmatrix}\!\!\begin{bmatrix} {\mat{E}_1}^T &\!\!{\mat{E}_2}^T &\!\!\!\!\dots\end{bmatrix}
    \!= \!\begin{bmatrix}
      \mat{E}_{11} &\! \mat{E}_{12} &\! \dots\\
      \mat{E}_{21} &\! \mat{E}_{22} &\! \dots\\
      \vdots &\! \vdots &\! \ddots\\
    \end{bmatrix}\!\!,
  \end{small}
\end{equation}
where $\mat{E}_a$ is a set of four rows in $\mat{H}_{\mathcal{G}}^W$ corresponding to the observation graph edge $a$ in the form of (\ref{eq:rigidity_matrix_world}), and we define $\mat{E}_{ab}\triangleq\mat{E}_a{\mat{E}_b}^T$ which is a 4$\times$4 matrix.
When the edges $a$ and $b$ do not share a vertex (an agent), then $\mat{E}_{ab} = \mat{0_4}$.
Each $\mat{E}_{aa}$ has the format of (\ref{eq:default_m}), substituting vertex indices $1$ and $2$ for the indices of vertices connected by edge $a$.
If edges $a$ and $b$ both lead \emph{out of} their shared vertex, $\mat{E}_{ab}$ has the format
\begin{equation}
  \mat{E}_{ab} =
    \begin{bmatrix}
      1+{\left(\left[\vect{p}_{a}^W\right]_2\right)}^2 & -\left[\vect{p}_{a}^W\right]_2 \left[\vect{p}_{a}^W\right]_1 & 0 &  \left[\vect{p}_{a}^W\right]_2 \\
      -\left[\vect{p}_{a}^W\right]_2 \left[\vect{p}_{a}^W\right]_1 &  1+{\left(\left[\vect{p}_{a}^W\right]_1\right)}^2 & 0 &  -\left[\vect{p}_{a}^W\right]_1 \\ 
      0 & 0 & 1 & 0 \\
      \left[\vect{p}_{a}^W\right]_2 & -\left[\vect{p}_{a}^W\right]_1 & 0 & 1
    \end{bmatrix},
\end{equation}
where $\vect{p}_{a}$ is the relative position represented by the observation edge $a$.
If edges $a$ and $b$ both lead \emph{into} their shared vertex, $\mat{E}_{ab}$ has the format of 4$\times$4 identity matrix
\begin{equation}
  \mat{E}_{ab} = 
    \mat{I_4}.
\end{equation}
If edge $a$ leads from vertex $i$ \emph{into} vertex $j$ and edge $b$ leads \emph{out of} vertex $j$, $\mat{E}_{ab}$ has the format
\begin{equation}
  \mat{E}_{ab} = 
    \begin{bmatrix}
      -{R\left(\psi_j\right)}^TR\left(\psi_i\right) & \begin{bmatrix}-\left[\vect{p}_{a}^W\right]_2\\ \left[\vect{p}_{a}^W\right]_1\\0\end{bmatrix}\\
        \begin{bmatrix}0&0&0\end{bmatrix}&-2
    \end{bmatrix}.
\end{equation}
Lastly, if edges $a$ and $b$ connect vertices $i$ and $j$ in mutually opposite directions, $\mat{E}_{ab}$ has the format
\begin{equation}
  \begin{small}
  \mat{E}_{ab} = 
    \begin{bmatrix}
      -{R\left(\psi_i\right)}^TR\left(\psi_j\right)-{R\left(\psi_j\right)}^TR\left(\psi_i\right) &  \begin{bmatrix}-\left[\vect{p}_{a}^W\right]_2\\ \left[\vect{p}_{a}^W\right]_1\\0\end{bmatrix}\\
        \begin{bmatrix}-\left[\vect{p}_{b}^W\right]_2& \left[\vect{p}_{b}^W\right]_1&0\end{bmatrix}&-2
    \end{bmatrix}.
  \end{small}
\end{equation}

The above can serve as building blocks for evaluating the stability of various relative pose-based formations, but a general analysis is out of the scope of this work.

\section{1 active agent and 1 passive agent}
We will analyze the performance of our controller in a simplified scenario with two unidimensional agents with positions $p_{1}[k],p_{2}[k]\in\mathbb{R}$.
Agent 2 is static while agent 1 implements our robust controller, so that its relative position to agent 2 ($p_{12}[k]\triangleq p_{2}[k]-p_{1}[k]$) reaches a desired relative position $p_{12}^d$.
We also introduce the displacement $\Delta_{12}^d[k]\triangleq p_{12}[k]-p_{12}^d$.
Agent 1 has a noisy unbiased measurement $\Delta_{12}^m[k]\triangleq \Delta_{12}^d[k]+e_{12}[k]$ with $e_{12}[k]\sim\mathcal{N}(0,\sigma_{12}[k]^2)$.
The position of agent 1 is then given by:
\begin{equation}
  \label{App:A.2.0}
  \begin{aligned}
    p_{1}[k+1]&=u_{1}[k]+p_{1}[k],\\
    u_{1}[k]&= k_{ef} \mathrm{clamp}(y_{12}[k],\Delta_{12}^m[k])\\
    &=\begin{cases}%
      k_{ef} y_{12}[k] & \text{if\quad} |\Delta_{12}^m[k]|>-\sigma_{12}[k]\Phi^{-1}(\ell),\\%
      0 & \text{otherwise}%
    \end{cases}\\
    y_{12}[k] &=\sign{(\Delta_{12}^m[k])}\sigma_{12}[k]\Phi^{-1}(\ell)+\Delta_{12}^m[k]
  \end{aligned}
\end{equation}
We assume $\ell\in\interval[open left]{0}{0.5}$, and thus $\Phi^{-1}(\ell)\leq0$. Let us start by calculating the conditional probability that the agent does not move at time instant $k$, given $\Delta_{12}^d[k]$. We refer to this probability as the conditional stopping probability:
\begin{equation}
\label{App:A.2.2}
  \begin{aligned}
    \mathrm{P}_{\text{stop}}[k](\Delta_{12}^d[k])&\triangleq\mathrm{P}\left(d_{1}[k]=0\, |\, \Delta_{12}^d[k]\right)\\
    &= \mathrm{P}\left(|\Delta_{12}^d[k]+e_{12}[k]|\leq-\sigma_{12}[k]\Phi^{-1}(\ell)\right)
  \end{aligned}
\end{equation}
Since $e_{12}[k]$ follows a Gaussian distribution,
it follows that:
\begin{equation}
\label{App:A.2.3}
  \begin{aligned}
    \mathrm{P}_{\text{stop}}[k](\Delta_{12}^d[k])&=\Phi\left(-{\Delta_{12}^d[k]}/{\sigma_{12}[k]}-\Phi^{-1}(\ell)\right)\\
    &-\Phi\left(-{\Delta_{12}^d[k]}/{\sigma_{12}[k]}+\Phi^{-1}(\ell)\right).    
  \end{aligned}
\end{equation}
Note, that the arguments of the right-hand side terms of (\ref{App:A.2.3}) are antisymmetric.
This stopping probability is caused by the clamping function.
\begin{figure}
    \centering
    \includegraphics[width=\linewidth]{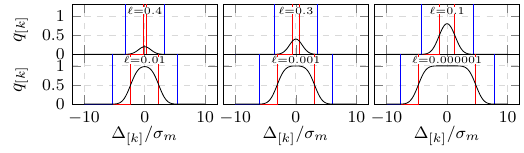}
    \caption{Conditional stopping probability conditioned on the offset $\Delta_{}[k]$ from the measured target position with measurement variance $\sigma_m$.
    The red line shows the symmetric restrained target position derived from $\ell$, while the blue lines delimit the area into which the system state converges akin to proportional control.}
    \label{fig:App.1}
\end{figure}
From Fig. \ref{fig:App.1}, we can see that clamping introduces a spatial probabilistic motion filter. The motion probability of the agent decreases as $|\Delta_{12}^d[k]|$ becomes smaller, with the minimum motion probability:
\begin{equation}
\label{App:A.2.4}
    \bar{\mathrm{P}}_{\text{stop}}[k](0)=2\ell,
\end{equation}
where $\bar{\mathrm{P}}_{\text{stop}}[k](\Delta_{12}^d[k])=1-\mathrm{P}_{\text{stop}}[k](\Delta_{12}^d[k])$.
We observe that the effect of this probabilistic motion filter becomes negligible for $|\hat{\Delta}_{12}^d[k]|\geq$ $ -\Phi^{-1}(\ell)+3$ where $\hat{\Delta}_{12}^d[k]\triangleq\frac{\Delta_{12}^d[k]}{\hat{\sigma}_{12}[k]}$ is the \emph{normalized control error}.
Thus, we can approximate the nonlinear system (\ref{App:A.2.0}) with a switched system 
\begin{equation}
\label{App:A.2.5}
    \!\!p_{1}[k\!+\!1]\!=\!
      \begin{cases}%
        k_{ef}\mathrm{clamp}(y_{12}[k],\Delta_{12}^m[k])+p_{1}[k]\\%
        \hfill \text{if\quad} |\hat{\Delta}_{12}^d[k]|\leq$ $ -\Phi^{-1}(\ell)+3,\\%
        k_{ef}y_{12}[k]+p_{1}[k] \quad \text{otherwise}.
      \end{cases}
\end{equation}
When the agent is close to reaching the target $p_{12}^d$ (i.e., when $
\hat{\Delta}_{12}^d[k]\in\mathcal{S}_{\ell}$ where ${S}_{\ell}=\left[\Phi^{-1}(\ell)-3,-\Phi^{-1}(\ell)+3\right]$), then the switched system (\ref{App:A.2.5}) behaves as the nonlinear system (\ref{App:A.2.0}). Otherwise, the switched system (\ref{App:A.2.5}) behaves like a linear system. 
\subsection{Linear System}
\label{App:A.L}
When the switched system operates as a linear system, we have:
  \begin{equation*}
    \begin{aligned}
      \label{App:A.L.1}
      &p_{1}[k+1]=\hfill\\
      &=k_{ef}\left(\sign{(\Delta_{12}^m[k])}\sigma_{12}[k]\Phi^{-1}(\ell)+\Delta_{12}^m[k]\right)+p_{1}[k],\\
      &=k_{ef}\left(\sign{(\Delta_{12}^d[k]+e_{12}[k])}\sigma_{12}[k]\Phi^{-1}(\ell)+\Delta_{12}^d[k]\right)\\
      &+k_{ef}e_{12}[k]+p_{1}[k],
    \end{aligned}
  \end{equation*}
since $p_2-p_{1}[k]-p_{12}^d=\Delta_{12}^d[k]$, then:  
\begin{equation}
\label{App:A.L.2}
  \begin{aligned}
    \Delta_{12}^d[k+1]&=-k_{ef}\sign{(\Delta_{12}^d[k]+e_{12}[k])}\sigma_{12}[k]\Phi^{-1}(\ell)\nonumber\\
    &-k_{ef}e_{12}[k]+(1-k_{ef})\Delta_{12}^d[k],\nonumber
  \end{aligned}
\end{equation}
We divide the equation by $\sigma_{12}[k]$:
\begin{equation}
\label{App:A.L.3}
  \begin{aligned}
    \frac{\Delta_{12}^d[k+1]}{\sigma_{12}[k]}&=-k_{ef}\sign{(\Delta_{12}^d[k]+e_{12}[k])}\Phi^{-1}(\ell)\nonumber\\
    &-\frac{k_{ef}e_{12}[k]}{\sigma_{12}[k]}+(1-k_{ef})\frac{\Delta_{12}^d[k]}{\sigma_{12}[k]},\nonumber
  \end{aligned}
\end{equation}
Let us assume that the standard deviation of the error for two consecutive instants is approximately constant, i.e., ${\sigma}_{12}[k]\approx {\sigma}_{12}[k+1]$.
Therefore, $\frac{\Delta_{12}[k+1]}{\sigma_{12}[k]}\approx\frac{\Delta_{12}[k+1]}{\sigma_{12}[k+1]}=\hat{\Delta}_{12}^d[k+1]$.
We then take the expected value \wrt{} the measurement noise
\begin{eqnarray}
\label{App:A.L.5}
    \mathbb{E}[\hat{\Delta}_{12}^d[k+1]]&=&-k_{ef}\mathbb{E}[\sign{(\Delta_{12}[k]+e_{12}[k])}]\Phi^{-1}(\ell)\nonumber\\
    &+&(1-k_{ef})\mathbb{E}[\hat{\Delta}_{12}^d[k]],\nonumber
\end{eqnarray}
where
\begin{equation}
\label{App:A.L.6}
  \begin{aligned}
  \mathbb{E}[\sign{(\Delta_{12}[k]+e_{12}[k])}]
    &=\sign(\Delta_{12}[k])\bar{p}_e(\Delta_{12}^d[k])\\
    &-\sign(\Delta_{12}[k])p_e(\Delta_{12}^d[k]),\\ 
  \end{aligned}
\end{equation}
and 
\begin{equation}
\label{App:A.L.7}
  \begin{aligned}
    p_e(\Delta_{12}^d[k])&=\mathrm{P}\left(\sign(\Delta_{12}^d[k]) \neq \sign(\Delta_{12}^d[k] + e_{12}[k])\right)\\
    &=\mathrm{P}\left(\frac{\Delta_{12}^d[k] + e_{12}[k]}{\Delta_{12}^d[k]}<0\right)=1-\Phi\left(\abs{{\hat{\Delta}_{12}^d[k]}}\right).
  \end{aligned}
\end{equation}
However, this system  is valid only for $
\hat{\Delta}_{12}[k] \notin \mathcal{S}_{\ell}$ and under this condition, $p_e(\Delta_{12}[k])\ll1$. 
As a consequence, $\mathbb{E}[\sign{(\Delta_{12}[k]+e_{12}[k])}]\approx\sign(\Delta_{12}[k])$. In addition, since $\sigma_{12}[k]>0$, then $\sign(\Delta_{12}[k])=\sign(\hat{\Delta}_{12}^d[k])$.
Therefore,
\begin{equation}
\label{App:A.L.8}
  \begin{aligned}
    \mathbb{E}\left[\hat{\Delta}_{12}^d[k+1]\right]&\approx-k_{ef}\sign{(\hat{\Delta}_{12}^d[k])}\Phi^{-1}(\ell)\\
    &+(1-k_{ef})\mathbb{E}\left[\hat{\Delta}_{12}^d[k]\right],
  \end{aligned}
\end{equation}
Next, we write the iterative eq. (\ref{App:A.L.8}) as a function of the initial state $\hat{\Delta}_{12}^d[0]$:
\begin{small}
\begin{equation}
\label{App:A.L.9}
  \begin{aligned}
    \mathbb{E}\left[\hat{\Delta}_{12}^d[k\!+\!1]\right]&=-k_{ef}\Phi^{-1}(\ell)\sum_{n=1}^{k}\sign(\hat{\Delta}_{12}^d[n])(1\!-\!k_{ef})^{n\!-\!1}\\
    &+(1-k_{ef})^{k\!+\!1}\hat{\Delta}_{12}^d[0].
  \end{aligned}
\end{equation}
\end{small}%
Since this system is valid for $%
\hat{\Delta}_{12}^d[k] \notin\mathcal{S}_{\ell}$ and $0\in\mathcal{S}_{\ell}$, we then have that $\sign(\hat{\Delta}_{12}^d[k])=\sign(\hat{\Delta}_{12}^d[0])$ for all $k$, as long as this system is valid. Thus, we obtain the following linear system:
\begin{equation}
\label{App:A.L.10}
\begin{aligned}
  \mathbb{E}\left[\hat{\Delta}_{12}^d[k+1]\right]&=-k_{ef}\sign(\hat{\Delta}_{12}^d[0])\Phi^{-1}(\ell)\sum_{n=1}^{k}(1-k_{ef})^{n-1}\\
  &+(1-k_{ef})^{k+1}\hat{\Delta}_{12}^d[0],
  \end{aligned}
\end{equation}
and using the geometric series formula, we obtain:
\begin{equation}
\label{App:A.L.11}
  \begin{aligned}
    \mathbb{E}\left[\hat{\Delta}_{12}^d[k+1]\right]&=-\sign(\hat{\Delta}_{12}^d[0])\Phi^{-1}(\ell)(1-(1-k_{ef})^k+1)\\
    &+(1-k_{ef})^{k+1}\hat{\Delta}_{12}^d[0].
  \end{aligned}
\end{equation}
If $k_{ef}\in(0,2)$, then $|1-k_{ef}|<1$ and the system is stable. Consequently, we have: 
\begin{equation}
\label{App:A.L.12}
   \lim_{k\rightarrow+\infty}\mathbb{E}\left[\hat{\Delta}_{12}^d[k+1]\right]=-\sign(\hat{\Delta}_{12}^d[0])\Phi^{-1}(\ell),
\end{equation}
and $-\sign(\hat{\Delta}_{12}^d[0])\Phi^{-1}(\ell)\in \mathcal{S}_{\ell}$.
This implies that if $\hat{\Delta}_{12}^d[0]\notin\mathcal{S}_{\ell}$ and the system is stable, then for some value $k$, we will have $\hat{\Delta}_{12}^d[k]\in\mathcal{S}_{\ell}$.
This demonstrates that if the system and operating in the linear region, then it will enter into the clamped region eventually.

\subsection{Nonlinear System}
\label{App:A.N}
Now, let us focus on $\hat{\Delta}_{12}^d[k]\in\mathcal{S}_{\ell}$.
We will assume that the standard deviation $\sigma_{12}[k]$ of the measurement error remains approximately constant, since the motion of agent 1 is small.
We will denote this final value of $\sigma_{12}[k]$ as $\sigma$.
\begin{equation}
  \begin{aligned}
    \label{App:A.N.1}
    &\Delta_{12}^d[k+1]=\Delta_{12}^d[k]\\%
    &-k_{ef}\mathrm{clamp}(\sign{(\Delta_{12}^m[k])}{\sigma}\Phi^{-1}(\ell)+\Delta_{12}^m[k],\Delta_{12}^m[k])
  \end{aligned}
\end{equation}
Then, we take the expected value \wrt{} the measurement noise:    
\begin{equation}
\begin{aligned}
  \label{App:A.N.2}
  &\mathbb{E}\left[\Delta_{12}^d[k+1]\right]=\mathbb{E}\left[\Delta_{12}^d[k]\right]\\
  &-k_{ef}\mathbb{E}\left[\mathrm{clamp}(\sign{(\Delta_{12}^m[k])}{\sigma}\Phi^{-1}(\ell)+\Delta_{12}^m[k],\Delta_{12}^m[k])\right],
\end{aligned}
\end{equation}
which becomes:
\begin{equation}
    \label{App:A.N.3}
  \begin{aligned}
    &\mathbb{E}\left[\Delta_{12}^d[k+1]\right]=\mathbb{E}\left[\Delta_{12}^d[k]\right]\\
    &+k_{ef}\mathbb{E}_k\left[\int_{-\infty}^{L_-}\left({\sigma}\Phi^{-1}(\ell)-\Delta_{12}^d[k]-x\right)f_e(x)\mathrm{d}x\right]\\
    &-k_{ef}\mathbb{E}_k\left[\int_{L_+}^{\infty}\left({\sigma}\Phi^{-1}(\ell)+\Delta_{12}^d[k]+x\right)f_e(x)\mathrm{d}x\right]
  \end{aligned}
\end{equation}
where $L_-=-\Delta_{12}^d[k]+\sigma\Phi^{-1}(\ell)$ and $L_+=-\Delta_{12}^d[k]-\sigma\Phi^{-1}(\ell)$.
$\mathbb{E}[\Delta_{12}^d[k]]$ is the expected value \wrt{} all the measurement errors (i.e., $e_{12}[0],e_{12}[1],\cdots,e_{12}[k]$), but $\mathbb{E}_k[\Delta_{12}^d[k]]$ is the expected value \wrt all the measurement errors except the current one (i.e., $e_{12}[0],e_{12}[1],\cdots,e_{12}[k-1]$).
Then $f_e(x)$ is the probability density function of $e_{12}[k]$.
After some algebraic operations and calculating the corresponding integrals, we obtain:
\begin{equation}
\begin{aligned}
\label{App:A.N.4}
    &\mathbb{E}\left[\Delta_{12}^d[k+1]\right]=\mathbb{E}\left[\Delta_{12}^d[k]\right]\\
    &+k_{ef}\sigma\Phi^{-1}(\ell)\mathbb{E}\left[\Phi\left({-\Delta_{12}^d[k]}/{\sigma}+\Phi^{-1}(\ell))\right)\right]\\
  &+k_{ef}\sigma\Phi^{-1}(\ell)\mathbb{E}\left[\Phi\left({-\Delta_{12}^d[k]}/{\sigma}-\Phi^{-1}(\ell)\right)\right]\\
  &-k_{ef}\mathbb{E}\left[\Delta_{12}^d[k]\right]-k_{ef}\sigma\Phi^{-1}(\ell)\\
   &-k_{ef}\mathbb{E}\left[\Delta_{12}^d[k]\Phi\left({-\Delta_{12}^d[k]}/{\sigma}+\Phi^{-1}(\ell)\right)\right]\\
   &+k_{ef}\mathbb{E}\left[\Delta_{12}^d[k]\Phi\left({-\Delta_{12}^d[k]}/{\sigma}-\Phi^{-1}(\ell)\right)\right]\\
  &+\frac{\sigma k_{ef}}{\sqrt{2\pi}}\mathbb{E}\left[\exp\left(-(-\Delta_{12}^d[k]+\Phi(\ell))^2/2\sigma^2\right)\right]\\
  &-\frac{\sigma k_{ef}}{\sqrt{2\pi}}\mathbb{E}\left[\exp\left(-(-\Delta_{12}^d[k]-\Phi(\ell))^2/2\sigma^2\right)\right]
\end{aligned}
\end{equation}
We approximate the nonlinear terms in the previous equation  with their first-order Taylor series \wrt{} $\Delta_{12}^d[k]$ centered around $\Delta_{12}^d[k]=0$.
Thus, we obtain:
  \begin{equation}
  \label{App:A.N.5}
    \begin{aligned}
      &\mathbb{E}\left[\Delta_{12}^d[k+1]\right]\approx-2k_{ef}\ell\mathbb{E}[\Delta_{12}^d[k]]+\mathbb{E}\left[\Delta_{12}^d[k]\right]\\%
      &-k_{ef}\sqrt{\frac{2}{\pi}}\Phi^{-1}(\ell)\exp\left(-\frac{(\Phi^{-1}(\ell))^2}{2\sigma^2}\right)\mathbb{E}\left[\Delta_{12}^d[k]\right]\\
      &+k_{ef}\sqrt{\frac{2}{\sigma^2\pi}}\Phi^{-1}(\ell)\exp\left(-\frac{(\Phi^{-1}(\ell))^2}{2\sigma^2}\right)\mathbb{E}\left[\Delta_{12}^d[k]\right]\\
      &=(1-k_{ef}^{'})\mathbb{E}\left[\Delta_{12}^d[k]\right],
    \end{aligned}
  \end{equation}
where
\begin{equation}
\label{App:A.N.6}
   k_{ef}^{'}\!=\!k_{ef}\left(\!\!\left(1\!-\!\frac{1}{\sigma}\right)\!\sqrt{\frac{2}{\pi}}\Phi^{-1}(\ell)\exp\!\!\left(-\!\frac{(\Phi^{-1}(\ell))^2}{2\sigma^2}\right)\!+\!2\ell\right)\!\!.
\end{equation}
As long as $|1-k_{ef}^{'}|\in(0,1)$, the system (\ref{App:A.N.5}) will be stable and we will have:
\begin{equation}
\label{App:A.N.7}
 \lim_{k\rightarrow\infty}   \mathbb{E}\left[\Delta_{12}^d[k]\right]=0.
\end{equation}
If $k_{ef}^{'}\in(0,1)$, then the system operates in an overdamped regime, but if $k_{ef}^{'}\in(1,2)$, then it operates in an underdamped regime.

The conditional variance of the system given $\Delta_{12}^d[k]=0$, which corresponds to the expected value of the system in the steady state was obtained using a symbolic solver.
It is given by
\begin{equation}
  \label{App:A.N.8}
  \begin{aligned}
    &\mathrm{var}[\Delta_{12}^d[k\!\!+\!\!1]|\Delta_{12}^d[k]\!\!=\!\!0]=2k_{ef}^2\sigma^2\Phi^{-1}\!(\ell)\ell+2k_{ef}^2\sigma^2\!\ell\\
    &+2k_{ef}^2\sigma^2\Phi^{-1}(\ell)/\sqrt{2\pi}\exp\left(-(\Phi^{-1}(\ell))^2/2\right).
  \end{aligned}
\end{equation}
This equation was obtained by first setting $\Delta_{12}^d[k]=0$ and following a similar procedure as was used to calculate the expected value in (\ref{App:A.N.4}). The conditional variance (\ref{App:A.N.8}) is a strictly increasing function of $\ell$ for $\ell\in[0,0.5]$.

We refer to the duration $T$ expressing the number of consecutive time steps in which $p_1[k]$ remains constant ($\forall o \in \interval{1}{T} : p_{1}[k+o] = p_{1}[k]$) as the coherence time $k_{\text{coh}}$ of the system. This coherence time conditioned on $\Delta_{12}^d[k]$ is a random variable that follows a geometrical distribution
\begin{equation}
\label{App:A.2.6}
   \mathrm{P}(T=n\,|\, \Delta_{12}^d[k]=z)=({\mathrm{P}}_{1}[k](z))^{n-1}\bar{\mathrm{P}}_{\text{stop}}[k](z), 
\end{equation}
with conditional expected value
\begin{equation}
\label{App:A.2.7}
    \mathbb{E}[T\, |\, \Delta_{12}^d[k]=z)]=\frac{1}{\bar{\mathrm{P}}_{\text{stop}}[k](z)}.    
\end{equation}
The unconditional distribution and expected values are:
\begin{eqnarray}
\label{App:A.2.8}
\mathrm{P}(T=n )&=& \int_{-\infty}^{+\infty}\mathrm{P}(T=n\,|\, \Delta_{12}^d[k]=z)f_{\Delta_{12}^d}[k](z)\mathrm{d}z\nonumber,\\
\label{App:A.2.9}
  \mathbb{E}[T]&=& \int_{-\infty}^{+\infty}\frac{f_{\Delta_{12}^d}[k](z)}{\bar{\mathrm{P}}_{\text{stop}}[k](z)}\mathrm{d}z,
\end{eqnarray}
where $f_{\Delta_{12}^d}[k](z)$ is the probability density function of $\Delta_{12}^d[k]$ during the steady state.
Analytically deriving such a probability density function is complicated, but after performing a large-scale numerical analysis using simulations, we observed that
\begin{enumerate}
  \item{the steady-state probability distribution $f_{\Delta_{12}^d}[k](z)$ of the system (\ref{App:A.N.1}) can be considered Gaussian for the range $\{\ell,k_{ef}\}\in[0.1,0.5]\times [0.1,1]$.
    This hypothesis is validated by the extremely low values of the Kullback-Leiber divergence calculated between the density function obtained through simulations and the Gaussian distribution, see Table \ref{tab:KullbackLeiblerDivergence};}
  \item{the variance of $\Delta_{12}^d[k]$ decreases with $\ell$;}
    \item{for $\ell\in[0.01,0.5]$ and $k_{ef}\in[0.1,1.9]$, a good approximation for the variance of $\Delta_{12}^d[k]$ is
    \footnote{This approximation was obtained heuristically by an extensive analysis of the simulation data and its error was observed to be significantly low.}:
\begin{equation}
\label{App:A.2.10}
  {\sigma_{ss,\text{res}}}^2=\frac{k_{ef}\sigma^2}{2-k_{ef}}\exp{\left(\beta(k_{ef})\Phi^{-1}(\ell)\right)}
\end{equation}
where  $\beta(k_{ef})$ is given in Table \ref{tab:Beta}.
    }
\end{enumerate}
\begin{table}
  \begin{center}
    \begin{tabular}{c|c|c|c|} 
      & $k_{ef}=0.1$ & $k_{ef}=0.5$ & $k_{ef}=1.0$ \\
      \hline     
      $\ell=0.1$ & 0.0018 & 0.0035  & 0.0122  \\
      \hline
      $\ell=0.3$ & 0.0005 & 0.0009 & 0.0051  \\
      \hline
      $\ell=0.5$ & 0.0003 & 0.0002 & 0.0002  \\
      \hline
    \end{tabular}
  \end{center}
  \vspace{-1em}
    \caption{Kullback-Leibler divergence between the $\Delta_{12}^d[k]$ probability density function measured from simulations $p_{S}$ and $p_{G}(z)=1/(\hat{\sigma}_{m}\sqrt{2\pi})\exp\!\left(-z^2/2\hat{\sigma}_{m}^2\right)$. $\hat{\sigma}_{m}$ is the variance of $\Delta_{12}^d[k]$ estimated from simulations.\label{tab:KullbackLeiblerDivergence}}
\end{table}
\begin{table}
  \begin{center}
    \begin{tabular}{c|c|c|c|c|c} 
      $k_{ef}$ & 0.1 & 0.5 & 1.0 & 1.5 & 1.9 \\
      \hline     
      $\beta(k_{ef})$ & 0.7251 & 0.8266 & 1.043 & 1.498 & 3.177
    \end{tabular}
  \end{center}
  \vspace{-1em}
    \caption{Values of $\beta(k_{ef})$ for different $k_ef$.\label{tab:Beta}}
\end{table}
We then observe in Table \ref{tab:CoherenceTime} a comparison of the expected coherence time estimated by simulation and the numerical approximation.
\begin{table}
  \begin{center}
    \begin{tabular}{c|c|c|c|c|c|c} 
      $\ell$ & 0.45 & 0.3 & 0.2 & 0.1 & 0.05\\
      \hline
       From simulations & 1.1084 & 1.6491 & 2.4722 & 4.8897 & 9.6895 \\
       From (\ref{App:A.2.9}), (\ref{App:A.2.10}) & 1.1083 & 1.6494 & 2.4604 & 4.8912 & 9.7489 \\
    \end{tabular}
  \end{center}
    \caption{Expected coherence time $k_{\text{coh}}$ ($\sigma_m=0.1$, $k_{ef}=0.1$).\label{tab:CoherenceTime}}
\end{table}
\section{2 active agents}
\label{App:C}
We consider again the same unidimensional scenario, except that now agent 2 is also active and implements the same controller as agent 1.
This scenario corresponds to the real-world case when two agents are mutually observing.
The dynamics of such agents are expressed as follows:
\begin{equation}
  \begin{aligned}
    p_{1}[k+1]&=k_{ef}\mathrm{clamp}(y_{12}[k],\Delta_{12}^m[k])+p_{1}[k],\\
    p_{2}[k+1]&=k_{ef}\mathrm{clamp}(y_{21}[k],\Delta_{21}^m[k])+p_{2}[k],\\
    y_{ij}[k] &=\sign{(\Delta_{ij}^m[k])}\sigma_{ij}[k]\Phi^{-1}(\ell)+\Delta_{ij}^m[k],\\    
    \Delta_{ij}^m[k]&=\Delta_{ij}^d[k]+e_{ij}[k],\\
    \Delta_{ij}^d[k]&=p_{j}[k]-p_{i}[k]-p_{ij}^d,
  \end{aligned}
\end{equation}
where $e_{ij}[k]$ and $e_{ji}[k]$ are statistically independent, but we will assume for simplicity that they have the same variance.
Thus, $p_{21}^d=-p_{12}^d$, although in general, $\Delta_{21}^m[k]\neq -\Delta_{12}^m[k]$.
Next, we analyse the following system:
 \begin{eqnarray}
 \label{App:C.0.3}
   \Delta_{12}^d[k+1]&=&k_{ef}\mathrm{clamp}(y_{21}[k],\Delta_{21}^m[k])\\  
   &-&k_{ef}\mathrm{clamp}(y_{12}[k],\Delta_{12}^m[k])+\Delta_{12}^d[k]\nonumber
 \end{eqnarray} 
Using similar reasoning as in the previous case for 1 agent, we approximate system (\ref{App:C.0.3}) with the following switched system
\begin{equation}
\label{App:C.0.4}
  \begin{aligned}
    \Delta_{12}^d[k\!+\!1]&\!=\!\!
      \begin{cases}%
        k_{ef}C_{12}[k]+\Delta_{12}^d[k\!+\!1]\\
        \hspace{0.5cm} \text{if\quad} |\hat{\Delta}_{12}^d[k]|\leq$ $ -\Phi^{-1}(\ell)+3,\\%
        k_{ef}(y_{21}[k]-\!y_{12}[k])+\Delta_{12}^d[k\!+\!1]~\text{otw.},
      \end{cases}\\%
    C_{12}[k]&=\mathrm{clamp}(y_{21}[k],\Delta_{21}^m[k])\\
    &-\mathrm{clamp}(y_{12}[k],\Delta_{12}^m[k]),\\
  \end{aligned}
\end{equation}
with $\hat{\Delta}_{12}^d[k]\!\!=\!\Delta_{12}^d[k]/\sigma_{}[k]$.
For $\hat{\Delta}_{12}^d[k]\!\!\notin\!\mathcal{S}_{\ell}$ using (\ref{App:A.L.8}), we obtain:
\begin{equation}
    \begin{aligned}
      \label{App:C.0.5}
      \mathbb{E}\left[\hat{\Delta}_{12}^d[k\!+\!1]\right]&\approx k_{ef}\Phi^{-1}(\ell)\left(\sign{(\hat{\Delta}_{21}^d[k])}-\sign{(\hat{\Delta}_{12}^d[k])}\right)\\
      &+(1-k_{ef})\mathbb{E}\left[\hat{\Delta}_{12}^d[k]\right]+k_{ef}\mathbb{E}\left[\hat{\Delta}_{21}^d[k]\right],\\
      &=-2 k_{ef}\Phi^{-1}(\ell)\sign{(\hat{\Delta}_{12}^d[k])}\\
      &+(1-2k_{ef})\mathbb{E}\left[\hat{\Delta}_{12}^d[k]\right]
    \end{aligned}
\end{equation}
From (\ref{App:C.0.5}), we observe that this system is stable for $|1-2k_{ef}|\leq1$, i.e., for $k_{ef}\in[0,1]$ as opposed to the system (\ref{App:A.L.8}) that is stable for $k_{ef}\in[0,2]$.
This shows that as the number of agents interacting with each other increases, the range of the gain $k_{ef}$ over which the system is stable shrinks.

Finally, if  $k_{ef}\in[0,1]$, then the system (\ref{App:C.0.5}) will enforce $\hat{\Delta}_{12}^d[k]\in\mathcal{S}_{\ell}$ for some $k$.
Once this happens, the probability that one agent has its input clamped rapidly increases, leading to the 1 mobile agent case where we have established that $\mathbb{E}[\Delta_{12}^d[k]]\rightarrow0$.
Eventually, the other agent may stop having its input clamped, but $\Delta_{12}^d[k]$ will be smaller, and the probability that any of the agents will have its input clamped will be even higher.
This demonstrates the stability of the system.%
\begin{acronym}
  \acro{CNN}[CNN]{Convolutional Neural Network}
  \acro{IR}[IR]{infrared}
  \acro{GNSS}[GNSS]{Global Navigation Satellite System}
  \acro{MOCAP}[mo-cap]{Motion capture}
  \acro{MPC}[MPC]{Model Predictive Control}
  \acro{MRS}[MRS]{Multi-robot Systems Group}
  \acro{ML}[ML]{Machine Learning}
  \acro{MAV}[MAV]{Micro-scale Unmanned Aerial Vehicle}
  \acro{UAV}[UAV]{Unmanned Aerial Vehicle}
  \acro{UV}[UV]{ultraviolet}
  \acro{UVDAR}[\emph{UVDAR}]{UltraViolet Direction And Ranging}
  \acro{UT}[UT]{Unscented Transform}
  \acro{RTK}[RTK]{Real-Time Kinematic}
  \acro{ROS}[ROS]{Robot Operating System}
  \acro{wrt}[w.r.t.]{with respect to}
  \acro{FEC}[FEC]{Formation-Enforcing Control}
  \acro{DIFEC}[DIFEC]{Distributed Formation-Enforcing Control}
  \acro{LIDAR}[LiDAR]{Light Detection And Ranging}
  \acro{UWB}[UWB]{Ultra-wideband}
\end{acronym}
\bibliographystyle{elsarticle-num}
\bibliography{bibfile}
\end{document}